\documentclass[lettersize,journal]{IEEEtran}

\makeatletter
\def\endthebibliography{%
	\def\@noitemerr{\@latex@warning{Empty `thebibliography' environment}}%
	\endlist
}
\makeatother

\usepackage{amsmath,amsfonts}
\usepackage{algorithmic}
\usepackage{algorithm}
\usepackage{array}
\usepackage[caption=false,font=scriptsize,labelfont=sf,textfont=sf]{subfig}
\usepackage{textcomp}
\usepackage{stfloats}
\usepackage{url}
\usepackage{verbatim}
\usepackage{graphicx}
\usepackage{cite}
\usepackage{hyperref}       % hyperlinks
\hypersetup{colorlinks=true,urlcolor=black}
\usepackage{booktabs}       % professional-quality tables
\usepackage{nicefrac}       % compact symbols for 1/2, etc.
\usepackage{microtype}      % microtypography
\usepackage{epsfig}
\usepackage{amssymb}
\usepackage{amsthm}
\usepackage{threeparttable}
\usepackage[export]{adjustbox}
\usepackage[table]{xcolor}
\usepackage{diagbox}
\usepackage{makecell}
\usepackage{multirow}
\usepackage{etoolbox}
\usepackage{bbm}
\usepackage{color}
\usepackage{footnote}
\usepackage{enumitem}
\usepackage{times}
\usepackage{ragged2e}
\usepackage{arydshln}
\usepackage{sail}

\newtheorem{problem}{Problem}

\newcommand{\blue}[1]{{\color{black}#1}}

\def\mymethod{CMAP}

\begin{document}

\title{Adversarial Purification by Consistency-aware Latent Space Optimization on Data Manifolds}

\author{Shuhai Zhang$^*$, Jiahao Yang$^*$, Hui Luo, Jie Chen, Li Wang, Feng Liu, Bo Han, Mingkui Tan$^\dagger$
        % <-this % stops a space
% 
    \thanks{Shuhai Zhang and Mingkui Tan are with the School of Software Engineering, South China University of Technology, also with the Pazhou Laboratory, Guangzhou, China. E-mail: shuhaizhangshz@gmail.com, mingkuitan@scut.edu.cn.}
    \thanks{Jiahao Yang and Feng Liu is with the School of Computing and Information Systems, The University of Melbourne, Australia. E-mail: yijiahao81@gmail.com, fengliu.ml@gmail.com.}
    \thanks{Hui Luo is with the State Key Laboratory of Optical Field Manipulation Science and Technology, Institute of Optics and Electronics, CAS, Chengdu, China. E-mail: luohui19@mails.ucas.ac.cn.}
    \thanks{Jie Chen is with the School of Electronic and Computer Engineering, Peking University, Beijing 100871, China, and also with Peng Cheng Laboratory, Shenzhen 518066, China. E-mail: jiechen2019@pku.edu.cn}
    \thanks{Li Wang is with the Department of Mathematics and the Department of Computer Science and Engineering, University of Texas at Arlington, Arlington, TX 76019 USA. E-mail: li.wang@uta.edu.}
    % \thanks{Feng Liu is with Computing and Information Systems, University of Melbourne. E-mail: fengliu.ml@gmail.com.}
    \thanks{Bo Han is with Department of Computer Science, Hong Kong Baptist University. E-mail: bhanml@comp.hkbu.edu.hk.}
    \thanks{$^*$Authors contributed equally, $^\dagger$ Corresponding author.}
}
% The paper headers
\markboth{Journal of \LaTeX\ Class Files,~Vol.~14, No.~8, August~2021}%
{Shell \MakeLowercase{\textit{et al.}}: A Sample Article Using IEEEtran.cls for IEEE Journals}

% \IEEEpubid{0000--0000/00\$00.00~\copyright~2021 IEEE}
% Remember, if you use this you must call \IEEEpubidadjcol in the second
% column for its text to clear the IEEEpubid mark.

\maketitle

\begin{abstract}
Deep neural networks (DNNs) are vulnerable to adversarial samples crafted by adding imperceptible perturbations to clean data, potentially leading to incorrect and dangerous predictions. Adversarial purification has been an effective means to improve DNNs robustness by removing these perturbations before feeding the data into the model. 
However, it faces significant challenges in preserving key structural and semantic information of data, as the imperceptible nature of adversarial perturbations makes it hard to avoid over-correcting, which can destroy important information and degrade model performance.
In this paper, we break away from traditional adversarial purification methods by focusing on the clean data manifold. To this end, we reveal that samples generated by a well-trained generative model are close to clean ones but far from adversarial ones. Leveraging this insight, we propose Consistency Model-based Adversarial Purification (\mymethod), which optimizes vectors within the latent space of a pre-trained consistency model to generate samples for restoring clean data. 
Specifically, 1) we propose a \textit{Perceptual consistency restoration} mechanism by minimizing the discrepancy between generated samples and input samples in both pixel and perceptual spaces. 2) To maintain the optimized latent vectors within the valid data manifold, we introduce a \textit{Latent distribution consistency constraint} strategy to align generated samples with the clean data distribution. 3) We also apply a \textit{Latent vector consistency prediction} scheme via an ensemble approach to enhance prediction reliability. 
\mymethod~fundamentally addresses adversarial perturbations at their source, providing a robust purification. Extensive experiments on CIFAR-10 and ImageNet-100 show that our \mymethod~significantly enhances robustness against strong adversarial attacks while preserving high natural accuracy. 
Our code is available at \url{https://github.com/ZSHsh98/CMAP}.
\end{abstract}

\begin{IEEEkeywords}
Deep neural networks, Adversarial samples, Adversarial purification, Consistency models.
\end{IEEEkeywords}

\section{Introduction}
\IEEEPARstart{D}{eep} neural networks (DNNs) have achieved remarkable success across various domains, including image classification \cite{he2016deep,zhuang2018discrimination}, 
natural language processing \cite{bahdanau2014neural,de2024human} and speech recognition \cite{Alexei2021,afouras2018deep}, and even in complex tasks like navigation planning \cite{10120966,zeng2019graph}. However, DNNs are vulnerable to \textit{adversarial samples} \cite{goodfellow2014explaining,chen2022adversarial}, which are typically generated by introducing  small and often imperceptible perturbations to inputs, causing the model to yield undesirable outputs \cite{8725541}.
These adversarial samples pose a significant  threat to the deployment of DNNs in real-world scenarios, particularly in safety-critical applications, \eg, autonomous navigation systems \cite{wu2023human,wu2022yolop} and medical diagnosis \cite{ozbulak2019impact,zhao2021diagnose}. As DNNs become increasingly integrated into our lives, developing advanced adversarial defense methods becomes very urgent and imperative.

\begin{figure}[t]
    \begin{center}
        \subfloat[{CIFAR-10, consistency model} \cite{song2023consistency}]
        {\includegraphics[width=0.46\linewidth]{ 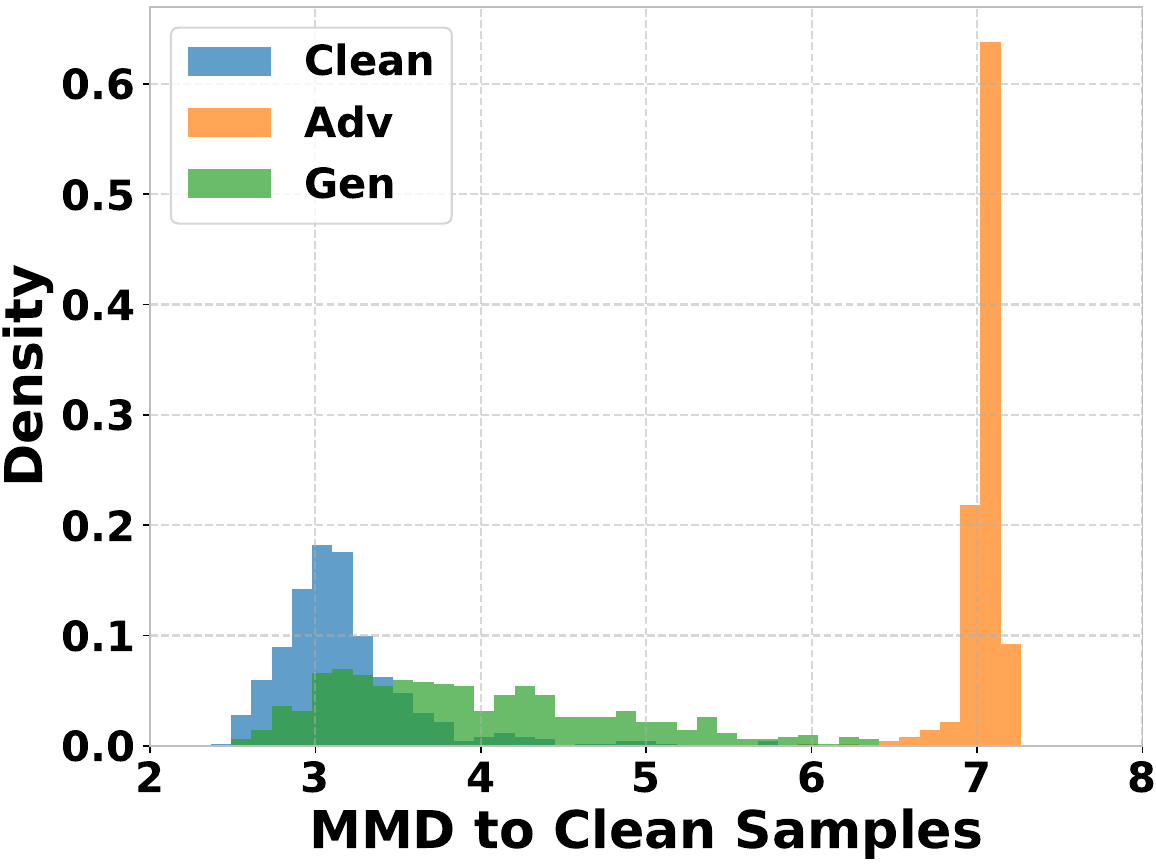}}
        \subfloat[ImageNet, diffusion model \cite{dhariwal2021diffusion}]
        {\includegraphics[width=0.49\linewidth]{ 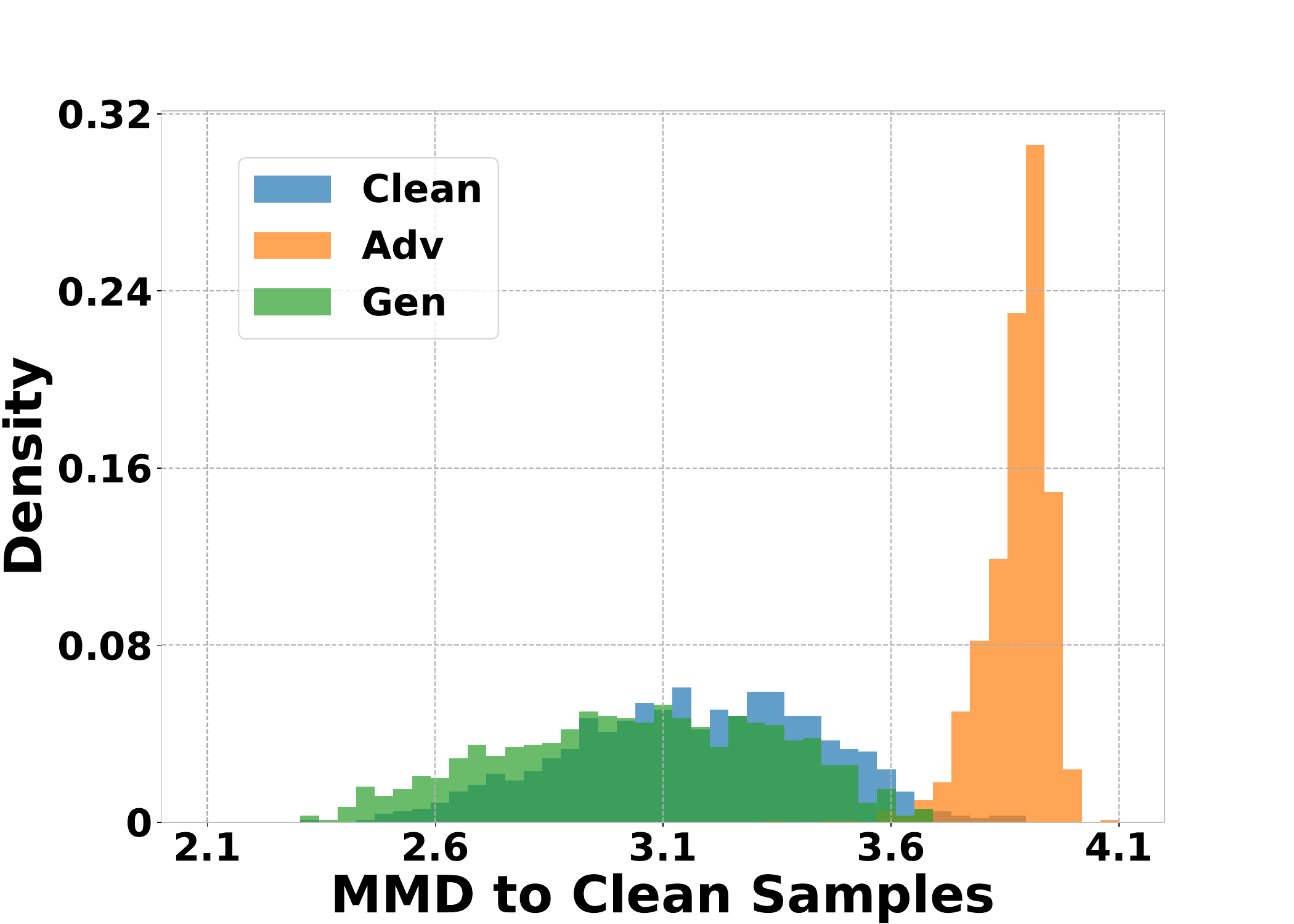}}
        \caption{ Histograms of MMD distances \cite{gretton2012kernel} between the features of clean (Cln) and clean samples \textit{v.s.} generated (Gen) and clean samples \textit{v.s.} adversarial (Adv) samples and clean samples on CIFAR-10 and ImageNet. The results demonstrate that the generated samples are close to the clean ones but are far from the adversarial ones.
        }
        \vspace{-12pt}
        \label{fig: motivition}
    \end{center}
\end{figure}

To improve the robustness of models against these threats, numerous adversarial defense strategies have been proposed \cite{madry2017towards,gowal2021improving,yoon2021adversarial,nie2022diffusion,lee2023robust,pei2025diffusion}. Among them, \textit{adversarial training} \cite{madry2017towards,pang2022robustness,zheng2020efficient,de2022make,gowal2021improving} stands out as a typical defense method, which involves augmenting the training data set with adversarial samples and then iteratively training the model with these augmented data, thereby enhancing its ability to defend these attacks. However, one significant drawback is the substantial computational cost of generating adversarial samples and incorporating them into the training process \cite{Wong2020Fast}.  Moreover, adversarial training tends to be specific to some types of attacks during training, leaving the model vulnerable to unseen attacks \cite{shionline,hillstochastic,yoon2021adversarial,nie2022diffusion}. 

Another parallel route to improve robustness is \textit{ adversarial purification}, which preprocesses inputs to eliminate or neutralize adversarial perturbations with generative models before feeding them into the model \cite{shionline,hillstochastic,yoon2021adversarial,nie2022diffusion}. This process is independent of the model architecture or training, making it  effective against various adversarial attacks compared with adversarial training, providing a versatile defense mechanism. However,  it still suffers from mode collapse and low-quality sampling in generative models, which can lead to incomplete purification and the introduction of artifacts \cite{goodfellow2014generative, lecun2006tutorial, nie2022diffusion}.

Recently, given the powerful capabilities of diffusion models in generating high-quality samples \cite{song2019generative, dhariwal2021diffusion,rombach2022high,ramesh2022hierarchical}, \textit{diffusion-based adversarial purification} methods \cite{yoon2021adversarial,nie2022diffusion,lee2023robust,pei2025diffusion} have shown state-of-the-art defense performance. The diffusion model operates by a forward diffusion process that gradually adds noise to an input, followed by a reverse denoising process that removes this noise to recover the original data \cite{song2019generative,ho2020denoising}. A key advantage is its ability to purify noisy samples back to a clean distribution, incorporating randomness, which enhances its effectiveness as an adversarial purification tool  \cite{nie2022diffusion,lee2023robust}. Given these merits, existing adversarial purification strategies \cite{nie2022diffusion,lee2023robust} add noise to adversarial samples at a specific diffusion timestep and recover the diffused samples through the reverse denoising.
However, \textbf{two major challenges remain}: 1) \textit{ Incomplete removal of adversarial noise}: Adversarial noise often differs significantly from the noise added during the diffusion process, so even with denoising, only portions of the noise resembling the added noise can be removed, making it challenging to completely eliminate adversarial perturbations, especially on large-scale datasets \cite{nie2022diffusion,lee2023robust}. 2) \textit{Inconsistent noise levels}:  A specific diffusion timestep struggles to handle varying noise-level attacks, making it difficult to use a unified approach against diverse adversarial attacks \cite{nie2022diffusion,lee2023robust}.

In this paper, we break away from traditional adversarial purification methods by focusing on the clean data \textit{manifold}. Using generative models, we seek to identify latent vectors corresponding to the input samples on this manifold, thus fundamentally removing the potential adversarial perturbations from the data source. To this end, we reveal a crucial observation: \textbf{samples generated by a well-trained generative model are close to clean ones but far from adversarial ones}. As shown in Fig. \ref{fig: motivition}, the MMD distances \cite{gretton2012kernel} between generated and clean sample features are \textit{comparable} to those between clean samples, but are \textit{significantly smaller} than those between adversarial and clean samples. This suggests that the latent space of the generative model aligns with the clean data manifold, enabling it to generate samples that resemble clean data and are far from adversarial perturbations.
Based on this, we can optimize a latent vector within the latent space of the generative model to generate a sample as close to the original as possible, restoring the sample to the clean data manifold.

Unlike traditional diffusion-model-based adversarial purification methods \cite{yoon2021adversarial,nie2022diffusion,lee2023robust,pei2025diffusion}, we employ a pre-trained consistency model \cite{song2023consistency}, a diffusion model with deterministic generation and efficient one-step image synthesis, making it well-suited for this task. With the pre-trained consistency model, we propose a \textbf{Consistency Model-based Adversarial Purification (\mymethod)} method, as illustrated in Fig. \ref{fig: overview}. 
During \textit{optimization}, we introduce a \textbf{perceptual consistency restoration} mechanism, which optimizes latent vectors within the latent space of the pre-trained consistency model, projecting the input sample back onto the data manifold. To maintain these optimized latent vectors in the valid manifold and avoid overfitting to adversarial noise, we enforce a \textbf{latent distribution consistency constraint} strategy, ensuring alignment with the underlying distribution of clean data. For the final \textit{prediction}, we employ a \textbf{latent vector consistency prediction} scheme to aggregate labels from multiple optimized latent vectors for each input, leveraging their diversity to improve the stability and reliability of the overall outputs.
By focusing on the clean data manifold and leveraging the latent space of a pre-trained consistency model, \mymethod~fundamentally addresses adversarial noise at its source. Unlike traditional defenses operating in the input space, our approach optimizes latent vectors that inherently align with the structure of clean data, effectively filtering out adversarial perturbations off the clean data manifold.
 
Specifically, the \textit{perceptual consistency restoration} mechanism minimizes the mean absolute error (MAE\cite{willmott2005advantages}) while maximizing the structure similarity index measure (SSIM \cite{brunet2011mathematical}) between the test input and the sample generated by the optimized latent vector. The rationale is that the adversarial samples typically reside close to the data manifold of clean examples in pixel space \cite{wang2022guided,prakash2018deflecting,yuan2019adversarial}. 
MAE focuses on correcting pixel-wise deviations, ensuring generated samples closely resemble the test sample. 
Meanwhile, SSIM restores perceptual and structural features \cite{zhao2016loss}, ensuring the recovered samples align with the manifold on pixel discrepancies while preserving the essential high-level features that define the data distribution. This dual optimization strategy leverages the manifold structure to achieve both pixel-level fidelity and perceptual integrity, resulting in semantically meaningful generated images.

The \textit{latent distribution consistency constraint} strategy imposes a penalty on the optimized latent vectors during optimization using a mean squared error (MSE \cite{wang2009mean}) loss, which enforces a Gaussian distribution constraint across multiple sampled latent vectors for each test sample. This ensures that the latent vectors remain aligned with the \textit{clean} data manifold, facilitating the generation of samples that reflect the clean data distribution.
MSE is particularly effective here as it offers a straightforward and effective way to quantify distributional differences by focusing on means and variances, which are the key characteristics defining the manifold structure. 

The \textit{latent vector consistency prediction} scheme employs label voting across multiple optimized latent vectors for each test sample. 
This scheme ensures that the final prediction is robust and closely aligned with the underlying data manifold by fully utilizing the diversity of latent vectors for each sample.
By aggregating predictions from multiple points on the manifold, it reduces the risk of outliers or deviations caused by adversarial perturbations. 
Similar multi-sampling defense mechanisms have been effectively used in prior works \cite{yoon2021adversarial,xiao2023densepure}.

Furthermore, we provide a consistency-disruption attack tailored to our \mymethod. The attack results highlight the strong robustness of \mymethod, suggesting that crafting adversarial samples within the latent space of a well-trained diffusion model is difficult, which aligns with the findings in Fig. \ref{fig: motivition}.
We also empirically compare our method with the latest adversarial training and purification methods against various strong attack benchmarks \cite{croce2020reliable,lee2023robust}. Extensive experiments on CIFAR-10 and ImageNet-100 demonstrate the superior performance of our method.  {Notably, our \mymethod~shows absolute improvements of up to $18.73\% \uparrow$ against  PGD\cite{madry2017towards}+EOT \cite{athalye2018synthesizing} attacks and $6.54\% \uparrow$ against AutoAttack \cite{croce2020reliable} on CIFAR-10}, and up to $6.47\% \uparrow$ against PGD+EOT attacks on ImageNet-100 compared with SOTA methods, respectively, in robust accuracy.  

We summarize our main contributions as follows: 
\begin{itemize}[leftmargin=*]
    \item We reveal that samples generated by a well-trained generative model are close to clean ones but far from adversarial ones. Leveraging this, we move beyond input-space purification toward the clean data manifold. We propose \mymethod, which, to the best of our knowledge, is the first adversarial purification framework to exploit consistency models' deterministic generation for latent-space optimization. \mymethod~optimizes latent vectors to generate samples resembling the original, effectively restoring them to the clean data manifold.

    \item We introduce a perceptual consistency restoration mechanism to align generated samples with the data manifold at the pixel level while preserving high-level perceptual features. Additionally, we theoretically show that a distribution shift occurs between the latent vectors of clean and their adversarial samples, potentially introducing adversarial noise during restoration. To address this, we enforce a latent distribution consistency constraint strategy to maintain optimized latent vectors within the valid manifold, ensuring the samples are generated from the clean data distribution.

    \item We incorporate a latent vector consistency prediction scheme to enhance the stability and reliability of the final prediction. We also provide a consistency-disruption attack tailored to our \mymethod. Experiments on CIFAR-10 and ImageNet-100 exhibit the superior performance of \mymethod~in robust and standard accuracy across various attacks. Notably, \mymethod~is agnostic to classifier architectures or attack types, using  a unified defense operation, unlike prior adversarial purification that requires distinct hyperparameters for different attacks.
\end{itemize}

%-------------------------------------------------------------------------

\section{Related Work}
\subsection{Diffusion Models and Consistency Model}
Diffusion models have emerged as a potent family of probabilistic generative models, exhibiting outstanding performance across various applications \citet{ho2020denoising,song2019generative,song2020improved,songscore}. The key to all approaches in this family is to progressively perturb images to noise via Gaussian perturbations and then generate samples from noise via sequential denoising steps. The concept of the diffusion model originated from Sohl-Dickstein et al. \cite{sohl2015deep}. Enlightened by non-equilibrium statistical physics, they use two learnable Markov chains to implement the perturbation and denoising process. Subsequently, Ho et al. \cite{ho2020denoising} introduce Denoised Diffusion Probabilistic Model (DDPM) to reconstruct noise instead of image in the reverse  by training a U-Net network. Song et al. \cite{song2019generative,song2020improved} propose a similar strategy, called the Score-based Generative Model (SGM), which learns the score function of the intensifying noise sequence injected to images using deep neural networks, and leverages annealed Langevin dynamics to remove noise and generate samples. Immediately after that, DDPM and SGM are unified into the form of the score stochastic differential equation by Song et al. \cite{songscore}, extending the case to infinite time steps or noise levels, leading to the continuous-time diffusion model.

However, diffusion models are bottlenecked by their slow sampling speed due to the large number of evaluation steps. The Denoising Diffusion Implicit Model (DDIM) \cite{song2021denoising} extends the original DDPM to non-Markovian cases, enabling faster generation with fewer denoising steps. DPM-solver \cite{lu2022dpm} exploits the semi-linear structure of probability flow ODE to develop a more efficient ODE solver. Progressive Distillation \cite{salimans2022progressive} suggests distilling the full sampling process into a faster sampler parameterized as a neural network.

Inspired by the theory of the continuous-time diffusion model, the Consistency Model \cite{song2023consistency} supports single-step generation that obtains consistent images from arbitrary sampling points that belong to the same PF ODE trajectory, enforcing the proposed self-consistency property. Consistency models can be trained in either the distillation mode or the isolation mode. Moreover, it still allows iterative generation for zero-shot data editing and trade-offs between sample quality and computing.

\subsection{Adversarial Attack}
Numerous studies have been explored to attack deep models by introducing imperceptible perturbations into input data \cite{Goodfellow15Explaining,madry2017towards,croce2020reliable,athalye2018obfuscated}. Depending on the attacker's level of access to the model, adversarial attacks are categorized into white-box attacks and black-box attacks. White-box attackers have complete access to the target model's structure and parameters, enabling them to induce incorrect predictions directly through gradient ascent. In contrast, black-box attackers can only control the input and output of the target model, lacking internal details. They typically employ a proxy model to generate adversarial samples, which are then applied to the target.

The phenomenon of misleading classifiers with small perturbations in images is first discovered and demonstrated by Szegedy et al. \cite{Szegedy14Intriguing}. Inspired by this, Goodfellow et al. \cite{Goodfellow15Explaining} propose the Fast Gradient Sign Method (FGSM), which introduces a single-step perturbation along the gradient direction of the loss function to generate adversarial samples. Madry et al. \cite{madry2017towards} further provide an iterative implementation of FGSM called Projected Gradient Descent (PGD), which projects the perturbed sample in each iteration to maintain similarity with the original one. While these attacks adopt $\ell_{\infty}$-norm or $\ell_{2}$-norm constraints, Papernot et al. \cite{papernot2016limitations} propose the Jacobian Saliency Map Attack (JSMA) to generate adversarial samples by constraining $\ell_{0}$-norm of the perturbation, with only a few pixel modifications required. 
Unlike traditional methods that seek a single perturbation, AutoAttack \cite{croce2020reliable}  conducts diverse and targeted attacks to explore different vulnerabilities of the model.
Additionally, Backward Pass Differentiable Approximation (BPDA) \cite{athalye2018obfuscated} computes the gradient of the non-differentiable function using a differentiable approximation, which is often an effective attack for defense methods that rely on gradient obfuscation, particularly purification methods based on stochastic gradients of the diffusion process.

 {Beyond digital perturbations, adversarial attacks can also manifest in the physical world. Prior works \cite{sharif2016accessorize,eykholt2018robust,fang2021radar,liu2022point} show that carefully crafted adversarial patterns can deceive models in real‑world scenarios. Recently, this line of research has extended to safety-critical domains like remote sensing and autonomous systems. For example, Peng et al. \cite{peng2025physical} generate physical adversarial camouflage for optical remote sensing images, deceiving models in real-world aerial scenes. Similarly, Zhang et al. \cite{zhang2024physical} craft feature-aligned expandable textures to physically attack aerial object detection systems. These underscore the urgent need for robust defense mechanisms that can protect deep learning models across diverse scenarios.}

\subsection{Adversarial Defense}
In the realm of adversarial defenses, existing approaches are primarily categorized into two main strategies: adversarial training and adversarial purification. 

\textbf{Adversarial Training.}
Adversarial training \cite{madry2017towards,pang2022robustness,zheng2020efficient,de2022make,gowal2021improving} has emerged as one of the most effective defense mechanisms against adversarial examples by incorporating adversarial perturbations during the training process. 
Huang et al. \cite{huang2015learning} define this as a min-max optimization, aiming to minimize the worst-case classification error induced by adversarial examples during training. Shaham et al. \cite{shaham2018understanding} further approach the min-max problem from the perspective of robust optimization and propose a comprehensive framework for adversarial training. Madry et al. \cite{madry2017towards} claim that iterative attacks with more refined perturbation directions facilitate more effective adversarial training. These PGD-based adversarial training method significantly improves robustness by employing PGD attack to approximate the inner maximizing loop. Building on this foundation, Zhang et al. \cite{zhang2019theoretically} and Pang et al. \cite{pang2022robustness} focus on improving the trade-off between robustness and accuracy. Zheng et al. \cite{zheng2020efficient} leverage the high transferability of models across different training epochs to enhance the efficiency and effectiveness of adversarial training.

 Some data-augmentation techniques are introduced into the process of adversarial training. For instance, Jorge et al. \cite{de2022make} introduce noise around clean samples to reinforce single-step adversarial training and mitigate the risk of catastrophic overfitting. Gowal et al. \cite{gowal2021improving} leverage samples generated by the generative model to offer more diverse augmentations, thereby improving the efficiency of adversarial training. However, these adversarial training methods often suffer from substantial performance degradation when facing attacks not encountered during training. Furthermore, the computational demands for training across various classifiers and attack types are exceptionally high. In contrast, our \mymethod, being both attack-agnostic and classifier-agnostic, offers a more flexible ``plug-and-play'' solution without requiring extensive retraining.

\textbf{Adversarial Purification.}
Adversarial purification through generative models is a preprocessing strategy designed to purify adversarial samples before classification. It typically involves training generative models to capture the underlying distribution of clean images, enabling the reconstruction of the clean versions from their adversarial counterparts. For example, Samangouei et al. \cite{samangouei2018defensegan} introduce Defense-GAN, leveraging the generative power of GANs to purify adversarial images. Song et al. \cite{song2018pixeldefend} observe that adversarial images predominantly fall in low-probability regions of the training distribution and design PixelDefend, an autoregressive generative model, to purify adversarial images. Similarly, Srinivasan et al. \cite{srinivasan2021robustifying} drive off-manifold adversarial samples towards high-density regions of the data generating distribution by the Metropolis adjusted Langevin algorithm (MALA) \cite{roberts1998optimal}. Moreover, studies such as \cite{du2019implicit,Grathwohl2020Your,hill2021stochastic} demonstrate that energy-based models (EBMs) are also effective for adversarial purification.

Benefiting from the superiority of the diffusion-based generative models, there's a growing interest in their application for adversarial purification. Yoon et al. \cite{yoon2021adversarial} achieve purification by gradually denoising the adversarial images, with the stopping threshold determined through score matching to avoid over-purification. DiffPure \cite{nie2022diffusion} adopt a different approach by first adding noise to adversarial samples and then restoring clean images via a reverse denoising process. Similarly, GDMP \cite{wang2022guided} facilitate the generation of clean samples by introducing guidance information into the reverse denoising phase. However, these methods continue to face challenges in balancing the incomplete removal of adversarial noise with the precise restoration of structural textures, as well as addressing the inconsistency in noise levels across different attack types.

%-------------------------------------------------------------------------
\section{Preliminaries}
\subsection{Adversarial Attack}
The typical goal of adversarial attack is to mislead the target classifier by crafting a sample with imperceptible perturbation \cite{madry2017towards, zhang2019theoretically}, whose general process can be formulated as: 

\begin{deftn}
(Adversarial Sample)
Let $\mathcal{D}=\{(\mathbf{x}_{i}, y_{i})\}_{i=1}^{n}$ denote a collection of $\mathbf{x}_{i}$ from the input space $\mathcal{X} \subset \mathbb{R}^{d}$ and $y_{i}$ as its ground-truth label defined in a label set $\mathcal{C}=\{1, \ldots, C\}$, and $\hat{h}$ be a well-trained classifier on $\mathcal{D}$. An adversarial sample $\hat{\mathbf{x}}$ regarding $\mathbf{x}$ with perturbation $\epsilon$ is generated as:
\begin{equation}
    \hat{\mathbf{x}} = \mathop{\arg\max}\limits_{\tilde{\mathbf{x}} \in \mathcal{B}(\mathbf{x}, {\epsilon})} \mathcal{L} \left(\hat{h}(\hat{\mathbf{x}}), \by\right),
\end{equation}
where $\mathcal{B}(\mathbf{x}, {\epsilon})=\left\{\mathbf{x}^{\prime} \in \mathcal{X} \mid d\left(\mathbf{x}, \mathbf{x}^{\prime}\right) \leq \epsilon\right\}$, $d$ is some distance (\eg, $\ell_2$ or $\ell_\infty$ distance), and $\mathcal{L}$ is some loss function. For simplicity, we denote $\hat{\mathbf{x}} = \mathbf{x} + \boldsymbol{\epsilon}_a$ as the adversarial sample with the adversarial perturbation $\boldsymbol{\epsilon}_a$. 
\end{deftn}

\subsection{Adversarial Purification}
Adversarial purification aims to use a generative model to restore the clean sample from the adversarial sample
\cite{samangouei2018defense,yoon2021adversarial,nie2022diffusion,lee2023robust,pei2025diffusion}, which is defined as below.

\begin{problem}
(Adversarial Purification)
    Let $\mathcal{X} \subset \mathbb{R}^{d}$ be a separable metric space and $p$ be a Borel probability measure on $\mathcal{X}$. Given IID samples $\mathcal{D}_{p}=\{\mathbf{x}^{(i)}\}_{i=1}^{n}$ from the distribution $p$ and  a ground-truth labeling mapping $h:\mathbb{R}^{d} \to \mathcal{C}$ with $\mathcal{C}=\{1, \ldots, C\}$ being a label set. Assuming that the attacker has access to some well-trained classifier $\hat{h}$ and generates samples $\mathcal{D}^{\prime}=\{\hat{\mathbf{x}}^{(i)}\}_{i=1}^{m}$ that mislead the classifier, we wish to restore samples $\hat{\mathbf{x}}_{i}$ in $\mathcal{D}^{\prime}$ back to the corresponding $\mathbf{x}_{i}$ in $\mathcal{D}_{p}$.
\end{problem}

\subsection{Continuous-time Diffusion Models}
Diffusion models, also referred to as score-based generative models \cite{songscore}, sequentially corrupt input data with slowly increasing noise, and then learn to reverse this corruption to form a generation process.

Given a data distribution $p(\mathbf{x})$, diffusion models initiate a forward diffusion process $\{\mathbf{x}_{t}\}_{t=0}^{T}$ indexed by continuous time $t \in [0, T]$, such that $\mathbf{x}_{0} := \mathbf{x} \sim p(\mathbf{x})$ represents  the data distribution, and $\mathbf{x}_{T} \sim p_{T}(\mathbf{x})$ is the prior distribution. This diffusion process can be modeled by a \textbf{stochastic differential equation (SDE)} with positive
time increments:
\begin{equation}
\label{forward_SDE}
    \mathrm{d} \mathbf{x} = \mathbf{f}(\mathbf{x}, t) \mathrm{d} t + g(t) \mathrm{d} \mathbf{w},
\end{equation}
where $\mathbf{f}(\cdot, t): \mathbb{R}^{d} \to \mathbb{R}^{d}$ is a vector-valued function called drift coefficient, $g(\cdot): \mathbb{R} \to \mathbb{R}$ is a scalar function called diffusion coefficient, and $\mathbf{w}$ is a standard Wiener process.

Denote by $p_{t}(\mathbf{x})$ the marginal distribution of $\mathbf{x}_{t}$ with $p_{0}(\mathbf{x}) := p(\mathbf{x})$. By starting from $\mathbf{x}_{T}$ and reversing the diffusion process, wherein the reverse of a diffusion is also a diffusion \cite{anderson1982reverse}, we can reconstruct samples $\mathbf{x}_{0} \sim p_{0}(\mathbf{x})$. The reverse process is given by the reverse-time SDE:
\begin{equation}
\label{reverse_SDE}
    \mathrm{d} \mathbf{x} = [\mathbf{f}(\mathbf{x}, t) - g(t)^{2} \nabla_{\mathbf{x}}\log p_{t}(\mathbf{x})] \mathrm{d} t + g(t) \mathrm{d} \bar{\mathbf{w}},
\end{equation}
where $\bar{\mathbf{w}}$ is a standard Wiener process when time flows backwards from $T$ to $0$, $\mathrm{d} t$ is an infinitesimal negative timestep.

Given the inherent denoising ability and randomness of diffusion models, many studies \cite{yoon2021adversarial,nie2022diffusion,lee2023robust,pei2025diffusion} use them for adversarial purification by adding noise at a specific timestep to the test sample during the forward diffusion process via Eqn. (\ref{forward_SDE}) and gradually removing it through reverse denoising steps via Eqn. (\ref{reverse_SDE}). However, if the noise level is too low, it fails to cover adversarial noise, resulting in poor defense performance; if too high, it significantly reduces natural accuracy. Additionally, different attacks require varying noise levels, making it difficult to defense diverse adversarial attacks \citet{nie2022diffusion,lee2023robust}.

\begin{figure*}[th]
    \begin{center}
    \includegraphics[width=0.98\textwidth]{ 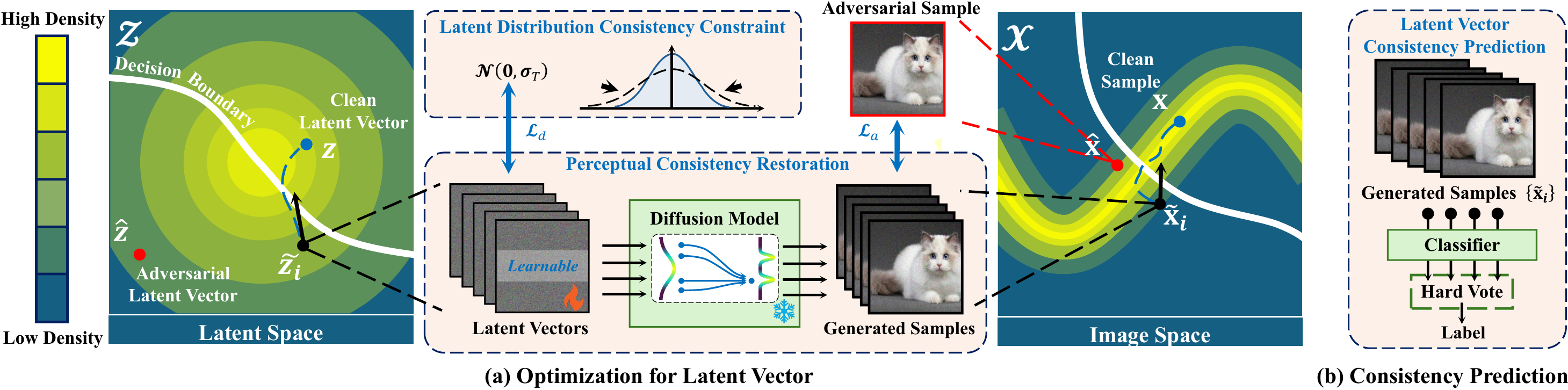}
    \vspace{-7pt}
    \caption{ {Overview of the proposed \mymethod. \textbf{(a)} Given a pre-trained consistency (diffusion) model, we optimize a set of latent vectors $\{\tilde{\mathbf{z}}_i\}$ in its latent space $\mathcal{Z}$ to generate samples $\{\tilde{\mathbf{x}}_i\}$ as close to the original test sample $\hat{\mathbf{x}}$ while removing potential adversarial perturbations by \textit{perceptual consistency restoration} mechanism and \textit{latent distribution consistency constrain} strategy, illustrated here for $\hat{\mathbf{x}}$ as an adversarial sample. The perceptual consistency restoration mechanism employs $\mathcal{L}_a$ consisting of MAE and SSIM losses to align generated samples with the clean data manifold, while the latent distribution consistency constraint strategy uses a Gaussian distribution constraint loss $\mathcal{L}_{d}$ to ensure that the optimized vectors $\{\tilde{\mathbf{z}}_i\}$ stay within the valid manifold. \textbf{(b)} After optimization, we employ a \textit{latent vector consistency prediction} scheme by a label voting across the final generated images to determine the final prediction for the test sample $\hat{\mathbf{x}}$.}}
    \label{fig: overview}
    \vspace{-12pt}
    \end{center}
\end{figure*}

\subsection{Consistency Model}
\label{sec: consistency}
The consistency model \cite{song2023consistency} is a new type of generative model that supports both single-step generation and multi-step generation for trade-offs between quality and computing. Its core design involves mapping each sampling point on an ODE-based diffusion trajectory to its origin by learning a consistency model $f_{\theta}:(\mathbf{x}_{t},t) \mapsto \mathbf{x}_{\delta}$, where $\delta$ is a fixed positive number close to $0$. Formally, the consistency model $f_{\theta}$ is required to satisfy the following self-consistency property:
\begin{equation}
    f_{\theta} (\mathbf{x}_{t},t) = f_{\theta} (\mathbf{x}_{t^{\prime}},t^{\prime}),  \forall t, t^{\prime} \in [\delta, T].
\end{equation}
Following \cite{song2023consistency}, to ensure that $f_{\theta} (\mathbf{x}_{\delta},\delta) = \mathbf{x}$, the consistency model $f_{\theta}$ is parameterized as:
\begin{equation}
    f_{\theta}(\mathbf{x}_{t},t) = c_{\text{skip}}(t)\mathbf{x} + c_{\text{out}}(t)F_{\theta}(\mathbf{x}_{t},t),
\end{equation}
where $c_{\text{skip}}(t)$ and $c_{\text{out}}(t)$ are differentiable functions with $c_{\text{skip}}(\delta){=}1$ and $c_{\text{out}}(\delta){=}0$, and $F_{\theta}(\mathbf{x}_{t},t)$ is a deep neural network. Throughout the paper, we omit the timestep $T$ and use $f_\theta(\cdot)$ to denote the generated sample for simplicity.

A consistency model can be either distilled from a well-trained diffusion model or directly trained from scratch, known as consistency distillation and consistency training \cite{song2023consistency}. 

\textbf{Consistency distillation} enforces the self-consistency property by defining a consistency distillation loss as follows:
\begin{equation}
\begin{split}
    \mathcal{L}_{\text{CD}}(\theta, \theta^{-}; \phi) = \mathbb{E}_{\mathbf{x},t}[d(f_{\theta}(\mathbf{x}_{t_{n+1}},t_{n+1}),f_{\theta^{-}}(\hat{\mathbf{x}}_{t_{n}}^{\phi},t_{n}))],
    \\
    \hat{\mathbf{x}}_{t_{n}}^{\phi} = \mathbf{x}_{t_{n+1}} + (t_{n}-t{n+1}) \Phi (\mathbf{x}_{t_{n+1}}, t_{n+1}, \phi),
\end{split}    
\end{equation}
where $\theta^{-}$ is a target model updated via the exponential moving average (EMA) of the parameter $\theta$, $\hat{\mathbf{x}}_{t_{n}}^{\phi}$ denotes a one-step estimation of $\mathbf{x}_{t_{n}}$ of $\mathbf{x}_{t_{n+1}}$ with $\Phi$ as the one-step ODE solver applied to PF ODE, and $d$ is a distance between two samples. When using the Euler solver, we have $\Phi (\mathbf{x}, t, \phi) = -t s_{\phi}(\mathbf{x}, t)$, with $s_{\phi}(\mathbf{x}, t)$ refers to the score model.

\textbf{Consistency training} approximates the score function $\nabla_{\mathbf{x}}\log p_{t}(\mathbf{x})$ with the following unbiased estimator, therefore avoid the pre-trained score model $s_{\phi}(\mathbf{x},t)$ altogether:
\begin{equation}
    \nabla_{\mathbf{x}}\log p_{t}(\mathbf{x}) = -\mathbb{E}[\frac{\mathbf{x}_{t}-\mathbf{x}}{t^{2}} \mid  \mathbf{x}_{t}],
\end{equation}
and similar to consistency distillation, the consistency training loss is obtained as follows:
\begin{equation}
    \mathcal{L}_{\text{CT}}(\theta, \theta^{-}) = \mathbb{E}_{\mathbf{x},t}[d(f_{\theta}(\mathbf{x}+{t_{n+1}}\mathbf{z},t_{n+1}),f_{\theta^{-}}(\mathbf{x}+{t_{n}}\mathbf{z},t_{n}))],
\end{equation}
where $\mathbf{z} \sim \mathcal{N}(\mathbf{0}, \mathbf{I})$. Moreover, it can be proven that $\mathcal{L}_{\text{CD}}(\theta, \theta^{-}; \phi) = \mathcal{L}_{\text{CT}}(\theta, \theta^{-}) + o(\Delta  t)$.

\textbf{Advantages of consistency models for adversarial purification}: 1) \textit{The ODE-based generation follows a deterministic trajectory}. Different from previous diffusion models \cite{songscore,ho2020denoising}, which may generate different samples with a latent vector from an Gaussian distribution, the ODE-based consistency model generates a unique sample from a latent vector, ensuring a \textit{deterministic} latent vector for each sample. This lays an important foundation for iterative optimization in finding the latent vector corresponding to a sample on the data manifold.  2) \textit{The consistency model enables one-step generation}. Unlike other diffusion models \cite{songscore,ho2020denoising} that require multiple iterations to generate high-quality images, the consistency model can produce a reasonably good image in a single step, facilitating efficient iterative latent vector optimization.

%-------------------------------------------------------------------------
\section{Proposed Methods}
\label{sec: methods}
Existing diffusion-based purification methods \cite{yoon2021adversarial,nie2022diffusion,lee2023robust,pei2025diffusion} inject Gaussian noise into the test sample and then recover them to purified samples through the denoising process. 
However, these methods suffer from incomplete removal of adversarial noise due to the inherent property of the diffusion and denoising processes, and inconsistent noise levels across different attacks.
To address this, we aim to break away from this traditional paradigm by focusing on the clean data manifold.

Motivated by the observation that samples generated by a well-trained generative model are close to clean ones but distant from adversarial ones (see Fig. \ref{fig: motivition} and Sec. \ref{sec:Observations}), we aim to optimize latent vectors within the latent space of a generative model to generate samples to restore clean data. Leveraging the consistency model with deterministic generation and efficient one-step image synthesis, we propose \textbf{Consistency Model-based Adversarial Purification} (\mymethod). Particularly, we introduce a \textit{perceptual consistency restoration} mechanism to optimize latent vectors to align generated samples with the data manifold (Sec. \ref{sec: Perceptual Consistency Restoration}), a \textit{latent distribution consistency constraint} strategy to maintain the optimized vectors within the valid manifold (Sec. \ref{sec: Latent Distribution Consistency Constraint}), and a \textit{latent vector consistency prediction} scheme to stabilize the final output by aggregating the results of the multiple optimized vectors (Sec. \ref{sec: Latent Vector Consistency Prediction}). The framework of \mymethod~is illustrated in Fig. \ref{fig: overview}, with its algorithm in Alg. \ref{alg: odepure with aug}. Last, we provide a consistency-disruption attack tailored to \mymethod~in Sec. \ref{sec: Adtaptive Attack} and its algorithm in Alg. \ref{alg: antiodepure}. 

Given a test sample $\hat{\mathbf{x}}$, the overall optimization for latent vectors $\tilde{\bz}=\{\tilde{\mathbf{z}}_i\}_{i=1}^K$ in our \mymethod~is formulated as:
\begin{equation}
\label{eq: overview}
    \min_{\tilde{\bz}} \mathcal{L}_{a}(\tilde{\bz}, \hat{\mathbf{x}})+\beta \mathcal{L}_{d}(\tilde{\bz}), 
\end{equation}
where $\mathcal{L}_{a}$ is the perceptual consistency restoration loss defined in Eqn.~(\ref{eqn: Align_Loss}), $\mathcal{L}_{d}$ is the latent distribution consistency constraint in Eqn.~(\ref{eqn: Dist_Loss}), and $\beta$ is a hyper-parameter.

\begin{algorithm}[t]
\caption{Adversarial purification with \mymethod.}\label{alg: odepure with aug}
\begin{algorithmic}
\STATE 
\STATE \textbf{Input:}  {Test samples $\hat{\bx}$, the consistency model $f_\theta$, the latent distribution $\mathcal{N}(\mathbf{0}, \sigma_T\mathbf{I})$, the number of sampling $K$, the total optimization iterations $T$, the step size $\eta$, the classifier $\hat{h}$}.

\STATE \textbf{Sample} $\tilde{\bz}^{0}=\{\tilde{\mathbf{z}}_i\}_{i=1}^K \sim \mathcal{N}(\mathbf{0}, \sigma_T\mathbf{I})$.

\FOR{$j = 1, \ldots, T$}

\STATE // \textit{Perceptual consistency restoration}. 
\STATE Obtain the  restoration loss  $\mathcal{L}_{a}(\tilde{\bz}^{j-1}, \hat{\mathbf{x}})$ using Eqn. (\ref{eqn: Align_Loss}).
\STATE // \textit{Latent distribution consistency
constraint}.
\STATE Obtain the latent distribution constraint loss $\mathcal{L}_{d}(\tilde{\bz}^{j-1})$ using Eqn. (\ref{eqn: Dist_Loss}).
\STATE Update the latent vectors by gradient descent:
\STATE ~~~~~~~$\tilde{\bz}^{j} \gets \tilde{\bz}^{j-1} - \eta \cdot \frac{ \partial \left[\mathcal{L}_{a}(\tilde{\bz}^{j-1}, \hat{\mathbf{x}})+\beta \mathcal{L}_{d}(\tilde{\bz}^{j-1})\right]}{ \partial {\bz}_{j-1}}$. 

\ENDFOR
\STATE // \textit{Latent vector consistency
prediction}.
\STATE Obtain the generated samples $\tilde{\mathbf{x}} \gets f_{\theta}(\tilde{\bz}^T)$.
\STATE Obtain the prediction $\hat{y}$ via the label voting using Eqn. (\ref{eqn: hard label voting}).

\STATE \textbf{Output:} $\hat{y}$.
\end{algorithmic}
\end{algorithm}

\subsection{Perceptual Consistency Restoration}
\label{sec: Perceptual Consistency Restoration}
Previous adversarial purification methods \cite{yoon2021adversarial,nie2022diffusion,lee2023robust,pei2025diffusion} typically restore the original data by directly modifying the test sample. However, this approach inherently struggles to fully eliminate adversarial perturbations, as it addresses the processed data rather than targeting the underlying causes of the perturbation. To address this, we propose a \textit{perceptual consistency restoration} mechanism focusing on the clean data manifold, which optimizes latent vectors within a generative model's latent space to generate samples resembling the test sample, thus removing potential adversarial perturbations at their source.

One straightforward approach to achieve the above goal is to optimize GAN latent vectors, which, however, often produces semantically unclear or structurally blurred images (see Fig. 9 in Appendix), leading to inferior defense performance shown in \cite{nie2022diffusion}. A possible reason is that the latent space, constrained by a low-dimensional normal distribution, limits its capacity for representation and semantic disentanglement \cite{xia2022gan,roich2022pivotal}. To overcome this, benefiting from the high-quality samples generated by editing latent vectors in diffusion models \cite{rombach2022high,avrahami2023blended,schramowski2023safe,xu2023geometric}, we can exploit a diffusion model to obtain an optimized latent vector. Unfortunately, one obvious limitation of diffusion models is their reliance on multi-step iterative processes to generate a single image, making them less suitable here. To circumvent this issue, we turn to consistency models \cite{song2023consistency}, which offer the advantages of deterministic generation and efficient one-step image synthesis (see more details in Sec. \ref{sec: consistency}), making it well-suited for this task.  {We defer more discussions of different generative models in Sec. \ref{sec: Ablation}}.

Formally, given a pre-trained consistency model $f_{\theta}$, we sample a set of latent vectors $\{\tilde{\bz}_i\}_{i=1}^K {\sim} \mathcal{N}(\mathbf{0}, \sigma_{T}\mathbf{I}) $ from its latent distribution. We employ multiple latent vectors for two reasons: 1) Multiple latent vectors reduce the risk of selecting an outlier deviating too far from the Gaussian distribution center, ensuring more reliable optimization results; 2) Multiple latent vectors provide a basis for imposing constraints on the distribution of latent vectors during optimization (see Sec. \ref{sec: Latent Distribution Consistency Constraint}). To restore the test sample, intuited that the adversarial samples typically reside near the manifold of clean data in the pixel space \cite{wang2022guided,prakash2018deflecting,yuan2019adversarial}, we employ Mean Absolute Error (MAE \cite{willmott2005advantages}) and Structure Similarity Index Measure (SSIM \cite{brunet2011mathematical}) to align the samples generated by the optimized latent vectors with the test sample, which is formulated as

\begin{align}
\label{eqn: Align_Loss}
    \mathcal{L}_{a}(\tilde{\bz}, \hat{\mathbf{x}}) {=} 
\frac{1}{K}\sum_{i=1}^K \left\| f_{\theta}(\tilde{\mathbf{z}}_i){-} \hat{\mathbf{x}}\right\|_1 {-} \alpha \cdot \mathrm{SSIM}(f_{\theta}(\tilde{\mathbf{z}}_i), \hat{\mathbf{x}}). 
\end{align}
Here, we omit the time $T$ in $f_\theta(\cdot)$ for simplicity. MAE focuses on preserving overall color and brightness, ensuring the sample aligns with the original test sample's appearance on the manifold. Meanwhile, SSIM captures high-frequency textures and structural details, critical for representing the local structure of the manifold. Furthermore, our experiments show that SSIM aids in faster convergence, enabling the model to rapidly achieve a high-quality restoration that represents both pixel-level and perceptual consistency with the original sample.

\begin{remark}
    Note that traditional white-box attacks \cite{Goodfellow15Explaining, madry2017towards, croce2020reliable} typically rely on manipulating the leaf nodes of the computational graph to generate adversarial samples that can bypass purification methods. However, in our approach, we employ the Gaussian vector to generate samples, treating them as non-leaf nodes in the optimization process. This makes it difficult for gradient-based white-box attacks to effectively target our defense module, as adversarial perturbations cannot directly influence the latent space optimization.
\end{remark}

\subsection{Latent Distribution Consistency Constraint}
\label{sec: Latent Distribution Consistency Constraint}
The perceptual consistency restoration mechanism can achieve moderate performance in both robust accuracy and clean accuracy. However, it heavily depends on selecting an appropriate number of optimization iterations (see Fig. \ref{fig: ablation_beta}). Excessive iterations may cause the latent vector to fit the adversarial noises embedded in the input sample, making the generated sample increasingly similar to the adversarial input at the pixel level, until they become exactly the same.  This issue occurs because the optimization may unintentionally steer the latent vector away from the clean data manifold when aligning with the adversarial input \cite{dombrowski2023diffeomorphic}. To further clarify this, 
we derive the following theorem to provide insight into the distribution discrepancy between the latent vectors of clean and adversarial samples when the natural data follow a Gaussian distribution.

\begin{thm}
\label{thm: adv vs natural}
    Assuming that the distribution of natural data $p(\bx){=} \mN({\boldsymbol{\mu}}_{\bx}, \sigma_{\bx}^2\mathbf{I})$,  {where $\mathbf{I}$ is an identity matrix}, given a PF ODE sampling $\mathrm{d}\bx=-t \nabla_{\mathbf{x}}\log p_{t}(\mathbf{x})$ with $\mathbf{f}(\bx,t)=\bf{0}$ and $g(t)=\sqrt{2t}$, then for $\forall~\bx \in p(\bx)$ and its adversarial sample $\hat{\bx}=\bx+\boldsymbol{\epsilon}_a$, we have
    \begin{equation}
         \bx_T -\hat{\bx}_T \sim \mN( \mathbf{0}, 2\sigma_{cl}^2\mathbf{I})+\boldsymbol{\mu}_{\epsilon},
    \end{equation}
    where $\boldsymbol{\mu}_\epsilon=\left(\mathbb{E}_t  \frac{ t}{\sigma_{\bx}^2+ t^2}-1\right)\boldsymbol{\epsilon}_a$ and $\sigma_{cl}^2=\mathbb{E}_t  \frac{ t^2}{\sigma_{\bx}^2+ t^2}$.
\begin{proof}
    The proof is provided in Appendix VII.1.
\end{proof}
\end{thm}
Theorem \ref{thm: adv vs natural} indicates that when tracing the ODE trajectory to find the corresponding latent vector for an adversarial sample, a significant distribution shift occurs between this latent vector and that of the original clean sample due to the term $\boldsymbol{\mu}_{\epsilon}$. Although the mean term can theoretically be zero (see Appendix VII.1), this scenario is highly unlikely in practical applications, where natural data distributions are typically diverse. Consequently, it is essential to ensure that the optimized latent vectors remain within a valid and meaningful region of the latent space, thereby maintaining the alignment of the restored image with the clean distribution and avoiding overfitting adversarial perturbations.

To achieve the above goal, we propose a \textit{latent distribution consistency constraint} strategy to enforce the optimized latent vectors to stay close to the latent distribution of the pre-trained consistency model. Constraining a single sample to belong to a specific distribution may be a challenging task. Fortunately, the perceptual consistency restoration mechanism depicted in Sec. \ref{sec: Perceptual Consistency Restoration} has already provided multiple optimized latent vectors corresponding to a single test input, which enables various techniques of the distribution alignment on the optimized vectors,such as Wasserstein distance \cite{villani2009optimal} and maximum mean discrepancy \cite{gretton2012kernel,borgwardt2006integrating}. 

In this work, we employ an MSE loss to align the mean and variance of the multiple optimized latent vectors with the latent distribution of the consistency model since it has been widely used as a metric to assess distribution alignment \cite{qin2023diverse,yin2020dreaming}. Formally, the latent distribution consistency constraint loss is provided as follows:
\begin{equation}
\label{eqn: Dist_Loss}
    \mathcal{L}_{d}(\tilde{\bz}) = \left \| \mathbf{\mu}(\{\tilde{\mathbf{z}}_i\}) - \mathbf{\mu}_{\mathbf{z}} \right\|_2^2+ \left\|\sigma(\{\tilde{\mathbf{z}}_i\}) - \sigma_{\mathbf{z}} \right\|_2^2,
\end{equation}
where $\mathbf{\mu}(\{\tilde{\mathbf{z}}_i\}){=}\frac{1}{K} \sum_{i=1}^K \tilde{\mathbf{z}}_i$, $\sigma^2(\{\tilde{\mathbf{z}}_i\}){=}\frac{1}{K} \sum_{i=1}^K (\tilde{\mathbf{z}}_i {-}\mathbf{\mu}(\{\tilde{\mathbf{z}}_i\})^2$ are the empirical mean and variance of the optimized vectors, respectively, $\mathbf{\mu}_{\mathbf{z}}=\mathbf{0}, \sigma^2_{\mathbf{z}}=\sigma_{T}\mathbf{I}$ are the mean and variance of the latent distribution w.r.t the pre-trained consistency model. 

MSE is computationally straightforward and can be particularly effective here. It directly aligns the first and second moments (\eg, mean and variance) of the latent vectors with the latent distribution. This approach effectively constrains the optimized latent vectors to maintain a statistical consistency to the latent  distribution, preserving the underlying structure of the clean data manifold.  Other techniques of the  distribution alignment can be explored in future work. This strategy helps maintain the alignment of the generated samples with the clean data, effectively preventing the optimized latent vectors from drifting toward adversarial artifacts. 

Thus far, we have detailed the optimization process in our \mymethod. In fact, optimizing the objective function in Eqn. (\ref{eq: overview}) implicitly minimizes an upper bound on the reconstruction loss for the \textit{clean} sample, as shown in the following proposition.
\begin{prop}
\label{prop: upper_bound}
Given a test sample $\hat{\bx} $, it holds that optimizing a set of latent vectors $\{\tilde{\bz}_i\}_{i=1}^K$ by Eqn. (\ref{eq: overview}) gives an upper bound on the reconstruction for the clean sample $\bx$:
\begin{equation}
\label{eq: upper_bound}
    \frac{1}{K} \sum_{i=1}^K\| f_{\theta}(\tilde{\mathbf{z}}_i) - \mathbf{x} \|_1 +\beta \mathcal{L}_{d}(\tilde{\bz}) \leq C + \mathcal{L}_{a}(\tilde{\bz}, \hat{\mathbf{x}})+\beta \mathcal{L}_{d}(\tilde{\bz}), 
\end{equation}
where $C$ is a constant related to the adversarial perturbation.
\begin{proof}
    The proof is provided in Appendix VII.2.
\end{proof}
\end{prop}

The conclusion in Proposition (\ref{prop: upper_bound}) is evident. When the test sample $\hat{\bx}$ is clean, minimizing $\mathcal{L}_{a}(\tilde{\bz}, \hat{\mathbf{x}}) + \beta \mathcal{L}_{d}(\tilde{\bz})$ directly reduces the loss between the generated and original clean samples; when $\hat{\bx}$ is adversarial, this objective effectively tightens the upper bound on the reconstruction loss of the original clean sample. This suggests that our \mymethod~method not only aims to restore the clean data but also constrains adversarial perturbations, preventing significant reconstruction errors associated with the clean sample. The latent distribution consistency term $\mathcal{L}_{d}(\tilde{\bz})$ further aligns the optimized latent vectors with the expected clean distribution, thereby reducing the likelihood of the reconstruction process yielding adversarial outputs. Consequently, the overall objective provides a strong regularization effect, pushing the optimization towards regions consistent with clean data, thus ensuring a smaller gap between the reconstructed sample and the original clean sample.

\subsection{Latent Vector Consistency Prediction}
\label{sec: Latent Vector Consistency Prediction}

While the perceptual consistency restoration mechanism and latent distribution consistency constraint strategy ensure that the generated samples are aligned with the clean data manifold and remain within a meaningful region of the latent space, there is still the potential for variability among the multiple optimized latent vectors. This variability can introduce inconsistencies in the generated samples, which may lead to fluctuations in the final classification results. To address this, we introduce the \textit{latent vector consistency prediction} scheme to stabilize the output and enhance the robustness of our purification method.

In this scheme, we aggregate the predictions from multiple generated samples to produce a more reliable classification. Specifically, given a set of optimized latent vectors $\{\tilde{\bz}_i\}_{i=1}^K$, we generate corresponding samples $\{\tilde{\bx}_i\}_{i=1}^K$ with $\tilde{\bx}_i=f_\theta(\tilde{\bz}_i)$ and obtain their predicted labels $\{\hat{h}(\tilde{\bx}_i)\}_{i=1}^K$ using the pre-trained classifier $\hat{h}$. The final prediction $\hat{y}$ is then determined by a label voting mechanism, which selects the label with the highest vote count across all predictions: 
\begin{align}
\label{eqn: hard label voting}
    \hat{y} = \arg\max_{y} \sum_{i=1}^K \mathbb{I}\left[\hat{h}(f_{\theta}( {\tilde{\bz}}_i)) = y\right],
\end{align}
where $\mathbb{I}[\cdot]$ is an indicator function that returns 1 if the condition is true and 0 otherwise.

This label voting mechanism effectively reduces the impact of any individual outlier prediction, thereby stabilizing the overall decision-making process. By aggregating predictions from multiple latent vectors, the method leverages the collective information from diverse perspectives within the latent space, leading to a more consistent and robust final output.

\textbf{Advantages of \mymethod~for adversarial purification}: 1) \textit{Shifted attack space}: Traditional adversarial attacks focus on generating perturbations in the input space, but our method operates in the latent space of a generative model. This \textit{shift} forces attackers to generate perturbations in the latent space, which is unfamiliar and challenging for them to exploit effectively, thereby complicating the attack process. 2) \textit{Latent distribution regularization:} Our method enforces regularization on the latent vectors, keeping them near a Gaussian distribution centered on clean data. This safeguard makes it difficult for adversarial perturbations to push the generated samples away from the clean data manifold. 3) \textit{Stability of multiple samplings}: Our approach introduces diversity in the generated outputs by sampling multiple latent vectors. This further increases robustness because the attacker would need to successfully perturb all sampled latent vectors simultaneously to achieve a consistent attack, significantly raising the difficulty.

\begin{algorithm}[t]
\caption{Consistency-Disruption Attack against \mymethod.}\label{alg: antiodepure}
\begin{algorithmic}
\STATE 
\STATE \textbf{Input:}  {Clean samples $\mathbf{x}$ and labels $y$, the consistency model $f_\theta(\cdot)$, 
the number of alignment iterations $T_\mathrm{def}$, the number of attack iterations $T_\mathrm{adv}$, the attack intensity $\epsilon$, the adversarial factor $\lambda$, the attack step size $\eta'$, the classifier $\hat{h}$.}\\

\vspace{2pt}
\STATE  {Obtain initialized latent vectors $\tilde{\bz}^{T_\mathrm{def}}$ via Alg. \ref{alg: odepure with aug} $(T{=}T_\mathrm{def})$.} 

\FOR{$j = T_\mathrm{def} + 1, \ldots, T_\mathrm{def} + T_\mathrm{adv}$}
\STATE Obtain the generated samples $\mathbf{x}^{j-1} \gets f_\theta(\tilde{\bz}^{j-1})$.

\STATE Project the samples into $\mathcal{B}(\mathbf{x}, \epsilon)$:
\STATE ~~~~~~~~$\mathbf{x}_\mathrm{adv} \gets Proj_{\mathcal{B}(\mathbf{x}, \epsilon)}(\mathbf{x}^{j-1})$.

\STATE Update the latent vectors by gradient descent:
\STATE ~$\tilde{\bz}^{j} \gets \tilde{\bz}^{j-1} {-} \eta' \cdot \frac{ \partial \left[\mathcal{L}_{a}(\tilde{\bz}^{j-1}, \hat{\mathbf{x}})+\beta \mathcal{L}_{d}(\tilde{\bz}^{j-1})- \lambda \cdot \mathcal{L}_{CE}(\mathbf{x}_\mathrm{adv},y) \right]}{ \partial {\bz}_{j-1}}$.
\ENDFOR

\STATE Obtain the adversarial samples $\mathbf{x}_\mathrm{adv} \gets f_\theta(\tilde{\bz}^{T_\mathrm{def} + T_\mathrm{adv}})$.
\STATE Project the samples into $\mathcal{B}(\mathbf{x}, \epsilon)$: $\mathbf{x}_\mathrm{adv} \gets Proj_{\mathcal{B}({\bx}, \epsilon)}(\mathbf{x}_\mathrm{adv})$.

\STATE \textbf{Output:} $\mathbf{x}_\mathrm{adv}$.
\end{algorithmic}
\end{algorithm}

\subsection{Consistency-Disruption Attack against \mymethod}
\label{sec: Adtaptive Attack}
Given that traditional white-box attacks \cite{Goodfellow15Explaining, madry2017towards, croce2020reliable} are ineffective against our method, we recognize the importance of an attack strategy specifically tailored to our defense mechanism to rigorously evaluate the robustness of our \mymethod. 
This attack should be aware of the defense mechanism and seek to exploit any potential vulnerabilities. Under such circumstances, unlike traditional attacks that optimize perturbations in the input space, we propose a \textit{consistency-disruption attack}, which 
navigates the latent space of a generative model to optimize the latent vectors during the purification process, steering generated samples away from the clean data manifold and toward adversarial regions.

To achieve the above goal, we modify the objective function in Eqn. (\ref{eq: overview}) of our defense method by introducing an additional term that maximizes the cross-entropy loss between the prediction of the classifier and the true label of the sample. This attack can be formulated as follows:
\begin{align}
\label{eq: overview_attack}
    \min_{\tilde{\bz}} \mathcal{L}_{a}(\tilde{\bz}, {\mathbf{x}})+\beta \mathcal{L}_{d}(\tilde{\bz})-\lambda \mathcal{L}_{CE}(\bx_\mathrm{adv},y), \\
    \st,~~\bx_\mathrm{adv}=Proj_{\mathcal{B}(\mathbf{x}, \epsilon)}(f_\theta(\tilde{\mathbf{z}})), \nonumber
\end{align}
where  $\mathcal{L}_{CE}$ is the cross-entropy loss, $Proj_{\mathcal{B}(\mathbf{x}, \epsilon)}(\cdot)$ projects the adversarial sample into a norm-ball. The hyper-parameter $\lambda$ controls the strength of the adversarial attack.

To efficiently optimize the attack objective in Eqn. (\ref{eq: overview_attack}), we initialize the latent vectors $\tilde{\bz}$ by leveraging our defense mechanism (without the loss term $\mathcal{L}_\mathrm{CE}$) to find the optimized latent vectors corresponding to the clean sample. Starting from this initialization helps in maintaining closer proximity to the clean data manifold and then gradually introducing adversarial perturbations in the latent space by adding additional $\mathcal{L}_\mathrm{CE}$ loss. We provide a detailed attack procedure in Alg. \ref{alg: antiodepure}. Note that this attack strategy can generate multiple adversarial samples corresponding to a single clean sample. The attack is considered successful if any of these samples manage to bypass the defense mechanism and lead to misclassification.

Our experiments in Sec. \ref{sec: exper on adaptive attack} demonstrate that while this attack poses a significant challenge to defense, our \mymethod~still exhibits strong robustness primarily due to our latent distribution consistency constraint strategy and the latent vector consistency prediction scheme, which jointly maintain the alignment of the purified samples with the clean distribution, highlighting the robustness and adaptability of our approach in adversarial settings. This attack provides a rigorous test of \mymethod's robustness, ensuring that our defense mechanisms are not easily bypassed by consistency-disruption strategies.

\begin{table*}[t]
    \caption{Standard and robust accuracy against PGD+EOT and AutoAttack with $\ell_{\infty}$-norm ($\epsilon=8/255$) and $\ell_{2}$-norm ($\epsilon=0.5$)) on CIFAR-10. Adversarial Training (AT) and Adversarial Purification (AP) Strategies are evaluated. ($^*$ Model is trained with extra data.)}
    \label{tab: comparison_on_cifar}
    \begin{minipage}[c]{0.487\linewidth}
    \renewcommand\arraystretch{1.3}
    \begin{center}
    \begin{threeparttable}
    \resizebox{\linewidth}{!}{
    \begin{tabular}{lclccc}
    \toprule
    ~ & Strategy &  Method & Standard & PGD+EOT & AutoAttack \\ \midrule
    \multirow{7}{*}{\rotatebox{90}{WRN-28-10}} & \multirow{3}{*}{AT} & Gowal et al. \cite{gowal2021improving}
 & $87.51$ & \underline{$66.01$} & $63.38$ \\
    ~ & ~ & Gowal et al. \cite{gowal2020uncovering} $^*$
 & $88.54$ & $65.93$ & $62.76$ \\
    ~ & ~ & Pang et al. \cite{pang2022robustness}
 & $88.62$ & $64.95$ & $61.04$ \\ \cline{2-6}
    ~ & \multirow{4}{*}{AP} & ADP \cite{yoon2021adversarial}
 & $85.66_{\pm 0.51}$ & $33.48_{\pm 0.86}$ & $59.53_{\pm 0.87}$ \\
    ~ & ~ & DiffPure \cite{nie2022diffusion}
 & $90.27_{\pm 0.81}$ & $48.27_{\pm 1.86}$ & $64.93_{\pm 2.14}$ \\
    ~ & ~ & GNSP \cite{lee2023robust}
 & \underline{$90.40_{\pm 1.40}$} & $55.87_{\pm 0.50}$ & $70.40_{\pm 1.80}$ \\
    ~ & ~ &   {FreqPure} \cite{pei2025diffusion}
 &  $\mathbf{93.96_{{\pm}0.09}}$ &  $45.00_{{\pm}0.59}$ &  \underline{$70.84_{{\pm}0.92}$} \\
    ~ & ~ & \mymethod~(Ours)
 & $88.73_{\pm 0.50}$ & $\mathbf{74.60_{\pm 1.59}}$ & $\mathbf{78.67_{\pm 1.90}}$ \\ \midrule
    \multirow{7}{*}{\rotatebox{90}{WRN-70-16}} & \multirow{3}{*}{AT} & Rebuffi et al. \cite{rebuffi2021fixing} $^*$
 & \underline{$92.22$} & \underline{$69.97$} & $66.56$ \\
    ~ & ~ & Gowal et al. \cite{gowal2021improving}
 & $88.75$ & $69.03$ & $66.10$ \\
    ~ & ~ & Gowal et al. \cite{gowal2020uncovering} $^*$
 & $91.10$ & $68.66$ & $65.87$ \\ \cline{2-6}
    ~ & \multirow{4}{*}{AP} & ADP \cite{yoon2021adversarial}
 & $86.76_{\pm 1.15}$ & $37.11_{\pm 1.35}$ & $37.11_{\pm 1.35}$ \\
    ~ & ~ & DiffPure \cite{nie2022diffusion}
 & $90.00_{\pm 0.87}$ & $50.93_{\pm 1.30}$ & $63.87_{\pm 1.86}$ \\
    ~ & ~ & GNSP \cite{lee2023robust}
 & $90.53_{\pm 0.58}$ & $56.07_{\pm 0.83}$ & $71.67_{\pm 0.64}$ \\
     ~ & ~ &   {FreqPure} \cite{pei2025diffusion}
 &  $\mathbf{94.18_{{\pm}0.20}}$ &  $53.54_{{\pm}0.78}$ &  \underline{$74.79_{{\pm}0.53}$} \\
    ~ & ~ & \mymethod~(Ours)
 & $88.27_{\pm 0.50}$ & $\mathbf{74.80_{\pm 1.56}}$ & $\mathbf{81.33_{\pm 0.31}}$ \\
  \bottomrule
    \end{tabular}
    }
    \end{threeparttable}
    \end{center}
    \end{minipage}
    \begin{minipage}[c]{0.5\linewidth}
    \renewcommand\arraystretch{1.3}
    \begin{center}
    \begin{threeparttable}
    \resizebox{\linewidth}{!}{
    \begin{tabular}{lclccc}
    \toprule
         ~ & Strategy &  Method & Standard & PGD-$\ell_{2}$+EOT & AutoAttack-$\ell_{2}$ \\ \midrule
        \multirow{7}{*}{\rotatebox{90}{WRN-28-10}} & \multirow{3}{*}{AT} & Rebuffi et al. \cite{rebuffi2021fixing} $^*$
 & $91.79$ & $85.05$ & $78.80$ \\
         ~ & ~ & Augustin et al. \cite{augustin2020adversarial}
 & $\underline{93.96}$ & $\mathbf{86.14}$ & $78.79$ \\
         ~ & ~ & Sehwag et al. \cite{sehwag2022robust}
 & $90.93$ & $83.75$ & $77.24$ \\ \cline{2-6}
         ~ & \multirow{4}{*}{AP} &  ADP \cite{yoon2021adversarial}
 & $85.66_{\pm 0.51}$ & $73.32_{\pm 0.76}$ & $79.57_{\pm 0.38}$ \\
         ~ & ~ & DiffPure \cite{nie2022diffusion}
 & $90.27_{\pm 0.81}$ & $82.00_{\pm 0.72}$ & $83.40_{\pm 0.20}$ \\
        ~ & ~ & GNSP \cite{lee2023robust}
 & $90.40_{\pm 1.40}$ & $84.80_{\pm 1.59}$ & $\underline{87.47_{\pm 0.70}}$ \\
         ~ & ~ &   {FreqPure} \cite{pei2025diffusion}
 &  $\mathbf{93.96_{{\pm}0.09}}$ &  $\underline{85.83_{\pm 0.17}}$ &  $\mathbf{88.26_{{\pm}0.51}}$  \\
         ~ & ~ & \mymethod~(Ours)
 & $88.73_{\pm 0.50}$ & $84.00_{\pm 1.71}$ & $83.40_{\pm 1.44}$ \\ \midrule
        \multirow{7}{*}{\rotatebox{90}{WRN-70-16}} & \multirow{3}{*}{AT} & Rebuffi et al. \cite{rebuffi2021fixing} $^*$
 & $\mathbf{95.74}$ & $\mathbf{89.62}$ & $82.32$ \\
         ~ & ~ & Gowal et al. \cite{gowal2020uncovering} $^*$
 & \underline{$94.74$} & \underline{$88.18$} & $80.53$ \\
         ~ & ~ & Rebuffi et al. \cite{rebuffi2021fixing}
 & $92.41$ & $86.24$ & $80.42$ \\ \cline{2-6}
         ~ & \multirow{4}{*}{AP} & ADP \cite{yoon2021adversarial}
 & $86.76_{\pm 1.15}$ & $75.66_{\pm 1.29}$ & $80.43_{\pm 0.42}$ \\
         ~ & ~ & DiffPure \cite{nie2022diffusion}
 & $90.00_{\pm 0.87}$ & $82.07_{\pm 1.79}$ & $82.67_{\pm 1.94}$ \\
        ~ & ~ & GNSP \cite{lee2023robust}
 & $90.53_{\pm 0.58}$ & $85.40_{\pm 0.60}$ & \underline{$87.73_{\pm 1.01}$} \\
        ~ & ~ &   {FreqPure} \cite{pei2025diffusion}
 &  $94.18_{{\pm}0.20}$ &  $85.83_{{\pm}0.75}$ &  $\mathbf{89.89_{\pm1.64}}$  \\
         ~ & ~ & \mymethod~(Ours)
 & $88.27_{\pm 0.50}$ & $82.93_{\pm 1.45}$ & $83.47_{\pm 0.46}$ \\
  \bottomrule
    \end{tabular}
    }
    \end{threeparttable}
    \end{center}
    \end{minipage}
\end{table*}

%-------------------------------------------------------------------------
\section{Experiments}

\subsection{Experimental Settings}
\label{sec:Experimental_settings}
\textbf{Datasets and network architectures.}
We conduct the experiments on two datasets: CIFAR-10\footnote{\scriptsize\url{https://www.tensorflow.org/datasets/catalog/cifar10}}\cite{krizhevsky2009learning} and ImageNet\footnote{\scriptsize\url{https://image-net.org}}\cite{russakovsky2015imagenet}. For ImageNet, we adopt its sub-dataset\footnote{\scriptsize\url{https://www.kaggle.com/datasets/ambityga/imagenet100}}, ImageNet-100, which includes 100 classes of the original 1000. 
This choice balances computational feasibility and dataset complexity, as training consistency models on larger datasets like ImageNet-1K is resource-intensive \cite{song2023consistency}. Notably, our method remains applicable on larger datasets if pre-trained consistency models are available.
% We implement various size of pre-trained WideResNet \cite{Zagoruyko2016WRN} as the classification models on the CIFAR-10, and extend the use of ResNet \cite{he2016deep} on the ImageNet-100. 
We use pre-trained WideResNet \cite{Zagoruyko2016WRN} models of varying sizes for CIFAR-10 classification and ResNet \cite{he2016deep} for ImageNet-100.
Particularly, we fine-tune the fully connected (FC) layer of classifiers to align predictions with the label space of ImageNet-100. By employing the automatic mixup \cite{liu2022automix}, the classification accuracies
are $78.62\%$ for ResNet50, $78.08\%$ for ResNet101 and $80.34\%$ for WRN-50-2. 

\textbf{Implementation details.} For the consistency models in our \mymethod, we adopt the pre-trained model released from the paper \cite{song2023consistency} on CIFAR-10, with only a conversion from the JAX implementation\footnote{\scriptsize\url{https://openaipublic.blob.core.windows.net/consistency/jcm_checkpoints/cd-lpips/checkpoints/checkpoint_80}} to the PyTorch version. While on ImageNet-100, we train the consistency model according to the consistency training following \cite{song2023consistency}. 
The diffusion trajectory of the consistency model aligns with that of the corresponding diffusion model \cite{karras2022elucidating}, characterized by the parameters $\mathbf{\mu}_{T}=\mathbf{0}$, $\sigma_{0}=0.002$ and $\sigma_{T}=80$.
Through the optimization parameters in our \mymethod, we set $\alpha=2, \beta=5 \times 10^{-4}$ with $200$ iterations on CIFAR-10 and $\alpha=5 \times 10^{-3}, \beta=1 \times 10^{-4}$ with $300$ iterations on ImageNet-100.

\textbf{Attack methods.}
Following \cite{lee2023robust}, we consider the commonly used $\ell_{\infty}$ and $\ell_{2}$ white-box attack methods to evaluate our method, 
including PGD \cite{madry2017towards}, AutoAttack \cite{croce2020reliable} and BPDA \cite{athalye2018obfuscated}. To strengthen these attacks, we approximate the surrogate gradient by simulating the entire denoising process from the midpoint of the diffusion, as done in \cite{lee2023robust}. This approach substantially increases the effectiveness of attack, providing a more rigorous robustness evaluation of the defense methods.
Specifically, we use 200 iterations for PGD and BPDA, while using the standard version of AutoAttack. Due to the stochasticity introduced by the randomized defenses, we also employ Expectation over Transformation (EOT) \cite{athalye2018synthesizing} with $n_{EOT} = 20$ when applying these attacks.  For $\ell_{\infty}$-norm attacks, we set $\epsilon = 8/255$ on CIFAR-10 and $\epsilon = 4/255$ on ImageNet-100, and for $\ell_{2}$-norm attacks, we use $\epsilon = 0.5$. To further demonstrate the superiority of our method, we also consider the relatively high attack intensities on CIFAR-10, i.e., $\ell_{\infty}$-norm with $\epsilon=16/255$ and $\ell_{2}$-norm with $\epsilon=1$.

Note that although existing strong white-box attacks such as AutoAttack, which are designed for direct perturbation in the input space, may not fully exploit the unique nature of our approach, applying them to our method ensures we are benchmarking against the most challenging and widely accepted adversarial settings. To this end, similar to attacking other diffusion-based adversarial purification methods \citet{yoon2021adversarial,nie2022diffusion,lee2023robust,pei2025diffusion}, we select a timestep $t_{diff}=0.3$ for attack based on the attack success rate (ASR) on both the classifier and purification module, as shown in Tab. \ref{tab: select_of_t}, with other settings kept the same.
Moreover,  we also provide a consistency-disruption attack tailored to our \mymethod~in Sec. \ref{sec: Adtaptive Attack} to evaluate its robustness in Sec. \ref{sec: exper on adaptive attack}. We leave more white-box attacks specifically targeting generative-model-based purification frameworks like ours as an important direction for future work.

\begin{table}[t]
    \caption{Standard and robust accuracy against PGD+EOT attack with $\ell_{\infty}$-norm ($\epsilon=4/255$) and $\ell_{2}$-norm ($\epsilon=0.5$) on ImageNet-100.}
    \begin{center}
    \begin{threeparttable}
    \resizebox{\linewidth}{!}{
    \begin{tabular}{clccc}
    \toprule
         Classifier &  Method & Standard & PGD+EOT  & PGD-$\ell_{2}$+EOT  \\ \midrule
         \multirow{3}{*}{ResNet50} & DiffPure \cite{nie2022diffusion}
 & $58.40_{\pm 1.10}$ & $28.57_{\pm 1.82}$ & $46.96_{\pm 1.44}$ \\
         ~ & GNSP \cite{lee2023robust}
 & \underline{$59.92_{\pm 1.24}$} & \underline{$33.40_{\pm 1.44}$} & \underline{$51.06_{\pm 1.74}$} \\
        ~ &   {FreqPure} \cite{pei2025diffusion}
 &  $56.36_{\pm 1.02}$ &  $30.83_{\pm 1.59}$ &  $43.54_{\pm 1.90}$ \\
         ~ & \mymethod~(Ours)
 & $\mathbf{61.93_{\pm 1.86}}$ & $\mathbf{39.87_{\pm 2.58}}$ & $\mathbf{54.67_{\pm 1.40}}$ \\ \midrule
        \multirow{3}{*}{ResNet101} & DiffPure \cite{nie2022diffusion}
 & $58.93_{\pm 2.15}$ & $28.97_{\pm 1.77}$ & $50.26_{\pm 2.98}$ \\
         ~ & GNSP \cite{lee2023robust}
 & \underline{$60.38_{\pm 2.89}$} & \underline{$32.61_{\pm 1.39}$} & \underline{$51.98_{\pm 2.15}$} \\
         ~ &   {FreqPure} \cite{pei2025diffusion}
 &  $56.46_{\pm 1.38}$ &  $30.83_{\pm 1.45}$ &  $44.58_{\pm 1.98}$ \\
         ~ & \mymethod~(Ours)
 & $\mathbf{61.53_{\pm 1.45}}$ & $\mathbf{39.27_{\pm 1.79}}$ & $\mathbf{52.60_{\pm 3.47}}$ \\ \midrule
         \multirow{3}{*}{WRN-50-2} & DiffPure \cite{nie2022diffusion}
 & $61.44_{\pm 1.02}$ & $28.57_{\pm 2.41}$ & \underline{$51.46_{\pm 1.35}$} \\
         ~ & GNSP \cite{lee2023robust}
 & \underline{$62.10_{\pm 1.21}$} & \underline{$31.68_{\pm 0.12}$} & $50.99_{\pm 2.10}$ \\
         ~ &   {FreqPure} \cite{pei2025diffusion}
 &  $57.04_{\pm 1.43}$ &  $30.26_{\pm 1.10}$ &  $46.81_{\pm 1.79}$ \\
         ~ & \mymethod~(Ours)
 & $\mathbf{63.33_{\pm 2.32}}$ & $\mathbf{40.47_{\pm 3.59}}$ & $\mathbf{54.73_{\pm 3.18}}$ \\
  \bottomrule
    \end{tabular}
    }
    \end{threeparttable}
    \end{center}
    \label{tab: comparison_on_imagenet}
    \vspace{-1.8em}
\end{table}

\textbf{Baselines.}
To evaluate the effectiveness of our method, we compare with the state-of-the-art defense methods as reported in the standard benchmark RobustBench \cite{croce2021robustbench}, including \textbf{adversarial training} \cite{pang2022robustness,gowal2021improving,gowal2020uncovering,rebuffi2021fixing,augustin2020adversarial,sehwag2022robust} and \textbf{adversarial purification} \cite{yoon2021adversarial,nie2022diffusion,lee2023robust,pei2025diffusion}. Following the settings in \cite{lee2023robust}, we inherit the results of the adversarial training methods from \cite{lee2023robust} due to the huge computational cost. For adversarial purification methods, DiffPure \cite{nie2022diffusion} uses diffusion timesteps of $t_{diff}=0.1$ for the $\ell_{\infty}$-norm and $t_{diff}=0.075$ for the $\ell_{2}$-norm on CIFAR-10, and $t_{diff}=0.15$ on ImageNet-100, respectively. GNSP \cite{lee2023robust} conducts purification in eight steps with $t_{diff} \in \{0.03 \times 4, 0.05 \times 2, 0.125 \times 2 \}$ and $t_{diff} \in \{0.03 \times 4, 0.05 \times 2, 0.2 \times 2 \}$ for both datasets.  {FreqPure \cite{pei2025diffusion} use conducts purification in one steps with $t_{diff} = 0.1$ on CIFAR-10 and eight steps with $t_{diff} \in \{ 0.05 \times 8 \}$ on ImageNet-100. Additionally, we also incorporate a blind image restoration method DiffBIR \cite{lin2024diffbir} as a baseline, where we train its denoiser using its official code to evaluate its robustness.}

\textbf{Evaluation metrics.}
We consider two metrics to evaluate the performance of defense approaches: \textbf{standard accuracy} and \textbf{robust accuracy}. The standard accuracy assesses the classification performance of the defense model on clean samples, while the robust accuracy evaluates the model's performance against adversarial attacks. An ideal purification module is supposed to enhance robust accuracy and strive to maintain standard accuracy.
Considering the computational expense of applying several attacks and purification methods, especially on ImageNet-100, following \citet{gao2021maximum}, we randomly select a subset containing 500 samples and compute accuracy over it unless otherwise specified. Additionally, to ensure more stable and reliable results, we report the mean and standard deviation over 3 independent runs.

\begin{table}[t]
    \caption{Standard and robust accuracy against BPDA+EOT attack with $\ell_{\infty}$-norm on CIFAR-10 ($\epsilon=8/255$) and ImageNet-100 ($\epsilon=4/255$). WideResNet-28-10 is used as classifier on CIFAR-10 while ResNet50 is used on ImageNet-100.}
    \vspace{-0.5em}
    \renewcommand\arraystretch{1.3}
    \begin{center}
    \begin{threeparttable}
    \resizebox{0.9\linewidth}{!}{
    \begin{tabular}{clcc}
    \toprule
         Dataset & Method & Standard & BPDA+EOT  \\ \midrule
        \multirow{4}{*}{\rotatebox{0}{CIFAR-10}} 
         ~ & ADP \cite{yoon2021adversarial}
 & $85.66_{\pm 0.51}$ & $66.91_{\pm 1.75}$ \\
         ~ & DiffPure \cite{nie2022diffusion}
 & $90.27_{\pm 0.81}$ & $82.33_{\pm 0.42}$ \\
         ~ & GNSP \cite{lee2023robust}
 & $\underline{90.40_{\pm 1.40}}$ & $\mathbf{88.53_{\pm 1.14}}$ \\
         ~ &   {FreqPure} \cite{pei2025diffusion}
 &  $\mathbf{93.96_{\pm 0.09}}$ &  \underline{$86.46_{\pm0.37}$} \\
         ~ & \mymethod~(Ours)
 & $88.73_{\pm 0.50}$ & $83.53_{\pm 1.80}$ \\ \midrule
        \multirow{3}{*}{\rotatebox{0}{ImageNet-100}} & DiffPure \cite{nie2022diffusion}
 & $58.40_{\pm 1.10}$ & $46.56_{\pm 2.12}$ \\
        ~ & GNSP \cite{lee2023robust}
 & \underline{$59.92_{\pm 1.24}$} & \underline{$48.68_{\pm 1.71}$} \\
        ~ &   {FreqPure} \cite{pei2025diffusion}
 &  $56.36_{\pm 1.02}$ &  $19.58_{\pm 2.39}$ \\
        ~ & \mymethod~(Ours)
 & $\mathbf{61.93_{\pm 1.86}}$ & $\mathbf{49.47_{\pm 2.12}}$ \\
  \bottomrule
    \end{tabular}
    }
    \end{threeparttable}
    \end{center}
    \label{tab: comparison_on_bpda}
    \vspace{-1em}
\end{table}

\begin{table}[t]
    \caption{ {Comparisons of our CMAP improved by DiffPure \cite{nie2022diffusion} with other baselines against $\ell_{\infty}$-norm ($\epsilon=8/255$) and $\ell_{2}$-norm ($\epsilon=0.5$) attacks on CIFAR-10, with WideResNet-28-10 used as a classifier.}
    }
    \vspace{-0.5em}
    \begin{center}
    \begin{threeparttable}
    \resizebox{\linewidth}{!}{
    \begin{tabular}{lcccc}
    \toprule
         Method & Standard & PGD-$\ell_{2}$+EOT  & AutoAttack-$\ell_{2}$ & BPDA+EOT  \\ \midrule
        DiffPure \cite{nie2022diffusion}
& $90.27_{\pm 0.81}$ & $82.00_{\pm 0.72}$ & $83.40_{\pm 0.20}$ & $82.33_{\pm 0.42}$ \\
         GNSP \cite{lee2023robust}
& $90.40_{\pm 1.40}$ & $\underline{84.80_{\pm 1.59}}$ & $87.47_{\pm 0.70}$ & \underline{$88.53_{\pm 1.14}$} \\
  {FreqPure} \cite{pei2025diffusion}
 &  $\mathbf{93.96_{{\pm}0.09}}$ &  $\mathbf{85.83_{\pm 0.17}}$ &  $\underline{88.26_{{\pm}0.51}}$ &   $86.46_{\pm0.37}$\\
         \mymethod
& $88.73_{\pm 0.50}$ & $84.00_{\pm 1.71}$ & $83.40_{\pm 1.44}$ & $83.53_{\pm 1.80}$ \\
        \mymethod+DiffPure \cite{nie2022diffusion}
& $\underline{91.33_{\pm 0.23}}$ & $83.60_{\pm 0.69}$ & $\mathbf{88.47_{\pm 0.31}}$ & $\mathbf{89.47_{\pm 0.23}}$ \\\bottomrule
    \end{tabular}
    }
    \end{threeparttable}
    \end{center}
    \label{tab: CMAP_aug}
    \vspace{-1em}
\end{table}

\begin{table}[t]
    \caption{ {Comparisons with blind image restoration methods against $\ell_{\infty}$-norm ($\epsilon=8/255$) and $\ell_{2}$-norm ($\epsilon=0.5$) attacks on CIFAR-10, with WideResNet-28-10 used as a classifier.}
    }
    \vspace{-0.5em}
    \begin{center}
    \begin{threeparttable}
    \resizebox{\linewidth}{!}{
    \begin{tabular}{lccccc}
    \toprule
         Method & Standard & PGD+EOT  & PGD-$\ell_{2}$+EOT  & AutoAttack & AutoAttack-$\ell_{2}$ \\ \midrule
          {DiffBIR} \cite{lin2024diffbir}
&   {$\mathbf{94.20_{\pm 0.31}}$} &  $23.44_{\pm 1.43}$ &   $81.02_{\pm 1.32}$ &   $20.10_{\pm 0.52}$ &   $55.06_{\pm 0.26}$ \\
         GNSP \cite{lee2023robust}
& $\underline{90.40_{\pm 1.40}}$ &  $\underline{55.87_{\pm 0.50}}$ &  $\mathbf{84.80_{\pm 1.59}}$ & $\underline{70.40_{\pm 1.80}}$ & $\mathbf{87.47_{\pm 0.70}}$ \\
         \mymethod~(Ours)
& $88.73_{\pm 0.50}$ & $\mathbf{74.60_{\pm 1.59}}$  & $\underline{84.00_{\pm 1.71}}$ & $\mathbf{78.67_{\pm 1.90}}$ & $\underline{83.40_{\pm 1.44}}$ \\ \bottomrule
    \end{tabular}
    }
    \end{threeparttable}
    \end{center}
    \label{tab: comparison_restore}
    \vspace{-1em}
\end{table}

\begin{table}[t]
    \caption{Standard and robust accuracy against high intensity attacks with $\epsilon=16/255$ for $\ell_{\infty}$-norm and $\epsilon=1$ for $\ell_{2}$-norm on CIFAR-10. All purification methods are implemented according to their original parameter settings.}
    \vspace{-0.5em}
    \begin{center}
    \begin{threeparttable}
    \resizebox{\linewidth}{!}{
    \begin{tabular}{lccccc}
    \toprule
         Method & Standard & PGD+EOT  & PGD-$\ell_{2}$+EOT  & AutoAttack & AutoAttack-$\ell_{2}$ \\ \midrule
        DiffPure \cite{nie2022diffusion}
& $90.27_{\pm 0.81}$ & $13.33_{\pm 1.33}$ & $65.98_{\pm 0.47}$ & $21.40_{\pm 1.04}$ & $64.09_{\pm 1.59}$ \\
         GNSP \cite{lee2023robust}
& \underline{$90.40_{\pm 1.40}$} & \underline{$21.10_{\pm 1.32}$} & \underline{$72.00_{\pm 0.71}$} & \underline{$33.80_{\pm 2.40}$} & \underline{$72.09_{\pm 1.80}$} \\
          {FreqPure} \cite{pei2025diffusion}
&    {$\mathbf{93.96_{\pm 1.24}}$} &    {$9.17_{\pm 0.69}$} &    {$68.16_{\pm 0.81}$} &    {$27.29_{\pm 1.34}$} &    {$66.46_{\pm 0.92}$}  \\
         \mymethod~(Ours)
& $88.73_{\pm 0.50}$ & $\mathbf{64.53_{\pm 1.89}}$ & $\mathbf{78.33_{\pm 1.17}}$ & $\mathbf{80.73_{\pm 1.63}}$ & $\mathbf{80.33_{\pm 1.75}}$ \\ \bottomrule
    \end{tabular}
    }
    \end{threeparttable}
    \end{center}
    \label{tab: comparison_on_high_intensity}
    \vspace{-1em}
\end{table}

\begin{table}[t]
    \caption{ {Comparisons on image fidelity metrics quantifying the similarity between purified samples (after PGD+EOT attack) and original clean samples on CIFAR-10 over WideResNet-28-10 ($\epsilon=8/255$) and ImageNet-100 over ResNet50 ($\epsilon=4/255$).}}
    \vspace{-0.5em}
    \begin{center}
    \begin{threeparttable}
    % --- 间距设置 ---
    \renewcommand{\arraystretch}{1.3} % 增加行高，避免拥挤
    \setlength{\tabcolsep}{4pt}       % 调整列宽
    % ---------------
    \resizebox{0.9\linewidth}{!}{
    {
    \begin{tabular}{l|l|ccc|c}
    \toprule
    Data & ~~Method & MSE$\downarrow$ & SSIM$\uparrow$ & LPIPS$\downarrow$ & Robust Accuracy$\uparrow$\\ \midrule
    \multirow{4}{*}{\rotatebox{90}{CIFAR-10}} 
    & {DiffPure} \cite{nie2022diffusion}
        & \underline{$0.0047$} & $0.8059$ & \underline{$0.0220$} & $48.27_{\pm 1.86}$ \\
    & {GNSP} \cite{lee2023robust}
        & $0.0076$ & $0.7162$ & $0.0493$ & $55.87_{\pm 0.50}$ \\
    & FreqPure \cite{pei2025diffusion}
        & $\mathbf{0.0017}$ & $\mathbf{0.9178}$ & $\mathbf{0.0089}$ & $45.00_{{\pm}0.59}$ \\
    &  {\mymethod~(Ours)}
        &  $0.0101$ &  \underline{$0.8332$} &  $0.0479$ &  $\mathbf{74.60_{\pm 1.59}}$ \\ \midrule
    % --- ImageNet-100 ---
    \multirow{4}{*}{\rotatebox{90}{ImageNet-100}} 
    & {DiffPure} \cite{nie2022diffusion}
        & $\mathbf{0.0040}$ & \underline{$0.7841$} & $\mathbf{0.0655}$ & $28.57_{\pm{1.82}}$ \\
    & {GNSP} \cite{lee2023robust}
        & $0.0068$ & $0.6437$ & \underline{$0.1132$} & $33.40_{\pm{1.44}}$ \\
    & FreqPure \cite{pei2025diffusion}
        & \underline{$0.0056$} & $0.5894$ & $0.1533$ & $30.83_{\pm{1.59}}$ \\
    &  {\mymethod~(Ours)}
        &  $0.0057$ &  $\mathbf{0.8473}$ &  $0.1143$ &  $\mathbf{39.87}_{\pm\mathbf{2.58}}$ \\ 
    \bottomrule
    \end{tabular}
    }}
    \end{threeparttable}
    \end{center}
    \label{tab: comparison_on_img_similarity_scaled}
    \vspace{-1em}
\end{table}

\begin{table}[t]
    \caption{ {Comparisons with different generative paradigms (\eg, GAN and VAE) against $\ell_{\infty}$-norm ($\epsilon=8/255$) and $\ell_{2}$-norm ($\epsilon=0.5$) attacks on CIFAR-10, with WideResNet-28-10 used as a classifier.}
    }
    \vspace{-0.5em}
    \begin{center}
    \begin{threeparttable}
    \resizebox{\linewidth}{!}{
    \begin{tabular}{lccccc}
    \toprule
         Method & Standard & PGD+EOT  & PGD-$\ell_{2}$+EOT  & AutoAttack & AutoAttack-$\ell_{2}$ \\ \midrule
         DiffPure \cite{nie2022diffusion}
& $90.27_{\pm 0.81}$ & $48.27_{\pm 1.86}$ & $82.00_{\pm 0.72}$ & $64.93_{\pm 2.14}$ & $83.40_{\pm 0.20}$ \\
         GNSP \cite{lee2023robust}
& $\underline{90.40_{\pm 1.40}}$ &  $55.87_{\pm 0.50}$ &  $\mathbf{84.80_{\pm 1.59}}$ & $70.40_{\pm 1.80}$ & $\mathbf{87.47_{\pm 0.70}}$ \\
          {\mymethod~(GAN)}
&   {$45.80_{\pm 0.74}$} &  $40.80_{\pm 0.81}$ &   $40.20_{\pm 1.62}$ &   $35.20_{\pm 1.64}$ &   $43.20_{\pm 0.55}$ \\
          {\mymethod~(VAE)}
&   {$\mathbf{92.80_{\pm 0.52}}$} &  $\underline{58.80_{\pm 1.20}}$ &   $82.80_{\pm 0.68}$ &   $\underline{74.60_{\pm 1.29}}$ &   $83.20_{\pm 0.93}$ \\
         \mymethod
& $88.73_{\pm 0.50}$ & $\mathbf{74.60_{\pm 1.59}}$  & $\underline{84.00_{\pm 1.71}}$ & $\mathbf{78.67_{\pm 1.90}}$ & $\underline{83.40_{\pm 1.44}}$ \\ \bottomrule
    \end{tabular}
    }
    \end{threeparttable}
    \end{center}
    \label{tab: comparison_generative_cifar}
    \vspace{-1em}
\end{table}

\begin{table}[t]
    \caption{ {Comparisons with different generative paradigms (\eg, VAE) against $\ell_{\infty}$-norm ($\epsilon=4/255$) and $\ell_{2}$-norm ($\epsilon=0.5$) attacks on ImageNet-100, with ResNet-50 used as a classifier.}
    }
    \vspace{-0.5em}
    \begin{center}
    \begin{threeparttable}
    \resizebox{0.8\linewidth}{!}{
    \begin{tabular}{lccc}
    \toprule
         Method & Standard & PGD+EOT  & PGD-$\ell_{2}$+EOT \\ \midrule
         DiffPure \cite{nie2022diffusion}
 & $58.40_{\pm 1.10}$ & $28.57_{\pm 1.82}$ & $46.96_{\pm 1.44}$ \\
         GNSP \cite{lee2023robust}
 & $59.92_{\pm 1.24}$ & $33.40_{\pm 1.44}$ & $51.06_{\pm 1.74}$ \\
          {\mymethod~(VAE)}
&   {$\mathbf{68.20_{\pm 0.32}}$} &  $\mathbf{43.60_{\pm 1.37}}$ &   $\underline{53.80_{\pm 0.84}}$ \\
         \mymethod~(Ours)
 & \underline{$61.93_{\pm 1.86}$} & $\underline{39.87_{\pm 2.58}}$ & $\mathbf{54.67_{\pm 1.40}}$ \\ \bottomrule
    \end{tabular}
    }
    \end{threeparttable}
    \end{center}
    \label{tab: comparison_generative_imagenet}
    \vspace{-1em}
\end{table}

\subsection{Observations on Generated Samples}
\label{sec:Observations}
We start by exploring the distributional characteristics of clean, adversarial and generated samples, which inspires the proposal of our \mymethod. Leveraging its effectiveness in quantifying the distributional discrepancy between two distributions, we calculate the Expected Perturbation Score (EPS, $\mathbb{E}_{t \sim U(0, T)} \nabla_{\mathbf{x}} \log p_t\left(\mathbf{x}\right)$) \cite{zhangs2023EPSAD} that introduces increasing Gaussian noise to the sample and averages their scores, thereby extracting comprehensive distribution information from its multi-view observations. For generated samples, we employ multiple generative models \cite{song2023consistency, nichol2021improved, dhariwal2021diffusion} on both CIFAR-10 and ImageNet, while clean samples are randomly drawn from the corresponding datasets. For adversarial samples, we craft them by PGD-$\ell_{\infty}$+EOT attack ($\epsilon=4/255$). Taking 500 clean samples as a reference set, we compute the Maximum Mean Discrepancy (MMD) \cite{gretton2012kernel} between this set and each of the 500 clean (distinct from the reference), adversarial, generated samples. The results are then visualized through histograms.

From Fig. \ref{fig: motivition}, two cases (consistency model \cite{song2023consistency} on CIFAR-10 and diffusion model \cite{dhariwal2021diffusion} on ImageNet) demonstrate that the generated samples tend to be closely aligned with clean samples, while remaining significantly distant from adversarial ones. This indicates that the latent space of the generative model is well-aligned with the clean data manifold, enabling the generation of samples that closely resemble clean data and are resistant to adversarial perturbations, underscoring the efficacy of our latent distribution consistency constraint.  {We also show two more generative models in Fig. \ref{fig: motivition_appendix} and various attacks in Fig. \ref{fig: motivation} in Appendix.}

\subsection{Comparison on Benchmark Datasets}
\label{sec:Comparison_with_SOTA}
We compare \mymethod~with the state-of-the-art adversarial training and purification methods against both $\ell_{\infty}$ and $\ell_{2}$ threat models, as shown in Tab. \ref{tab: comparison_on_cifar}, \ref{tab: comparison_on_imagenet}, \ref{tab: comparison_on_bpda}, \ref{tab: CMAP_aug} and \ref{tab: comparison_restore}.

\begin{table}[t]
    \caption{Attack success rate (ASR) with surrogate gradients approximated at different diffusion midpoints ($t_{diff}$). We select $t_{diff}=0.3$ to balance the ASR on both the classifier (Clf) and the purification (Pur).}
    \vspace{-0.7em}
    \label{tab: select_of_t}
    \renewcommand\arraystretch{1.3}
    \begin{center}
    \begin{threeparttable}
    \resizebox{\linewidth}{!}{
    \begin{tabular}{cccccccc}
    \toprule
         ~ & \multirow{2}{*}{Attack} & \multirow{2}{*}{ASR} & \multicolumn{5}{c}{$t_{diff}$} \\  \cline{4-8}
         ~ & ~ & ~ &$0.1$ & $0.2$ & $0.3$ & $0.4$ & $0.5$ \\ \midrule
         \multirow{4}{*}{\rotatebox{90}{CIFAR-10}} & \multirow{2}{*}{PGD+EOT } & Clf & $100.00\%$ & $99.20\%$ & $94.40\%$ & $28.80\%$ & $5.40\%$ \\
         ~ & ~ & Pur+Clf & $16.00\%$ & $16.60\%$ & $23.80\%$ & $28.00\%$ & $21.20\%$ \\  \cline{2-8}
         ~ & \multirow{2}{*}{AutoAttack-$\ell_{2}$} & Clf & $99.80\%$ & $93.00\%$ & $57.00\%$ & $5.40\%$ & $4.20\%$ \\
         ~ & ~ & Pur+Clf & $15.40\%$ & $14.60\%$ & $17.80\%$ & $15.60\%$ & $14.60\%$ \\ \midrule
        \multirow{4}{*}{\rotatebox{90}{ImageNet-100}} & \multirow{2}{*}{PGD+EOT } & Clf & $89.40\%$ & $86.60\%$ & $82.20\%$ & $51.20\%$ & $26.20\%$ \\
         ~ & ~ & Pur+Clf & $45.40\%$ & $50.20\%$ & $57.80\%$ & $59.80\%$ & $49.40\%$ \\  \cline{2-8}
         ~ & \multirow{2}{*}{PGD-$\ell_{2}$+EOT } & Clf & $94.80\%$ & $80.20\%$ & $67.00\%$ & $31.40\%$ & $24.40\%$ \\
         ~ & ~ & Pur+Clf & $43.60\%$ & $45.80\%$ & $47.00\%$ & $46.60\%$ & $45.40\%$ \\  \bottomrule
    \end{tabular}
    }
    \end{threeparttable}
    \end{center}
    \vspace{-1em}
\end{table}

\textbf{Results on CIFAR-10.}
Tab. \ref{tab: comparison_on_cifar} reports the defense performance against PGD+EOT and AutoAttack under $\ell_{\infty}$ ($\epsilon=8/255$) and $\ell_{2}$ ($\epsilon=0.5$) norm constraints on CIFAR-10, over two WideResNet classifiers \cite{Zagoruyko2016WRN}. Our \mymethod~demonstrates significant improvements over the latest adversarial training and purification methods against $\ell_{\infty}$ attacks, while maintaining comparable performance on $\ell_{2}$ attacks and clean samples.  {Particularly, \mymethod~surpasses other baselines by $18.73\%$ against PGD+EOT attack, and achieves improvements ranging from $6.54\%$ to $7.83\%$ on WideResNet-28-10 and WideResNet-70-16 for AutoAttack. We also include a comparison with blind image restoration techniques in Tab. \ref{tab: comparison_restore}. Although DiffBIR yields high standard accuracy ($94.20\%$ on CIFAR-10), its robust accuracy against PGD+EOT is very low ($23.44\%$), confirming that methods designed for natural image restoration often fail to remove adversarial noises effectively \cite{zhou2023eliminating, kostin2025robust}.} 

Tab. \ref{tab: comparison_on_bpda} shows the results of our \mymethod~against BPDA+EOT attack. 
 {While \mymethod~achieves slightly inferior performance than strong recent baselines like GNSP \cite{lee2023robust} and FreqPure \cite{pei2025diffusion}, its robustness can be further enhanced by integrating a diffusion-based pre-purification step (\eg, DiffPure). As shown in Tab. \ref{tab: CMAP_aug}, this combined approach significantly boosts the defense of \mymethod~against both AutoAttack-$\ell_2$ and BPDA+EOT, indicating that the performance gap can be effectively narrowed. Moreover, on more complex datasets such as ImageNet under BPDA attack, our method demonstrates superior robust accuracy (see Tab. \ref{tab: comparison_on_bpda}). }
In addition, while adversarial training outperforms adversarial purification in some cases, it exhibits a significant decline in robustness against unseen attacks \cite{nie2022diffusion}. 

\begin{figure*}[t]
\vspace{-1em}
    \begin{center}
    \vspace{-1em}
        \subfloat[Standard Accuracy]
        {\includegraphics[width=0.32\textwidth]{ 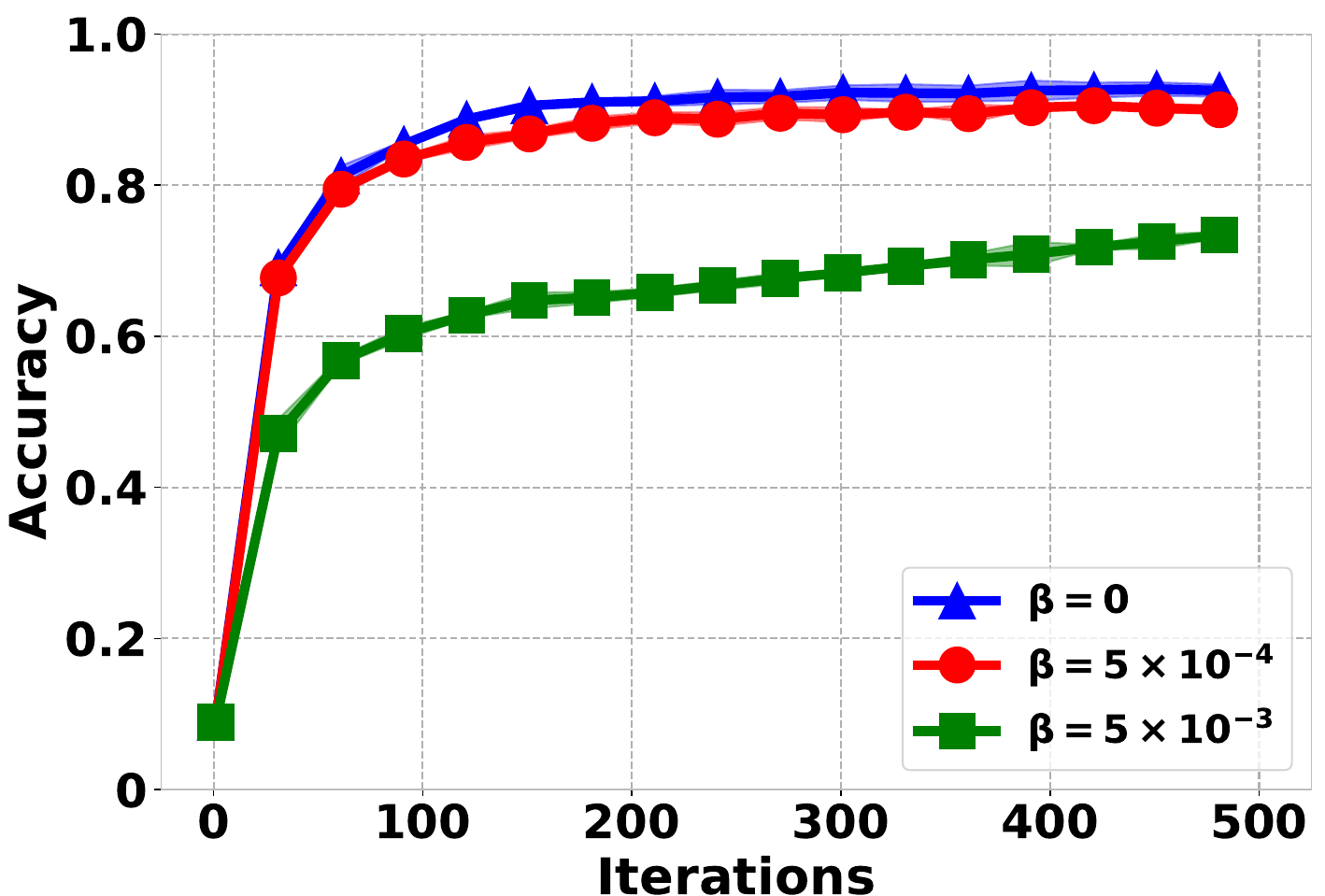}}
        \subfloat[PGD-$\ell_{\infty}$+EOT, $\epsilon=8/255$]
        {\includegraphics[width=0.32\textwidth]{ 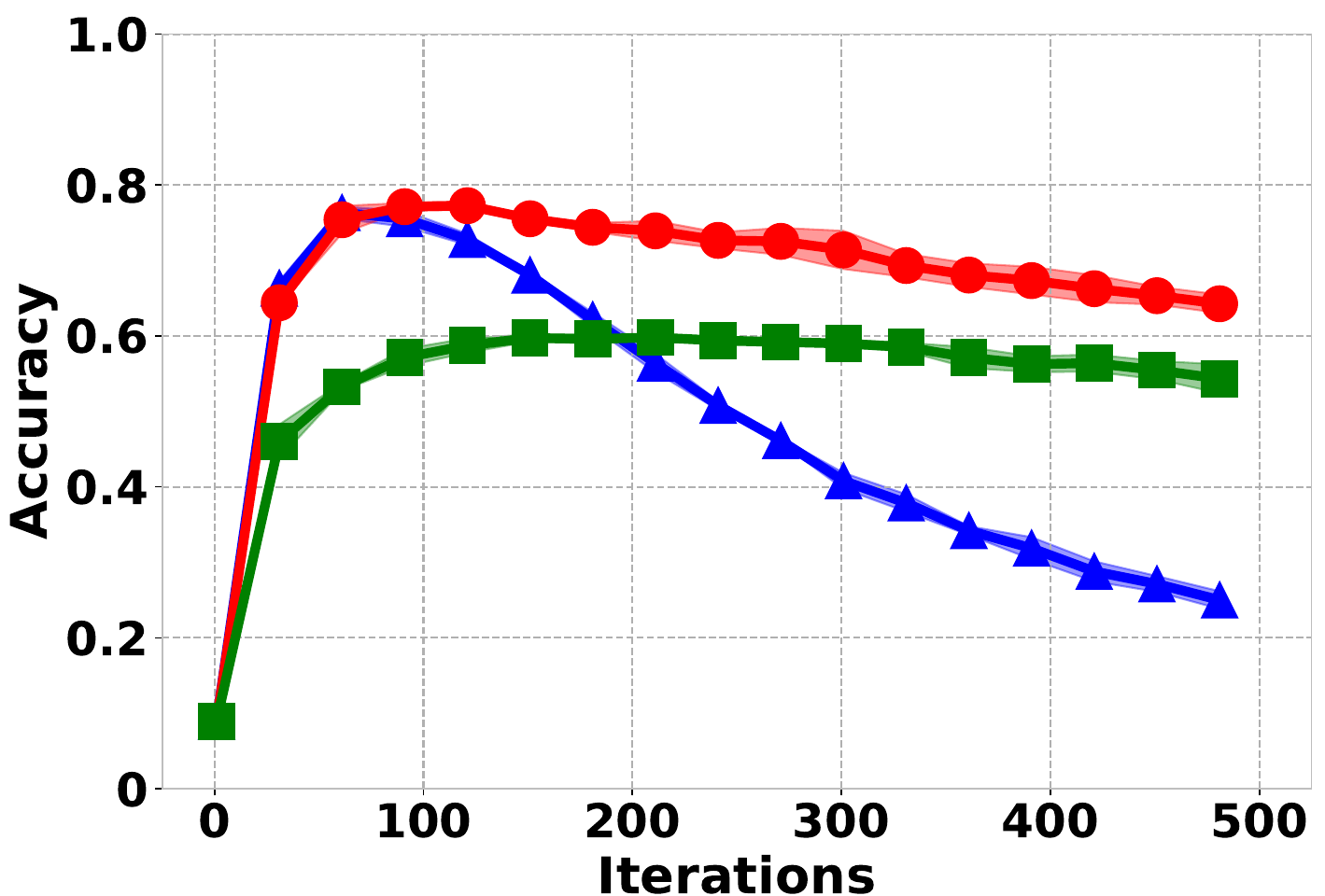}}
        \subfloat[PGD-$\ell_{2}$+EOT, $\epsilon=0.5$]
        {\includegraphics[width=0.32\textwidth]{ 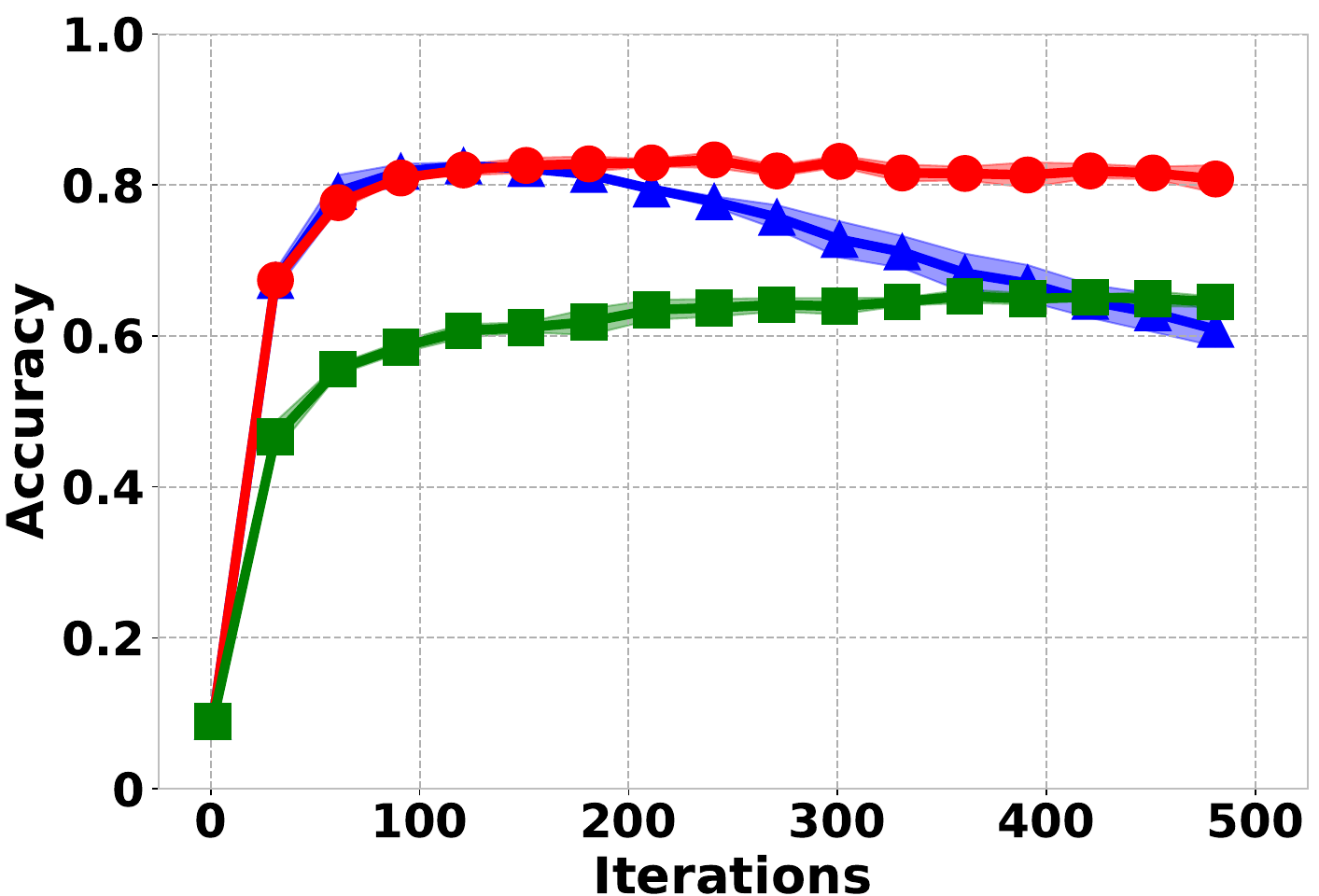}}
        \caption{Standard and robust accuracy curves under different $\beta$ against PGD+EOT attack with $\ell_{\infty}$-norm ($\epsilon=8/255$) and $\ell_{2}$-norm ($\epsilon=0.5$) on CIFAR-10, where we use WideResNet-28-10 as the classifier. The results indicate that the absence of constraint ($\beta=0$) leads to a significant drop in robust accuracy, while an excessive constraint ($\beta=5 \times 10^{-3}$) negatively impacts both standard and robust accuracy. In contrast, $\beta=5 \times 10^{-4}$ achieves a favorable balance between standard and robust accuracy, thereby simultaneously suppressing adversarial perturbations while maintaining effective image restoration.}
        \label{fig: ablation_beta}
    \end{center}
    \vspace{-1.8em}
\end{figure*}

\begin{figure}[t]
\vspace{-0.5em}
    \begin{center}
        \subfloat[CIFAR-10]
        {\includegraphics[width=0.5\linewidth]{ 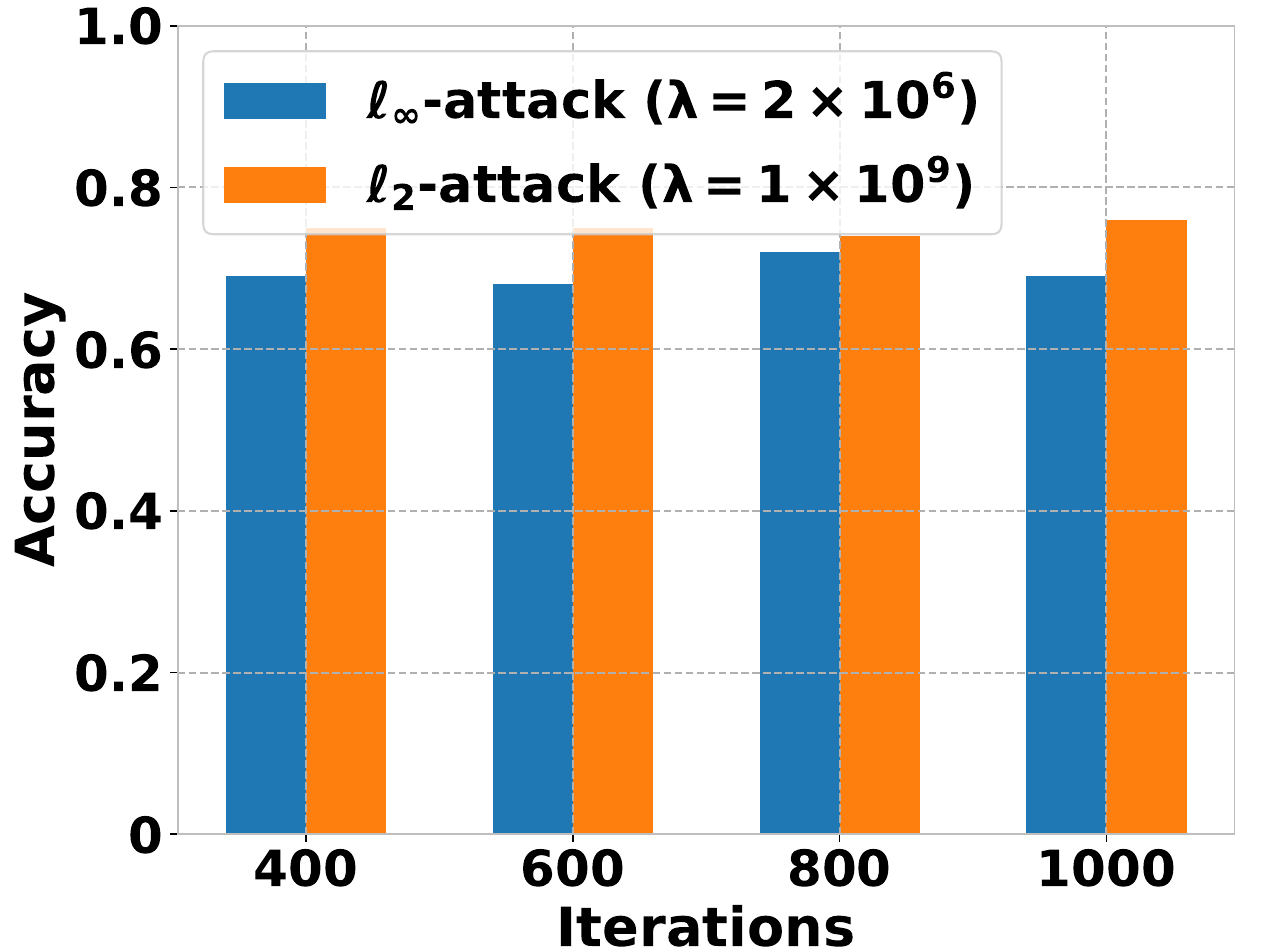}}
        \subfloat[ImageNet-100]
        {\includegraphics[width=0.49\linewidth]{ 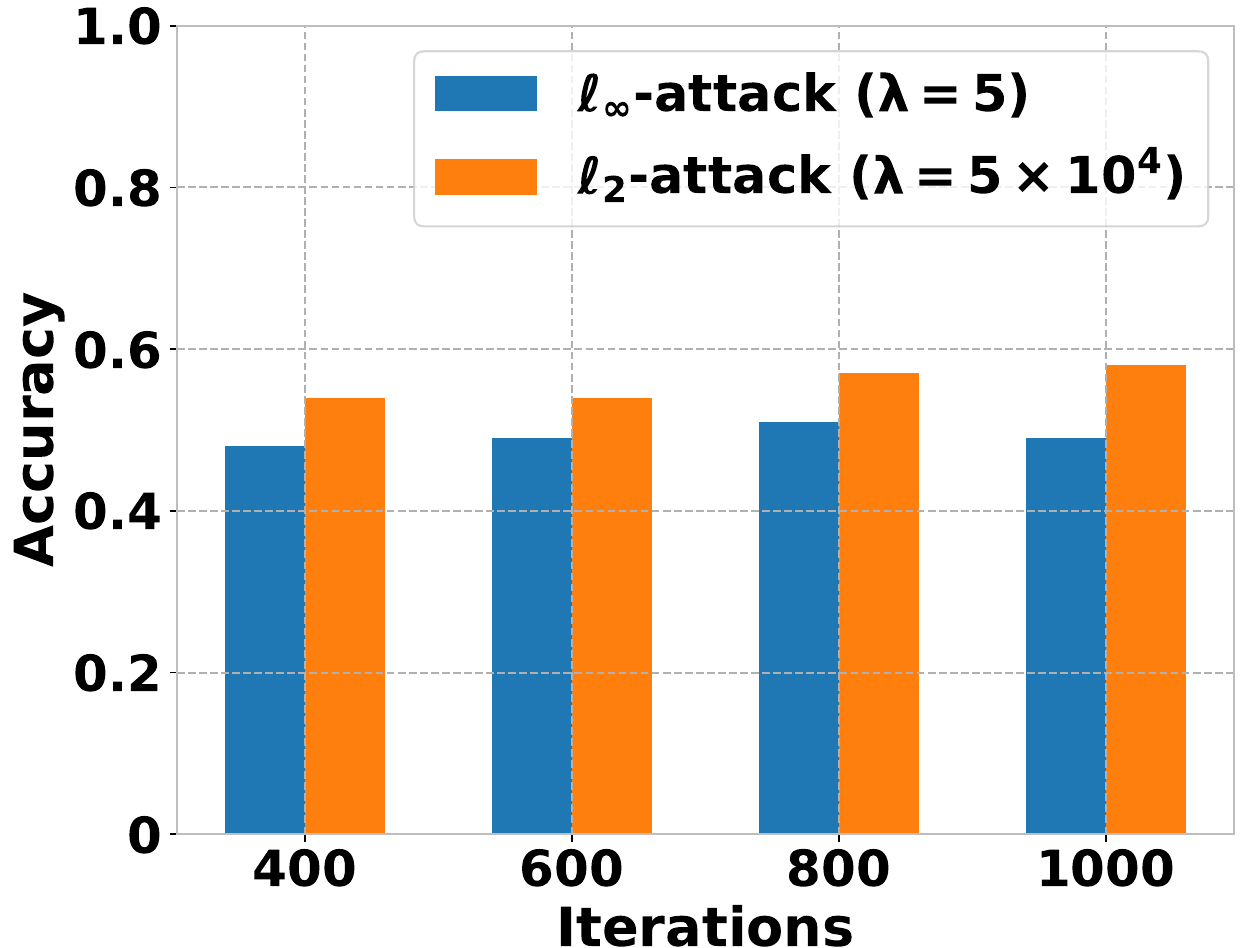}}
        \caption{Robust accuracy across iterations against consistency-disruption attack with $\ell_{\infty}$-norm and $\ell_{2}$-norm on WideResNet-28-10 for CIFAR-10 and ResNet50 for ImageNet-100. The alignment iterations $T_\mathrm{def}=200$ for CIFAR-10 and $T_\mathrm{def}=300$ for ImageNet-100, with subsequent attack iterations $T_\mathrm{adv}$ extending to $1000$. \mymethod~maintains high performance throughout the process.}
        \label{fig: adaptive_attack_accuracy}
    \end{center}
    \vspace{-1.7em}
\end{figure}

\begin{table}[t]
    \caption{Impact of the similarity factor $\alpha$ on standard and robust accuracy against $\ell_{\infty}$ ($\epsilon=8/255$) and $\ell_{2}$ ($\epsilon=0.5$) threat models, where we evaluate over WideResNet-28-10 on CIFAR-10.}
    \vspace{-0.5em}
    \begin{center}
    \begin{threeparttable}
    \resizebox{\linewidth}{!}{
    \begin{tabular}{l|ccccc}
    \toprule
     \diagbox{Acc}{$\alpha$} & $0.5$ & $1$ & $2$ & $5$ & $10$ \\ \midrule
     Clean & $81.20_{\pm 1.31}$ & $85.80_{\pm 0.53}$ & $88.73_{\pm 0.50}$ & $\mathbf{90.87_{\pm 0.46}}$ & $90.53_{\pm 0.61}$ \\
     PGD+EOT & $73.47_{\pm 1.55}$ & $\mathbf{75.07_{\pm 0.31}}$ & $74.60_{\pm 1.59}$ & $68.40_{\pm 2.42}$ & $63.73_{\pm 1.14}$ \\ 
     PGD-$\ell_{2}$+EOT & $77.53_{\pm 2.23}$ & $80.93_{\pm 1.29}$ & $\mathbf{84.00_{\pm 1.71}}$ & $82.67_{\pm 1.62}$ & $81.60_{\pm 1.25}$ \\
     BPDA+EOT & $77.53_{\pm 0.61}$ & $81.40_{\pm 0.35}$ & $\mathbf{83.53_{\pm 1.80}}$ & $80.60_{\pm 1.40}$ & $78.53_{\pm 1.55}$ \\ \bottomrule
    \end{tabular}
    }
    \end{threeparttable}
    \end{center}
    \label{tab: impact_of_alpha}
    \vspace{-1.8em}
\end{table}

\begin{table}[t]
    \caption{Impact of the Gaussian factor $\beta$ on standard and robust accuracy against $\ell_{\infty}$ ($\epsilon=8/255$) and $\ell_{2}$ ($\epsilon=0.5$) threat models, where we evaluate on WideResNet-28-10 for CIFAR-10.}
    \vspace{-0.5em}
    \begin{center}
    \begin{threeparttable}
    \resizebox{\linewidth}{!}{
    \begin{tabular}{l|ccccc}
    \toprule
     \diagbox{Acc}{$\beta$} & $1 \times 10^{-4}$ & $2 \times 10^{-4}$ & $5 \times 10^{-4}$ & $1 \times 10^{-3}$ & $2 \times 10^{-3}$ \\ \midrule
      Clean & $\mathbf{91.53_{\pm 0.31}}$ & $90.87_{\pm 0.64}$ & $88.73_{\pm 0.50}$ & $85.73_{\pm 0.81}$ & $80.67_{\pm 1.29}$ \\
     PGD+EOT & $65.07_{\pm 1.90}$ & $69.27_{\pm 2.60}$ & $\mathbf{74.60_{\pm 1.59}}$ & $74.27_{\pm 1.60}$ & $71.20_{\pm 0.35}$ \\ 
     PGD-$\ell_{2}$+EOT & $82.00_{\pm 0.69}$ & $83.00_{\pm 1.93}$ & $\mathbf{84.00_{\pm 1.71}}$ & $80.67_{\pm 1.22}$ & $74.40_{\pm 2.84}$ \\
     BPDA+EOT & $78.73_{\pm 0.81}$ & $80.67_{\pm 1.21}$ & $\mathbf{83.53_{\pm 1.80}}$ & $80.20_{\pm 2.36}$ & $75.67_{\pm 0.42}$ \\  \bottomrule
    \end{tabular}
    }
    \end{threeparttable}
    \end{center}
    \label{tab: impact_of_beta}
    \vspace{-1.8em}
\end{table}

\begin{table}[t]
    \caption{Impact of the latent vector number $K$ on standard and robust accuracy against $\ell_{\infty}$ ($\epsilon=8/255$) and $\ell_{2}$ ($\epsilon=0.5$) threat models, where we evaluate on WideResNet-28-10 for CIFAR-10.}
    \vspace{-0.5em}
    \begin{center}
    \begin{threeparttable}
    \resizebox{\linewidth}{!}{
    \begin{tabular}{l|cccc}
    \toprule
     \diagbox{Acc}{K} & $2$ & $5$ & $10$ & $20$ \\ \midrule
     Clean & $80.13_{\pm 1.17}$ & $87.27_{\pm 1.10}$ & $88.73_{\pm 0.50}$ & $\mathbf{89.27_{\pm 0.42}}$ \\
     PGD+EOT & $69.67_{\pm 2.19}$ & $73.67_{\pm 1.40}$ & $\mathbf{74.60_{\pm 1.59}}$ & $73.40_{\pm 1.64}$ \\ 
     PGD-$\ell_{2}$+EOT & $76.80_{\pm 3.27}$ & $81.20_{\pm 1.74}$ & $\mathbf{84.00_{\pm 1.71}}$ & $83.53_{\pm 1.14}$ \\
     BPDA+EOT & $76.13_{\pm 2.57}$ & $81.53_{\pm 1.75}$ & $\mathbf{83.53_{\pm 1.80}}$ & $82.47_{\pm 1.30}$ \\  \bottomrule
    \end{tabular}
    }
    \end{threeparttable}
    \end{center}
    \label{tab: impact_of_k}
    \vspace{-1.8em}
\end{table}

\textbf{Results on ImageNet-100.}
As shown in Tab. \ref{tab: comparison_on_imagenet} and \ref{tab: comparison_on_bpda}, our \mymethod~consistently outperforms state-of-the-art purification methods in both robust accuracy and standard accuracy on ImageNet-100 under $\ell_{\infty}$ ($\epsilon=4/255$) and $\ell_{2}$ ($\epsilon=0.5$) attacks. In particular, \mymethod~achieves  absolute improvements in robust accuracy ranging from $6.47\%$ to $8.79\%$ for PGD+EOT attacks, from $0.62\%$ to $3.74\%$ for PGD-$\ell_{2}$+EOT and $0.79\%$ for BPDA+EOT across different classifiers. Moreover, the standard accuracy also increases by $1.15\%$ to $2.01\%$.

The results clearly show the effectiveness of \mymethod~across various architectures and datasets in defending different attacks. Notably, \mymethod~is agnostic to classifier architectures or attack types, employing a unified defense operation, \eg, optimization iterations $T$. This contrasts with prior purification methods \cite{yoon2021adversarial,nie2022diffusion,lee2023robust}, which require different hyperparameters, \eg, the diffusion timestep $t_{diff}$, for different attacks.

\subsection{Purification on High Attack Intensity}

To further demonstrate the superiority of our \mymethod, we conduct experiments under conditions of extremely high attack intensity on CIFAR-10, \ie, using $\epsilon=16/255$ for $\ell_{\infty}$-norm attacks and $\epsilon=1$ for $\ell_{2}$-norm attacks. As shown in Tab. \ref{tab: comparison_on_high_intensity}, the diffusion-based purification baselines suffer from significant performance degradation. In contrast, our \mymethod~exhibits remarkable robustness, achieving improvements ranging from $43.43\%$ to $46.93\%$ for $\ell_{\infty}$-norm threats and $6.33\%$ to $8.24\%$ for $\ell_{2}$-norm threats. The results indicate that \mymethod~effectively maintains features of the clean data while removing adversarial noise, even under intense attack scenarios. This suggests that the latent space optimization by our method is more robust to high-intensity adversarial attacks, maintaining a stable purification process even under severe perturbations.

 {\subsection{Purification Fidelity and Visualizations}}

 {To assess the applicability of our purification framework beyond classification, we evaluate purification fidelity, which is defined as the perceptual and structural similarity between purified samples (after PGD+EOT attack) and original clean images, using standard metrics including MSE, SSIM, and LPIPS on both CIFAR-10 and ImageNet-100. Results in Tab. \ref{tab: comparison_on_img_similarity_scaled} show that CMAP achieves competitive or superior fidelity compared to leading baselines such as DiffPure and GNSP: on ImageNet-100, CMAP attains a notably higher SSIM score ($0.847$) than the best baseline ($0.784$), indicating stronger preservation of structural integrity, while on CIFAR-10, its slightly higher MSE and LPIPS values reflect an incidental denoising effect aligned with its manifold-aware design. The high fidelity of CMAP stems from its manifold-aware optimization in the latent space. Unlike methods that operate directly in pixel space, our joint perceptual consistency restoration and latent distribution consistency constraint ensure the generated sample is semantically anchored to the clean data manifold while closely matching the input, thereby inherently preserving high-level features. This demonstrates that CMAP effectively removes adversarial perturbations while maintaining the visual quality necessary for broader applications.}

 {To complement our quantitative fidelity analysis, we provide comprehensive visual comparisons of original clean images, their adversarial counterparts (generated via PGD+EOT attacks), and purified outputs from CMAP and key baselines (Fig. \ref{fig: cifar_part1}, \ref{fig: cifar_part2}, \ref{fig: imagenet_part1}, \ref{fig: imagenet_part2}, \ref{fig:comparison_part_cifar_high}). The results consistently show that CMAP effectively removes adversarial perturbations while recovering the essential visual content and structural details of the original clean image. On ImageNet-100, where natural image noise is relatively low, CMAP’s outputs are visually cleaner and more coherent than those of other methods, which often leave residual noise or introduce blurring/artifacts. On CIFAR-10, CMAP also incidentally smooths inherent natural noise in the original images, resulting in perceptually cleaner outputs—an observation that aligns with the analysis in Tab. \ref{tab: comparison_on_img_similarity_scaled}, where slightly elevated MSE/LPIPS values on CIFAR-10 reflect this denoising effect, while superior SSIM scores confirm better structural preservation. Together, the quantitative and qualitative evidence suggest that CMAP achieves a more faithful restoration of the clean data manifold, producing purified samples of high visual quality suitable for a wide range of downstream tasks.}

\begin{figure}[t]
    \begin{center}
        \subfloat[Mean]
        {\includegraphics[width=0.49\linewidth]{ 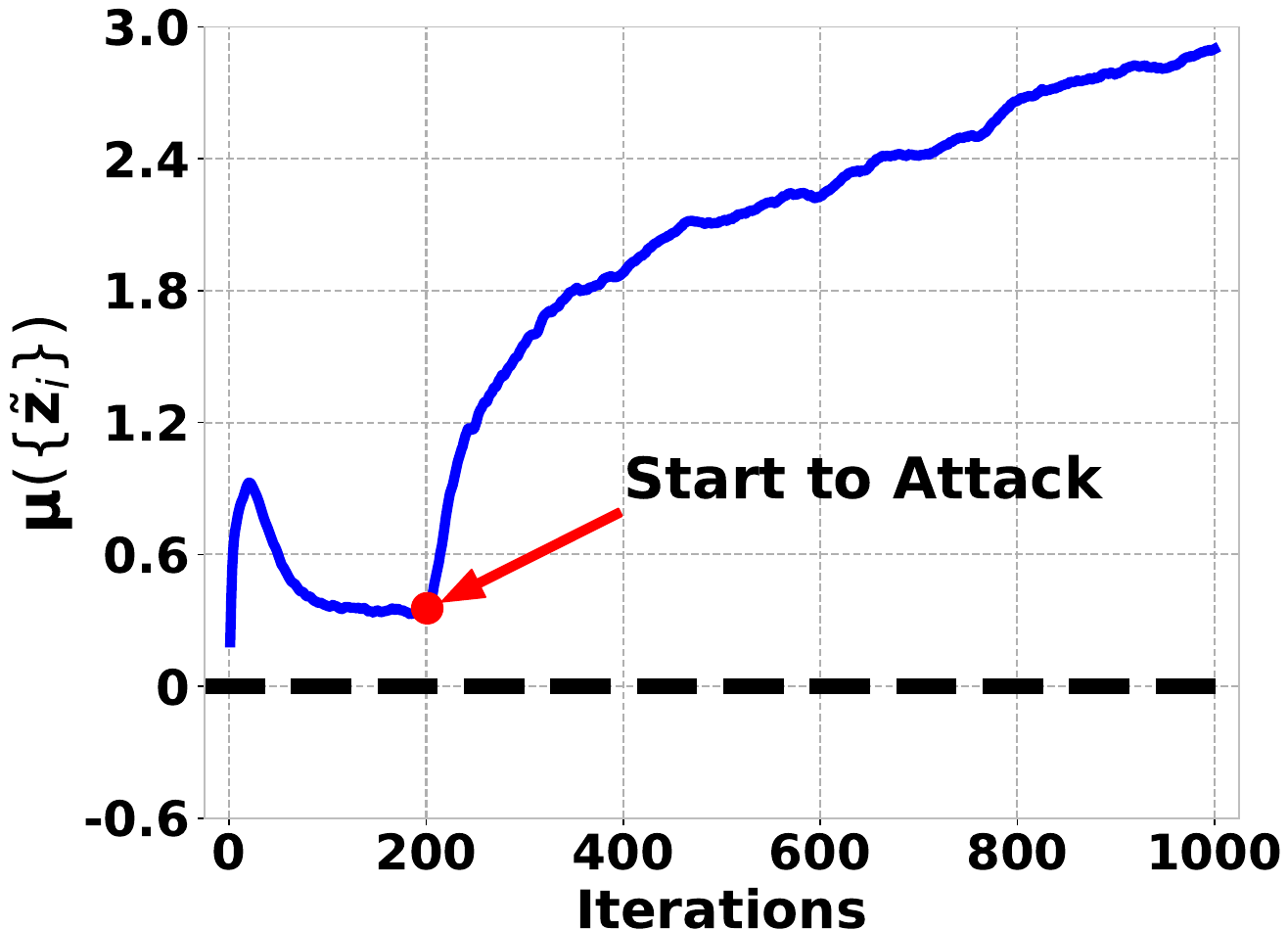}}
        \subfloat[Standard Deviation]
        {\includegraphics[width=0.48\linewidth]{ 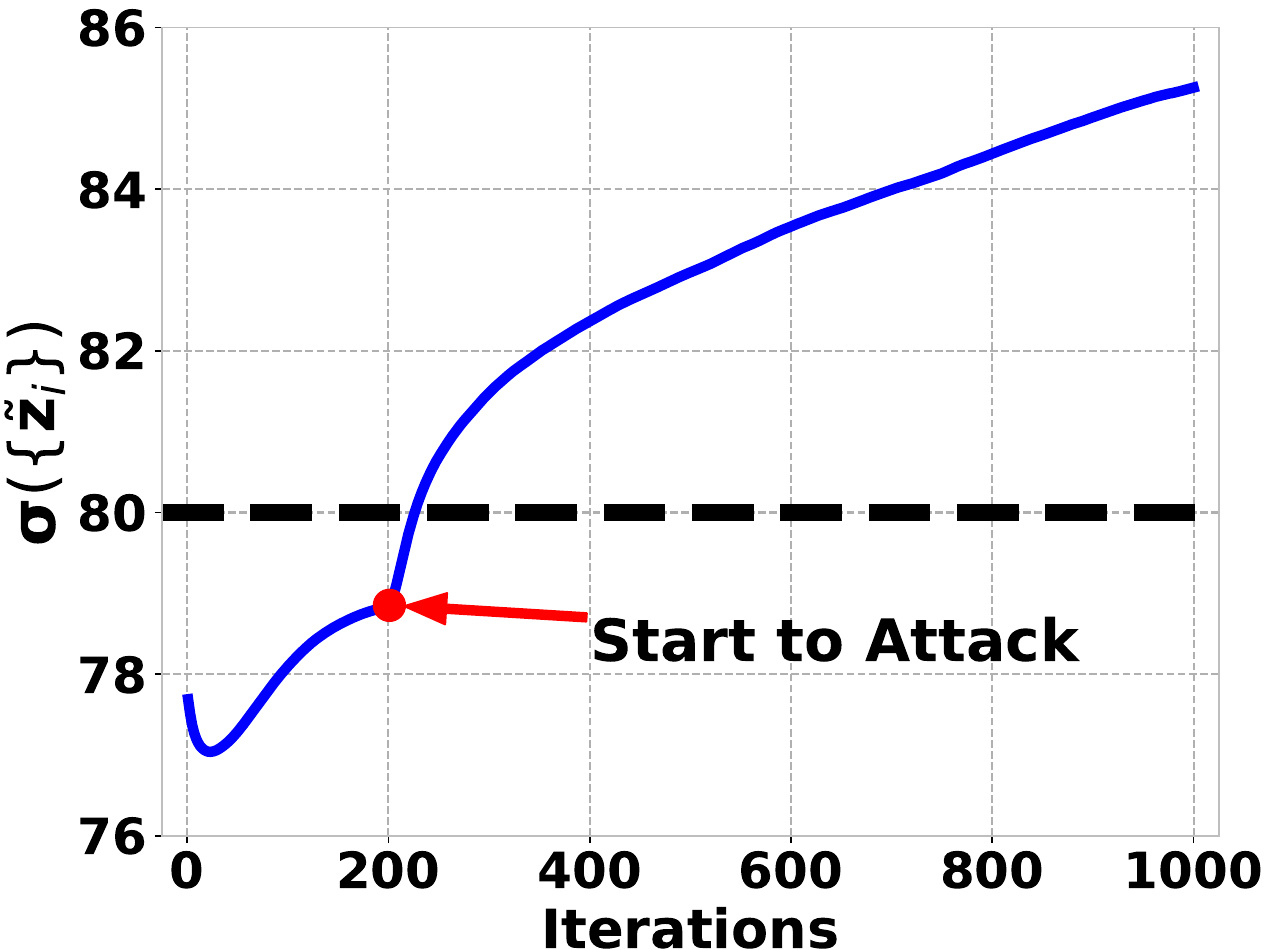}}
        \caption{Mean and standard deviation of optimized latent vectors by consistency-disruption attack ($\epsilon=8/255$) across optimization iterations on CIFAR-10, where $\mathbf{\mu}=0$ and $\sigma=80$ are the mean and variance of the latent distribution.              }
        \label{fig: latent_attack}
        \vspace{-1.7em}
    \end{center}
\end{figure}

\begin{table}[t]
    \caption{Robust accuracy against the consistency-disruption attack under different hyper-parameter $\lambda$. $\lambda$ is chosen to achieve the lowest accuracy (i.e., the highest attack intensity).}
    \vspace{-0.5em}
    \renewcommand\arraystretch{1.3}
    \begin{center}
    \begin{threeparttable}
    \resizebox{\linewidth}{!}{
    \begin{tabular}{cccccccc}
    \toprule
    % Dataset &  Intensity & ~ & ~ & ~ \\ \midrule
    \multirow{4}{*}{\rotatebox{90}{CIFAR-10}} & \multirow{2}{*}{\makecell{$\ell_{\infty}$ \\ ($8/255$)}} & $\lambda$ & $5 \times 10^{5}$ & $1 \times 10^{6}$ & $2 \times 10^{6}$ & $5 \times 10^{6}$ & $1 \times 10^{7}$ \\
    ~ & ~ & Acc & $71.00$ & $69.00$ & $68.00$ & $68.00$ & $69.00$ \\ \cline{2-8}
    ~ & \multirow{2}{*}{\makecell{$\ell_{2}$ \\ ($0.5$)}} & $\lambda$ & $2 \times 10^{8}$ & $5 \times 10^{8}$ & $1 \times 10^{9}$ & $2 \times 10^{9}$ & $5 \times 10^{9}$ \\
    ~ & ~ & Acc & $73.00$ & $73.00$ & $73.00$ & $75.00$ & $73.00$ \\ \midrule
    \multirow{4}{*}{\rotatebox{90}{ImageNet-100}} & \multirow{2}{*}{\makecell{$\ell_{\infty}$ \\ ($4/255$)}} & $\lambda$ & $1$ & $2$ & $5$ & $10$ & $20$ \\
    ~ & ~ & Acc & $51.00$ & $47.00$ & $47.00$ & $48.00$ & $47.00$ \\ \cline{2-8}
    ~ & \multirow{2}{*}{\makecell{$\ell_{2}$ \\ ($0.5$)}} & $\lambda$ & $1 \times 10^{4}$ & $2 \times 10^{4}$ & $5 \times 10^{4}$ & $1 \times 10^{5}$ & $2 \times 10^{5}$ \\
    ~ & ~ & Acc & $54.00$ & $54.00$ & $53.00$ & $54.00$ & $55.00$ \\  \bottomrule
    \end{tabular}
    }
    \end{threeparttable}
    \end{center}
    \label{tab: ataptive_attack_ablation}
    \vspace{-1.7em}
\end{table}

\subsection{Defense against Consistency-Disruption Attack}
\label{sec: exper on adaptive attack}

To further evaluate the robustness of our \mymethod, we apply the consistency-disruption attack in Sec. \ref{sec: Adtaptive Attack} to our \mymethod. This attack targets all components of \mymethod~and generates $K$ adversarial samples for each clean sample, aligned with the procedure of \mymethod. The attack is successful if any of these samples lead to a misclassification after performing our defense, and $\lambda$ is set to yield the lowest purification accuracy (see Tab. \ref{tab: ataptive_attack_ablation}). For this evaluation,  
we report the robust accuracy across optimization iterations. As shown in Fig. \ref{fig: adaptive_attack_accuracy}, our \mymethod~still maintains strong robust accuracy throughout the attack process across both datasets and against $\ell_2$ or $\ell_\infty$ attacks.

To explore the underlying reason for successful purification by our \mymethod, we report 
the mean and variance of latent vectors over iterations, as depicted in Fig. \ref{fig: latent_attack}. As the attack starts, the distribution of the $K$ latent vectors rapidly deviates from the latent distribution of the pre-trained consistency model. However, this deviation is effectively corrected by our latent distribution consistency constraint mechanism, guiding the generated samples back toward the clean data manifold.

\subsection{Ablation Studies}\label{sec: Ablation}
The preceding results show the satisfactory robustness of our \mymethod~against both white-box attacks and consistency-disruption attack.  {This stems from four key components: 1) the choice of generative model, lying the foundation of the latent space structure for purification;} 
2) the perceptual consistency restoration, capturing fine textures and details of the test sample; 3) the latent distribution consistency constraint, effectively avoiding fitting adversarial perturbations; 4) the latent vector consistency prediction, aggregating multiple generated samples for more reliable classification outcomes. Next, we validate the effectiveness of each related component.

 {\textbf{Impact of different generative paradigms.} CMAP’s purification pipeline relies on optimizing latent vectors within a generative model’s latent space, which requires two core properties: \textit{smoothness} (continuous latent changes yield semantically coherent image changes) and \textit{semantic coverage} (latent directions capture natural data variations). Diffusion/consistency models and VAEs inherently satisfy these criteria due to their continuous probabilistic latent spaces, enabling stable gradient-based optimization toward the clean data manifold \cite{kingma2013auto, higgins2017beta}. In contrast, off-the-shelf GANs often lack a smooth, disentangled latent geometry, making direct inversion and optimization unstable and leading to poor purification fidelity \cite{xia2022gan}.}

 {We evaluate CMAP with Projected-GAN \cite{sauer2021projected} and VAE variants \cite{esser2021taming, peebles2023scalable}) on CIFAR-10 and ImageNet-100. Since GAN-generated adversarial samples (via GAN+classifier pipeline) fail to effectively attack the target model (low attack success rate $<5\%$), we use classifier-only generated adversarial samples for evaluation. As in Tab. \ref{tab: comparison_generative_cifar}, GAN-based purification degrades standard accuracy to $45.80\%$ on CIFAR-10 due to unstable latent inversion, despite maintaining non-trivial robust accuracy. This result empirically confirms our analysis: while advanced GANs can generate high-quality samples, their latent spaces are not necessarily structured to support stable \textit{inversion via direct optimization}, leading to poor reconstruction fidelity during purification (see Fig. \ref{fig:comparison_part3}). Here, we do not demonstrate the results on some SOTA GANs like StyleGAN-XL \cite{sauer2022stylegan} due to their intrinsic difficulty of jointly optimizing in the complex latent space of multi-style vectors, which introduces non-trivial dependencies between latent components.} 

 {For VAE generative paradigm, we train a VAE on CIFAR-10 using the Taming Transformers framework\footnote{\scriptsize\url{https://github.com/CompVis/taming-transformers}} \cite{esser2021taming} and employ a pre-trained VAE from the DiET framework\footnote{\scriptsize\url{https://github.com/facebookresearch/DiT}} \cite{peebles2023scalable} for ImageNet-100. The results in Tab. \ref{tab: comparison_generative_cifar} and \ref{tab: comparison_generative_imagenet} shows that VAE-based CMAP achieves strong performance. On CIFAR-10, it achieves a high standard accuracy of $92.80\%$ and robust accuracy competitive with baselines. The advantage is even more pronounced on ImageNet-100, where CMAP(VAE) attains a standard accuracy of $68.20\%$ and a robust accuracy of $43.60\%$ under PGD+EOT, outperforming strong baselines like GNSP ($33.40\%$) by a large margin (over $10\%$).   These results validate that our core optimization principle is broadly effective for generative models with well-structured latent spaces.}

\textbf{Impact of the perceptual consistency restoration.} 
Both mean absolute error (MAE) and structure similarity index measure (SSIM) are used to restore test samples from latent vectors,
with a factor $\alpha$ to balance these terms. We demonstrate the effect of $\alpha$ over WideResNet-28-10 on CIFAR-10 in Tab. \ref{tab: impact_of_alpha}.
Notably, as $\alpha$ increases, both standard and adversarial accuracy gradually improve, indicating that introducing the SSIM term during restoration accelerates alignment optimization. However, when $\alpha$ becomes too large, \eg, $\alpha>2$, both robust and clean accuracy start to decline. This may result from an excessive focus on structural similarity at the expense of pixel-wise alignment, reducing the model's ability to generalize effectively to both adversarial and clean samples.

\textbf{Impact of the latent distribution
consistency constraint.}
% Varying the Gaussian factor $\beta$ effectively
The coefficient $\beta$ controls the strength of the latent distribution consistency constraint. To investigate its impact, we plot the standard and robust accuracy across iterations with different $\beta$ in Fig. \ref{fig: ablation_beta}. 
Without the distribution constraint, \ie, $\beta = 0$,
% When no constraint is applied (i.e., $\beta = 0$),
the restoration exhibits a significant decline in robust accuracy after reaching a peak, especially for PGD-$\ell_\infty$+EOT. In contrast, it remains stable throughout the optimization with $\beta=5 \times 10^{-4}$. Additionally, an excessively strong distribution constraint can impair the restoration of the test sample, as evidenced by reduced peak performance at $\beta=5 \times 10^{-3}$. More quantitative results in Tab. \ref{tab: impact_of_beta} further support these observations. Thus, we set $\beta = 5 \times 10^{-4}$ that simultaneously suppresses adversarial perturbations while maintaining effective image restoration.

\begin{figure}[t]
    \begin{center}
        \subfloat[Mean]
        {\includegraphics[width=0.49\linewidth]{ 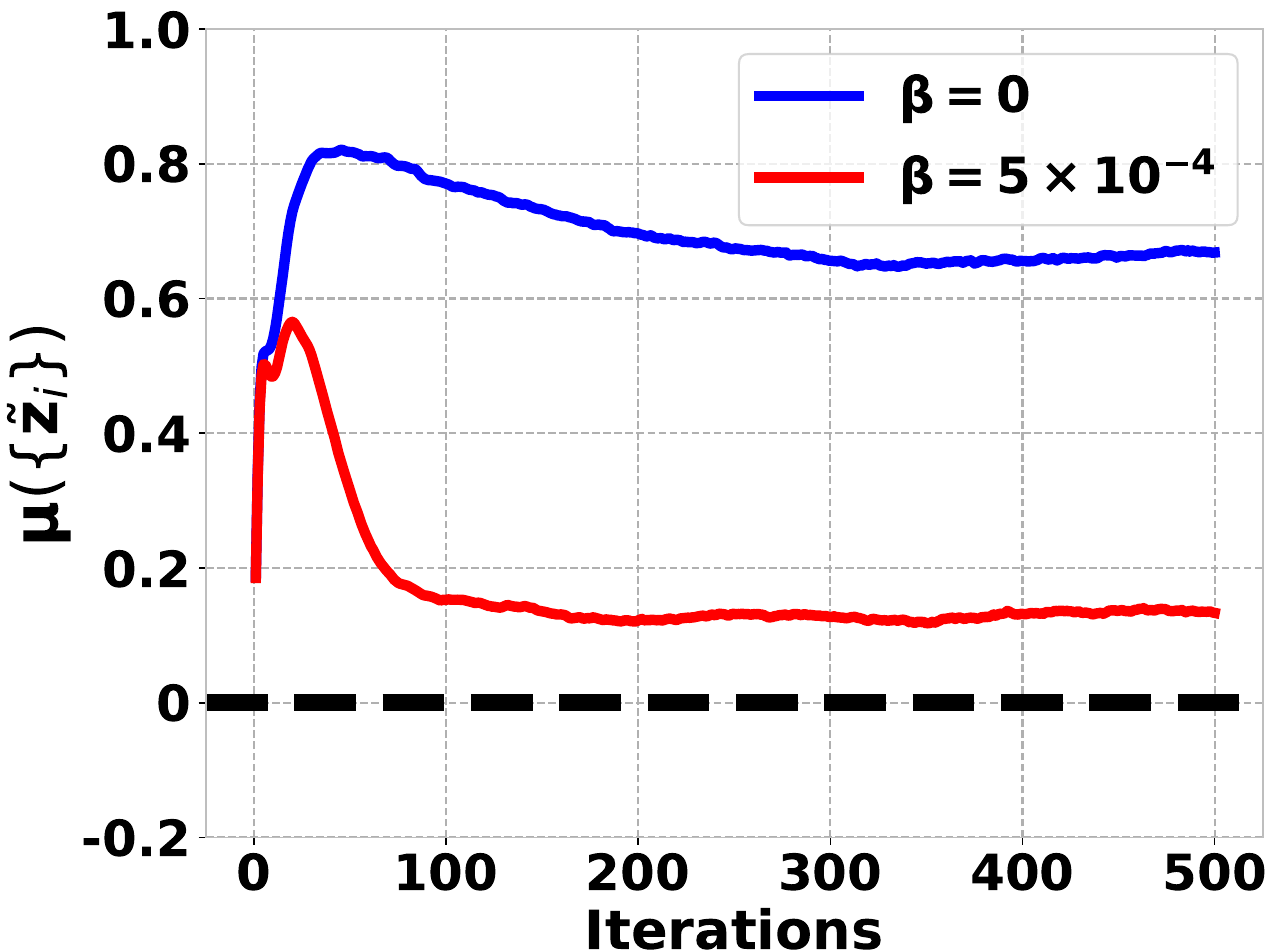}}
        \subfloat[Standard Deviation]
        {\includegraphics[width=0.48\linewidth]{ 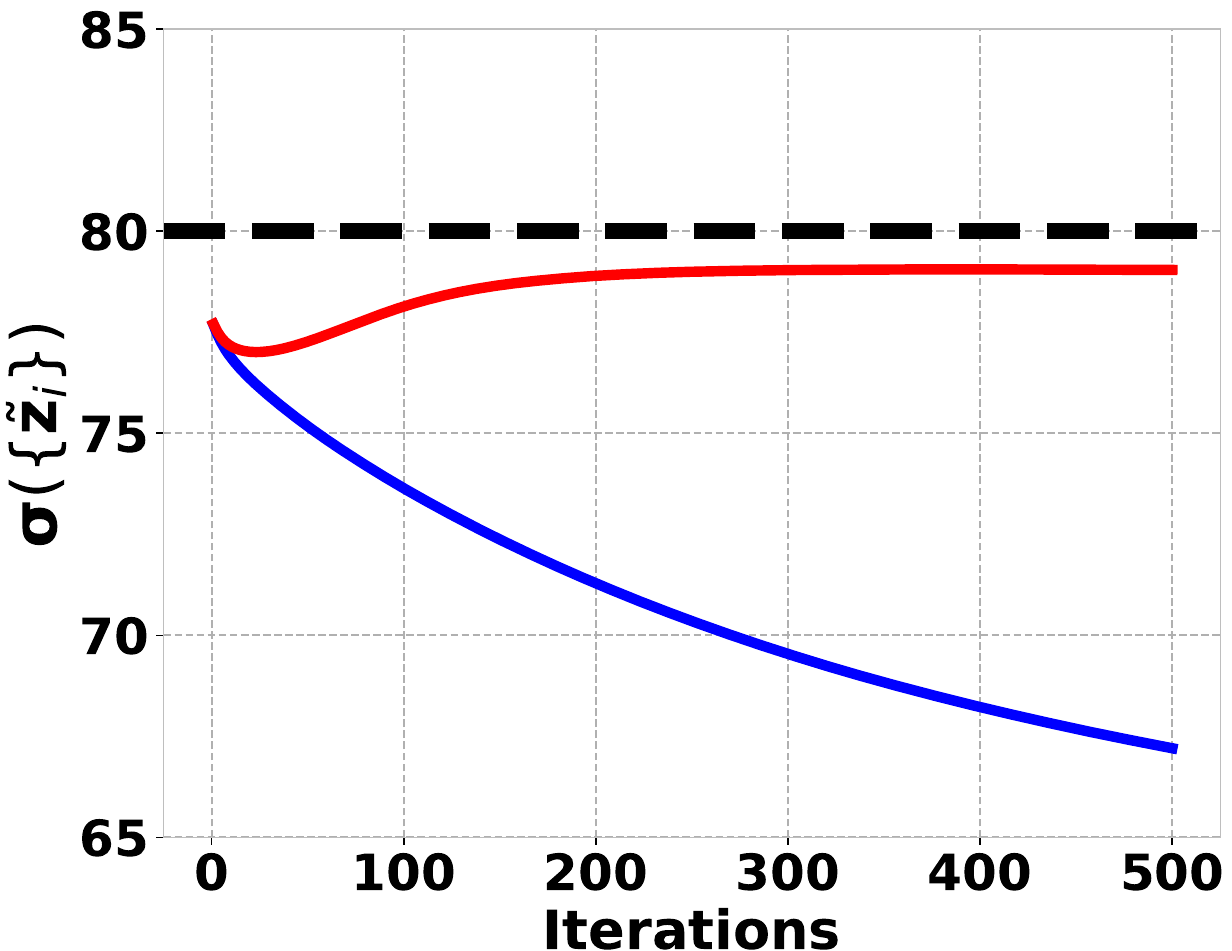}}
        \caption{Mean and standard deviation of optimized latent vectors by our \mymethod~in Alg. \ref{alg: odepure with aug} across optimization iterations under different $\beta$ on CIFAR-10 against PGD+EOT attack with $\epsilon=8/255$.}
        \label{fig: latent_purification}
        \vspace{-1.5em}
    \end{center}
\end{figure}

Furthermore, to delve into the mechanism behind the latent distribution consistency constraint, we examine its effect on latent vectors during optimization. To this end, we depict the mean and standard deviation of the latent vectors throughout the optimization process. As illustrated in Fig. \ref{fig: latent_purification}, without the distribution constraint, \ie, $\beta = 0$, the mean and standard deviation of the optimized latent vectors deviate significantly from the latent distribution, where the dotted lines represent the referenced statistics \wrt the latent distribution.

\begin{table}[t]
    \caption{Standard and robust accuracy against PGD+EOT and AutoAttack with $\ell_{\infty}$-norm on CIFAR-10 ($\epsilon=8/255$) and ImageNet-100 ($\epsilon=4/255$). The latent vector consistency prediction enhances both standard and robust accuracy.}
    \vspace{-0.5em}
    \label{tab: no_voting}
    \renewcommand\arraystretch{1.3}
    \begin{center}
    \begin{threeparttable}
    \resizebox{\linewidth}{!}{
    \begin{tabular}{cclccc}
    \toprule
         Dataset & Classifier & Method & Standard & PGD+EOT & AutoAttack \\ \midrule
         \multirow{4}{*}{CIFAR-10} & \multirow{2}{*}{WRN-28-10} & \mymethod~w/o Vote & $83.07_{\pm 0.50}$ & $69.87_{\pm 2.14}$ & $72.07_{\pm 2.00}$ \\
         ~ & ~ & \mymethod~w/ Vote & $\mathbf{88.73_{\pm 0.50}}$ & $\mathbf{74.60_{\pm 1.59}}$ & $\mathbf{78.67_{\pm 1.90}}$ \\
         ~ & \multirow{2}{*}{WRN-70-16} & \mymethod~w/o Vote & $82.44_{\pm 0.92}$ & $70.66_{\pm 1.14}$ & $75.23_{\pm 0.42}$ \\
         ~ & ~ & \mymethod~w/ Vote & $\mathbf{88.27_{\pm 0.50}}$ & $\mathbf{74.80_{\pm 1.56}}$ & $\mathbf{81.33_{\pm 0.31}}$ \\ \midrule
        \multirow{2}{*}{ImageNet-100} & \multirow{2}{*}{ResNet50} & \mymethod~w/o Vote & $58.72_{\pm 1.71}$ & $38.20_{\pm 1.35}$ & $-$ \\
         ~ & ~ & \mymethod~w/ Vote & $\mathbf{61.53_{\pm 1.45}}$ & $\mathbf{39.27_{\pm 1.79}}$ & $-$ \\ \bottomrule
    \end{tabular}
    }
    \end{threeparttable}
    \end{center}
    \vspace{-1em}
\end{table}

\begin{table}[t]
    \caption{ \blue{Comparisons in terms of performance, per-sample inference time and GPU memory, on CIFAR-10 over WideResNet-28-10 ($\epsilon=8/255$) and ImageNet-100 over ResNet50 ($\epsilon=4/255$).}}
    \vspace{-0.5em}
    \begin{center}
    \begin{threeparttable}
    \resizebox{\linewidth}{!}{
    {
    \begin{tabular}{clcccc}
    \toprule
      & Method & Standard (\%) $\uparrow$ & PGD+EOT (\%) $\uparrow$ & VRAM (MiB) $\downarrow$ & Inf. Time (s) $\downarrow$ \\ \midrule
     \multirow{7}{*}{\rotatebox{90}{CIFAR-10}} & {DiffPure} \cite{nie2022diffusion} & $90.27_{\pm 0.81}$ & $48.27_{\pm 1.86}$ & $3502$ & $7.44$ \\
     & {GNSP} \cite{lee2023robust} & $90.40_{\pm 1.40}$ & $55.87_{\pm 0.50}$ & $4206$ & $26.61$ \\ 
     & {FreqPure} \cite{pei2025diffusion} & $93.96_{{\pm}0.09}$ & $45.00_{{\pm}0.59}$ & $5912$ & $90.99$ \\
     & CMAP ($K=2$) & $80.13_{\pm 1.17}$ & $69.67_{\pm 2.19}$ & $2024$ & $18.98$ \\
     & CMAP ($K=5$) & $87.27_{\pm 1.10}$ & $73.67_{\pm 1.40}$ & $2046$ & $25.83$ \\
     & CMAP ($K=10$) & $88.73_{\pm 0.50}$ & $74.60_{\pm 1.59}$ & $2692$ & $35.39$ \\
     & CMAP ($K=20$) & $89.27_{\pm 0.42}$ & $73.40_{\pm 1.64}$ & $4290$ & $60.47$ \\ \midrule
     \multirow{6}{*}{\rotatebox{90}{ImageNet-100}} & {DiffPure} \cite{nie2022diffusion} & $58.40_{\pm 1.10}$ & $28.57_{\pm 1.82}$ & $3776$ & $40.86$ \\
     & {GNSP} \cite{lee2023robust} & $59.92_{\pm 1.24}$ & $33.40_{\pm 1.44}$ & $4010$ & $170.57$ \\ 
     & {FreqPure} \cite{pei2025diffusion} & $56.36_{\pm 1.02}$ & $30.83_{\pm 1.59}$ & $3358$ & $181.45$ \\
     & CMAP ($K=2$) & $47.84_{\pm 2.64}$ & $38.62_{\pm 1.27}$ & $6168$ & $46.27$ \\
     & CMAP ($K=5$) & $61.93_{\pm 1.86}$ & $39.87_{\pm 2.58}$ & $6220$ & $63.15$ \\
     & CMAP ($K=10$) & $60.85_{\pm 0.97}$ & $40.71_{\pm 1.71}$ & $6458$ & $107.01$ \\
     \bottomrule
    \end{tabular}
    }}
    \end{threeparttable}
    \end{center}
    \label{tab: efficient}
    \vspace{-1.8em}
\end{table}

\textbf{Impact of the number of latent vectors  $K$.}
The effectiveness of the latent distribution consistency constraint is also influenced by the number of latent vectors initialized for optimization, denoted as $K$. Intuitively, increasing $K$ reduces the bias in the estimation of mean and variance, thereby reinforcing the distribution constraint and leading to improved optimization outcomes. However, a larger $K$ comes with a linearly increasing computational cost. 
As shown in Tab. \ref{tab: impact_of_k}, both standard and robust accuracy generally improve with larger $K$. Nevertheless, when $K=20$, robust accuracy exhibits a slight decline. This issue can be alleviated by further adjusting $\beta$. However, to adhere to the principle of controlling for a single variable in ablation studies, we maintain consistent parameter settings in the reported results. 
Taking both performance and computational efficiency into account, we select $K=10$ for CIFAR-10 and $K=5$ for ImageNet-100.

\textbf{Impact of the latent vector consistency prediction.}
We ablate the impact of the latent vector consistency prediction by randomly selecting one of the $K$ samples for prediction. As shown in Tab. \ref{tab: no_voting}, this mechanism improves standard and robust accuracy by $4.14\%$ to $6.60\%$ on CIFAR-10, and by $1.07\%$ to $2.81\%$ on ImageNet-100, respectively. Notably, the performance without voting remains at an acceptable level (surpassing the DiffPure \cite{nie2022diffusion}), verifying the effectiveness of the entire purification process. Nevertheless, since our method already optimizes $K$ latent vectors, leveraging this information through voting is a natural and effective choice.

\subsection{{Discussion on Practical Applicability}}

\blue{To facilitate real-world deployment, we provide analysis on the efficiency, scalability, and deployment scope of CMAP.}

\blue{\textbf{Efficiency and tunable trade-off.} Table \ref{tab: efficient} reports per-sample inference time and GPU memory of CMAP (with different \(K\)) against baselines on both CIFAR-10 and ImageNet-100. On CIFAR-10, CMAP with \(K=5\) achieves $73.67\%$ robust accuracy with $25.83$s inference time and $2,046$ MiB memory, substantially outperforming GNSP ($55.87\%$, $26.61$s, $4,206$ MiB) and FreqPure ($45.00\%$, $90.99$s, $5,912$ MiB). On ImageNet-100, CMAP (\(K=5\)) achieves $39.87\%$ robust accuracy, surpassing DiffPure ($28.57\%$), GNSP ($33.40\%$), and FreqPure ($30.83\%$), while its inference time ($63.15$s) is significantly lower than GNSP ($170.57$s) and FreqPure ($181.45s$), though with higher GPU memory usage. These results show that CMAP offers a tunable robustness–efficiency trade-off controlled by \(K\): moderate \(K\) already delivers strong robustness with competitive efficiency.}

\blue{\textbf{Scalability to high-resolution data.} Our current experiments are on 64×64 resolution due to the lack of publicly available pre-trained consistency models for higher-resolution unconditional generation. Nevertheless, consistency models have been successfully extended to high-resolution generation \cite{song2023consistency}, and our framework is \textit{generator-agnostic}, naturally applicable to any pre-trained consistency model regardless of resolution. Table \ref{tab: efficient} also provides a guideline: increasing \(K\) yields diminishing returns, \eg, on ImageNet-100, raising \(K\) from 5 to 10 only slightly improves robust accuracy ($39.87\% \to 40.71\%$), but more than doubles GPU memory usage. In practice, one may adopt an adaptive strategy, \eg, starting with a small \(K\) and increasing it only when predictions from different latent vectors are inconsistent or the input is flagged as suspicious.}

\blue{\textbf{Deployment scope.} CMAP is particularly suitable for safety-critical or high-stakes scenarios, such as medical diagnosis, remote sensing, industrial inspection, security screening, forensic analysis, and pre-deployment robustness auditing, where reliable, attack-agnostic purification outweighs minimal latency. Moreover, existing input-space purification methods suppress perturbations via shallow operations (denoising, frequency filtering, or diffusion reversal) that do not explicitly enforce manifold consistency. In contrast, CMAP performs principled, manifold-consistent restoration through latent-space optimization,  contributing to its strong attack-agnostic robustness.}

\section{Conclusion}
In this paper, we reveal that samples from a well-trained generative model are close to clean ones but far from adversarial ones. Leveraging this, we propose a Consistency Model-based Adversarial Purification (\mymethod) method. By integrating a perceptual consistency restoration mechanism and a latent distribution consistency constraint strategy, our \mymethod~aligns the generated samples with the clean data manifold while preserving essential high-level perceptual features. Additionally, the latent vector consistency prediction scheme improves the stability and reliability of the final predictions. We also apply a consistency-disruption attack that takes into account the overall defense mechanism, aiming to exploit potential vulnerabilities. Extensive experiments on CIFAR-10 and ImageNet-100 across various classifier architectures such as ResNet and WideResNet demonstrate that our method achieves superior performance in robust and standard accuracy under diverse attack scenarios.

\vspace{-0.1in}
\ifCLASSOPTIONcompsoc
  % The Computer Society usually uses the plural form
  \section*{Acknowledgments}
\else
  % regular IEEE prefers the singular form
  \section*{Acknowledgment}
\fi

This work was partially supported by the Joint Funds of
the National Natural Science Foundation of China (Grant
No.U24A20327), Guangdong S\&T Program under Grant
2026B0101110001, Key-Area Research and Development Program Guangdong Province 2018B010107001, and TCL Science and Technology Innovation Fund, China. Jiahao Yang was supported by the China Scholarship Council (CSC) under Grant No. 202506150018. Bo Han was supported by RGC YCRG No. C2005-24Y, RGC GRF No. 12200725,  NSFC AIMRP No. 92570109, and NSFC GP No. 62376235.

%-------------------------------------------------------------------------

\ifCLASSOPTIONcaptionsoff
  \newpage
\fi
\bibliographystyle{ieeetr}
{
	\bibliography{longstrings,reference}
}
\vspace{-1.4cm}
% \clearpage
\begin{IEEEbiography}[{\includegraphics[width=1in,height=1.25in,clip,keepaspectratio]{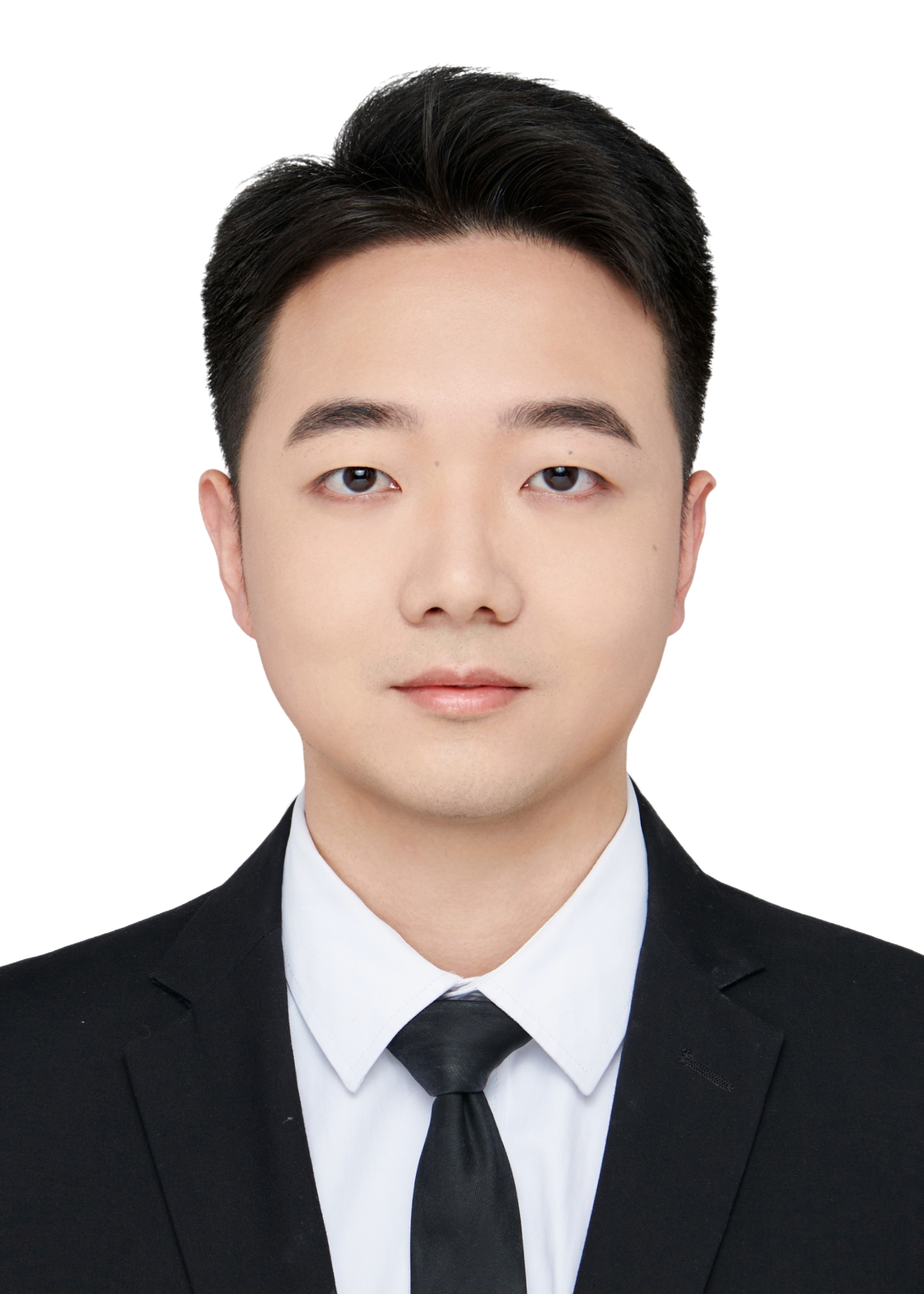}}]{Shuhai Zhang}
is currently a Ph.D. candidate at South China University of Technology in Guangzhou, China. His research interests are broadly in machine learning and mainly focus on adversarial robustness and AI generation detection. He has published papers in IEEE TIP, Neural Networks, T-CSVT, NeurIPS, ICCV, ICML, ICLR, CVPR.
\end{IEEEbiography}

\vspace{-13mm}

\begin{IEEEbiography}[{\includegraphics[width=1in,height=1.25in,clip,keepaspectratio]{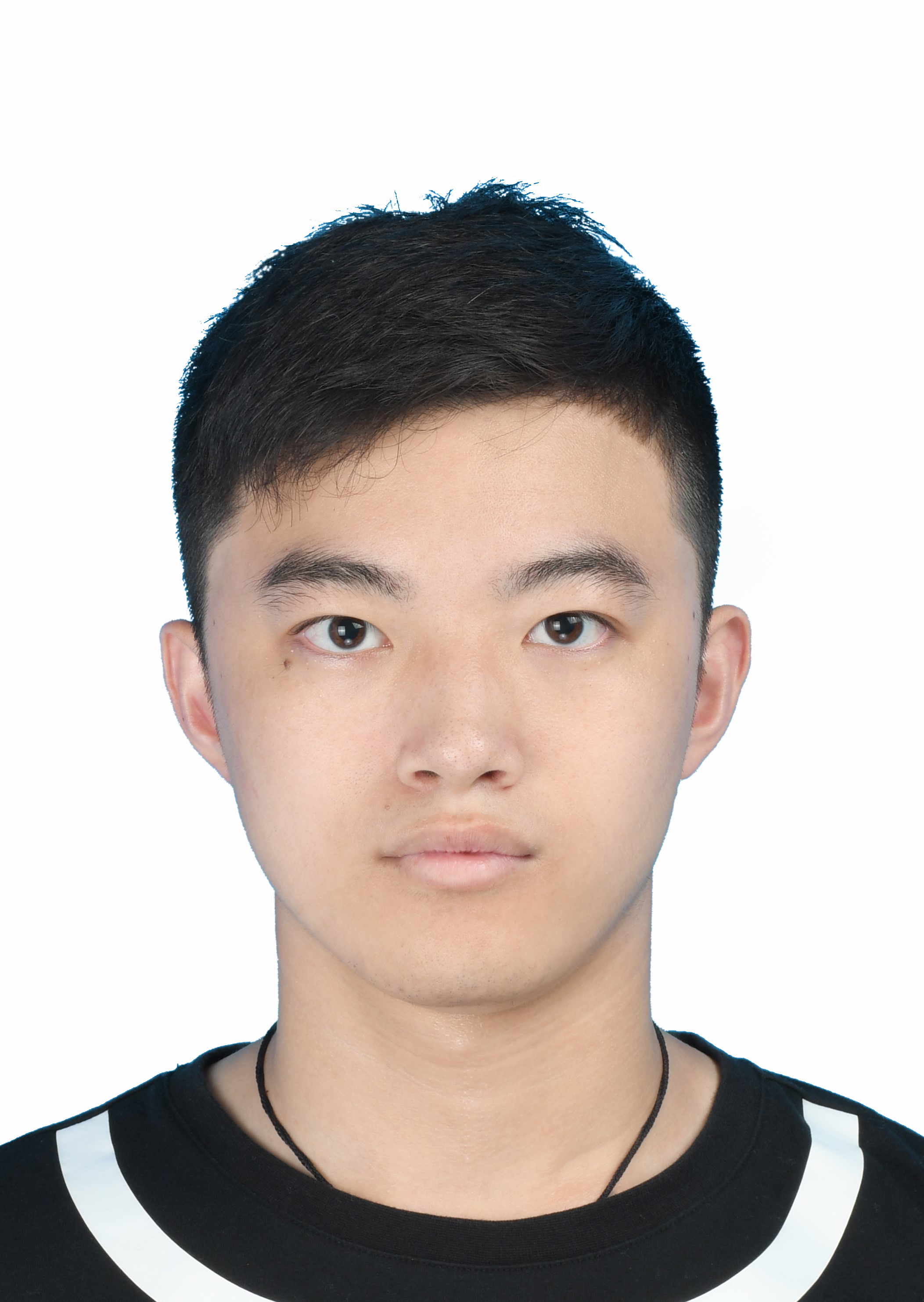}}]{Jiahao Yang}
is currently a Ph.D. candidate in the School of Computing and Information Systems at the University of Melbourne, Australia. He received his M.Eng. in Software Engineering and his B.Sc. in Mathematics from South China University of Technology, Guangzhou, China. His research interests broadly span machine learning, with a primary focus on trustworthy machine learning, including aspects such as robustness, fairness, and interpretability.
\end{IEEEbiography}

\vspace{-14mm}

\begin{IEEEbiography}[{\includegraphics[width=1in,height=1.25in,clip,keepaspectratio]{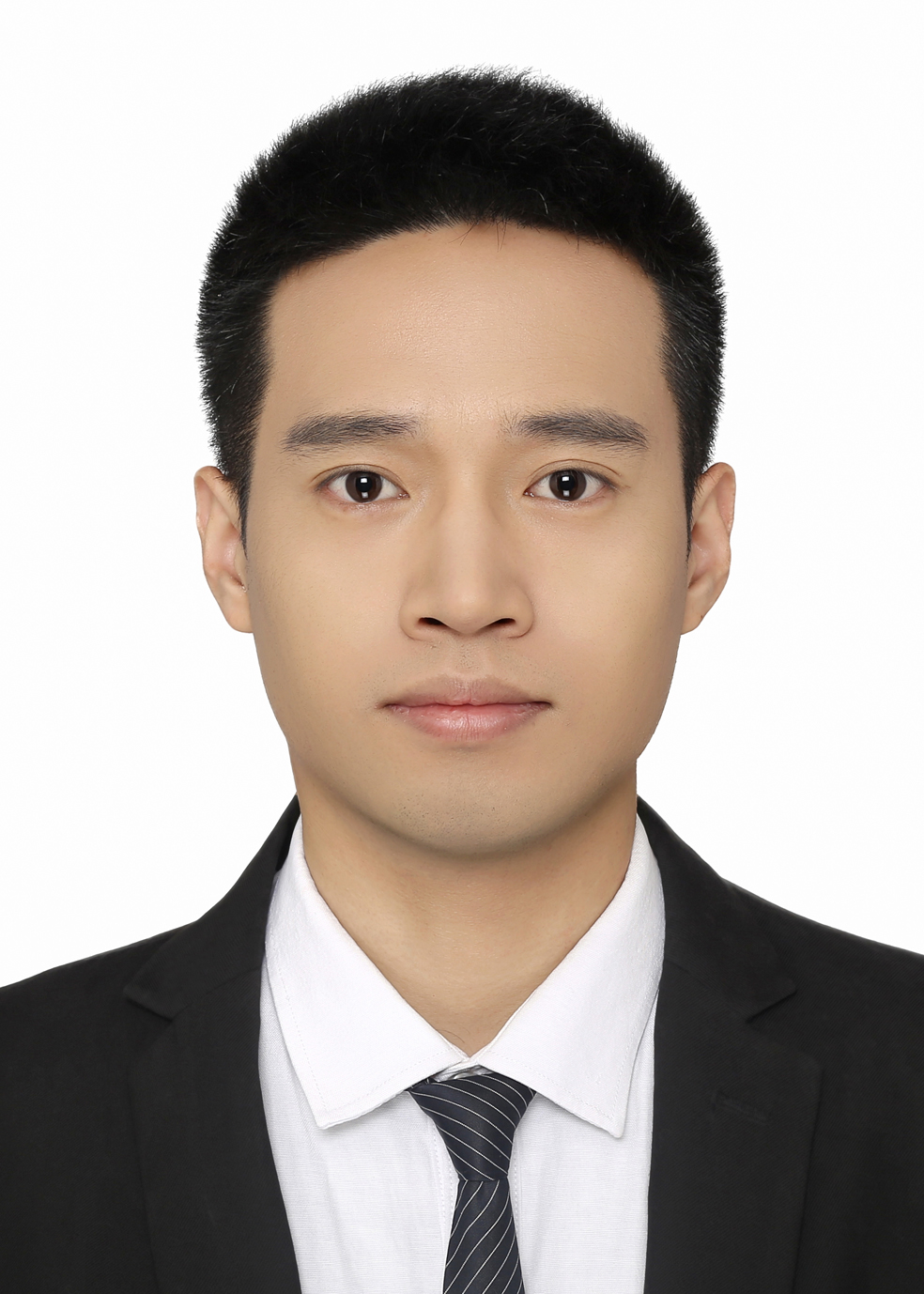}}]{Hui Luo}
is currently pursuing a Ph.D. in signal and information processing at the University of Chinese Academy of Sciences in Beijing, China. During his Ph.D. program, he also conducts research at the Institute of Optics and Electronics, Chinese Academy of Sciences, located in Chengdu, China. His research interests include model compression and acceleration, and the development of robust and reliable models.
\end{IEEEbiography}

\vspace{-14mm}

\begin{IEEEbiography}[{\includegraphics[width=0.9in,height=1.25in,clip,keepaspectratio]{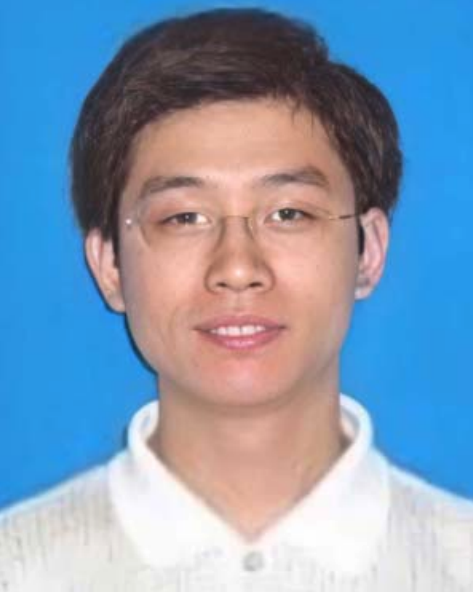}}]{Jie Chen}
(Member, IEEE) received the MSc and PhD degrees from the Harbin Institute of Technology, China, in 2002 and 2007, respectively. 
He joined the Shenzhen Graduate School, Peking University, as a faculty member in 2019 and is currently an associate professor with the School of Electronic and Computer Engineering, Peking University.
Since 2018, he has been working with the Peng Cheng Laboratory, China. From 2007 to 2018, he worked as a senior researcher with the Center for Machine Vision and Signal Analysis, University of Oulu, Finland. In 2012 and 2015, he visited the Computer Vision Laboratory, University of Maryland, and School of Electrical and Computer Engineering, Duke University, respectively. He was a co-chair of International Workshops at ACCV, CVPR, ICCV, and ECCV. He was a guest editor of special issues for IEEE Transactions on Pattern Analysis and Machine Intelligence, IJCV, and Neurocomputing. His research interests include deep learning, computer vision, and medical image analysis. He is an associate editor of the Visual Computer.
\end{IEEEbiography}

\vspace{-10mm}

\begin{IEEEbiography}[{\includegraphics[width=1in,height=1.25in,clip,keepaspectratio]{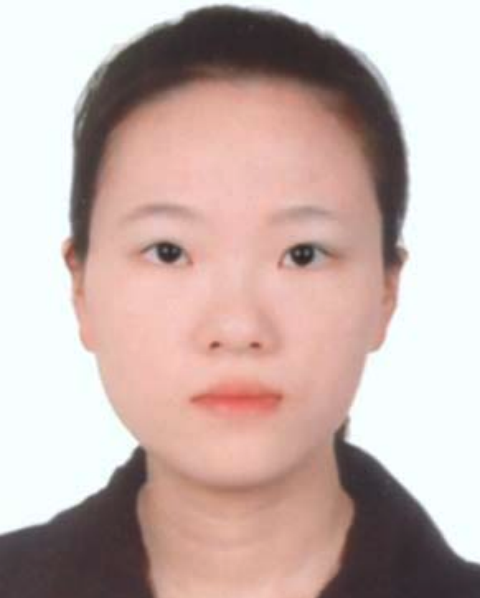}}]{Li Wang}
received the B.S. degree in information and computing science from China University of Mining and Technology, Jiangsu, China, in 2006, the M.S. degree from Xi’an Jiaotong University, Xi’an, Shaanxi, China, in 2009, and the Ph.D. degree from the Department of Mathematics, University of California at San Diego, San Diego, CA, USA, in 2014. She was a Research Assistant Professor with the Department of Mathematics, Statistics, and Computer Science, University of Illinois at Chicago, Chicago, IL, USA, from 2015 to 2017, and a Post-Doctoral Fellow with the University of Victoria, Victoria, BC, Canada, in 2015, and Brown University, Providence, RI, USA, in 2014. She is currently an Assistant Professor with the Department of Mathematics, The University of Texas at Arlington, Texas, USA, and also with the Department of Computer Science and Engineering. Her research interests include large-scale optimization, polynomial optimization, and machine learning.
\end{IEEEbiography}

\vspace{-10mm}

\begin{IEEEbiography}[{\includegraphics[width=1in,height=1.35in,clip,keepaspectratio]{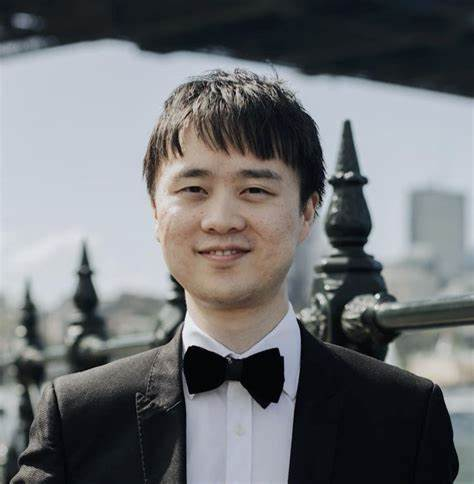}}]{Feng Liu}
(Member, IEEE) received the B.Sc. degree in mathematics and the M.Sc. degree in probability and statistics from the School of Mathematics and Statistics, Lanzhou University, Lanzhou, China, in 2013 and 2013, respectively, and the Ph.D. degree in computer science from the University of Technology Sydney, Sydney, NSW, Australia, in 2020. He is currently a Lecturer with the School of Mathematics and Statistics, Faculty of Science, The University of Melbourne, Melbourne, VIC, Australia. He is also a Visting Fellow of Australian Artificial Intelligence Institute, Faulty of Engineering and Information Technology, University of Technology Sydney. His research interests include hypothesis testing and trustworthy machine learning. He has served as a Senior Program Committee Member for ECAI and a Program Committee Member for NeurIPS, ICML, AISTATS, ICLR, KDD, AAAI, IJCAI, and FUZZ-IEEE. He also served as a reviewer for JMLR, MLJ, TPAMI, TNNLS and TFS. He received the Outstanding Reviewer Awards of ICLR in 2021 and NeurIPS in 2021, the UTS-FEIT HDR Research Excellence Award in 2019, and the Best Student Paper Award of FUZZ-IEEE in 2019.
\end{IEEEbiography}

\vspace{-10mm}

\begin{IEEEbiography}[{\includegraphics[width=1in,height=1.3in,clip,keepaspectratio]{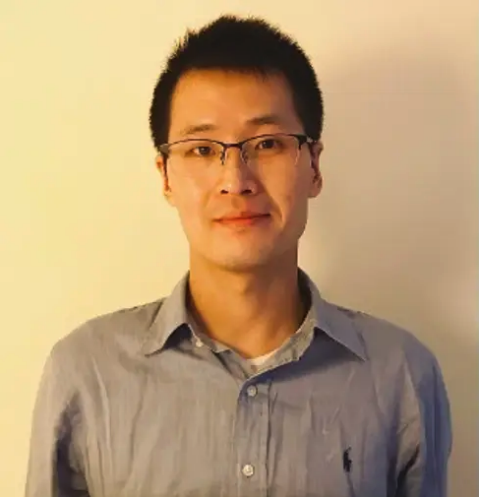}}]{Bo Han}
is currently an Assistant Professor in Machine Learning at Hong Kong Baptist University, and a BAIHO Visiting Scientist of Imperfect Information Learning Team at RIKEN Center for Advanced Intelligence Project (RIKEN AIP), where his research focuses on machine learning, deep learning, foundation models, and their applications. He was a Visiting Research Scholar at MBZUAI MLD (2024), a Visiting Faculty Researcher at Microsoft Research (2022) and Alibaba DAMO Academy (2021), and a Postdoc Fellow at RIKEN AIP (2019-2020). He received his Ph.D. degree in Computer Science from University of Technology Sydney (2015-2019). He has served as Senior Area Chair of NeurIPS, and Area Chairs of NeurIPS, ICML and ICLR. He has also served as Associate Editors of IEEE TPAMI, MLJ and JAIR, and Editorial Board Members of JMLR and MLJ. He received Outstanding Paper Award at NeurIPS, Most Influential Paper at NeurIPS, Notable Area Chair at NeurIPS, Outstanding Area Chair at ICLR, and Outstanding Associate Editor at IEEE TNNLS.
\end{IEEEbiography}

\vspace{-10mm}

\begin{IEEEbiography}[{\includegraphics[width=1.3in,height=1.3in,clip,keepaspectratio]{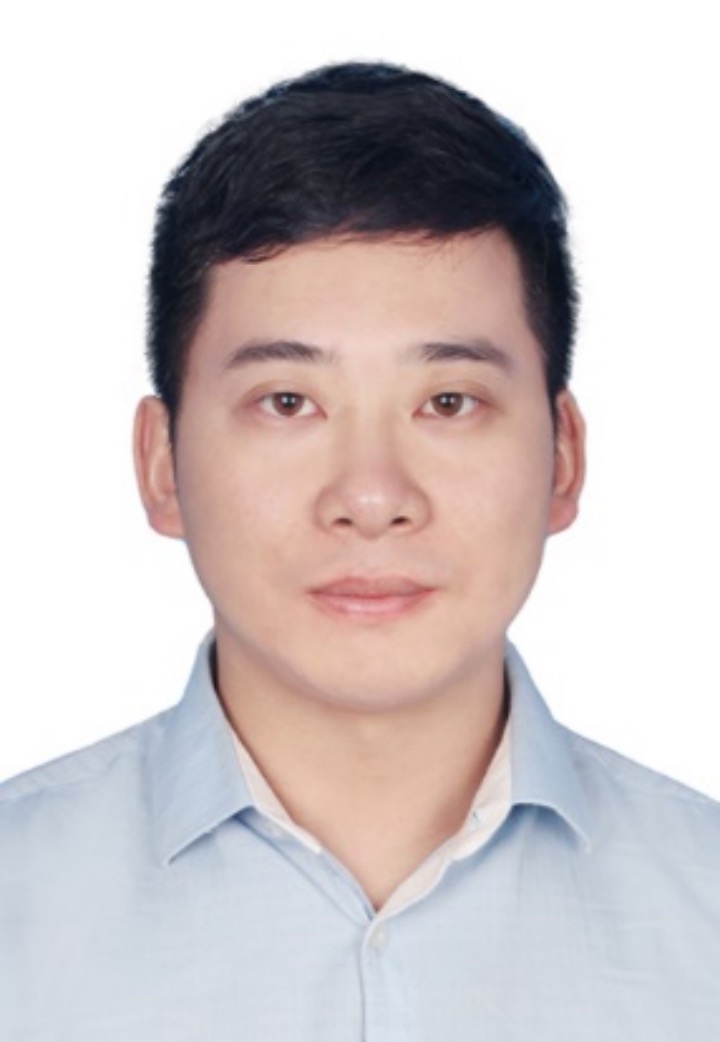}}]{Mingkui Tan}
is currently a Professor with the School of Software Engineering, South China University of Technology, Guangzhou, China. He received the Bachelor Degree in Environmental Science and Engineering in 2006 and the Master Degree in Control Science and Engineering in 2009, both from Hunan University in Changsha, China. He received the Ph.D. degree in Computer Science from Nanyang Technological University, Singapore, in 2014. From 2014-2016, he worked as a Senior Research Associate on computer vision in the School of Computer Science, University of Adelaide, Australia. His research interests include machine learning, sparse analysis, deep learning and large-scale optimization.
\end{IEEEbiography}

\clearpage

\vfill

\setcounter{page}{1}
\setcounter{subsection}{0} % 重置子章节计数器
\renewcommand{\thesubsection}{\thesection.\arabic{subsection}} % 设置为 A.1 格式

\section{Proofs in Section \ref{sec: methods}}

\subsection{Proof of Theorem \ref{thm: adv vs natural}}
\label{app: thm}
\noindent\textbf{Theorem \ref{thm: adv vs natural}}\emph{
% \begin{thm}
% \label{thm: adv vs natural}
    Assuming that the distribution of natural data $p(\bx){=} \mN({\boldsymbol{\mu}}_{\bx}, \sigma_{\bx}^2\mathbf{I})$,  {where $\mathbf{I}$ is an identity matrix}, given a PF ODE sampling $\mathrm{d}\bx=-t \nabla_{\mathbf{x}}\log p_{t}(\mathbf{x})$ with $\mathbf{f}(\bx,t)=\bf{0}$ and $g(t)=\sqrt{2t}$, then for $\forall~\bx \in p(\bx)$ and its adversarial sample $\hat{\bx}=\bx+\boldsymbol{\epsilon}_a$, we have
    \begin{equation}
         \bx_T -\hat{\bx}_T \sim \mN( \mathbf{0}, 2\sigma_{cl}^2\mathbf{I})+\boldsymbol{\mu}_{\epsilon},
    \end{equation}
    where $\boldsymbol{\mu}_\epsilon=\left(\mathbb{E}_t  \frac{ t}{\sigma_{\bx}^2+ t^2}-1\right)\boldsymbol{\epsilon}_a$ and $\sigma_{cl}^2=\mathbb{E}_t  \frac{ t^2}{\sigma_{\bx}^2+ t^2}$.
% \end{thm}
}

\begin{proof}
Starting from the PF ODE sampling equation, we derive the corresponding forward ODE:
\begin{equation}
\label{eqn: ode forward}
    \mathrm{d}\bx=t \nabla_{\mathbf{x}}\log p_{t}(\mathbf{x}).
\end{equation}
We then discretize this equation:
\begin{equation}
\label{eqn: ode forward discrete}
    \bx_{t+\Delta t}=\bx_{t}+ t \nabla_{\mathbf{x}}\log p_{t}(\mathbf{x}) \Delta t.
\end{equation}
Following \cite{song2023consistency}, with $\mathbf{f}(\bx,t)=\bf{0}$ and $g(t)=\sqrt{2t}$, we have $p(\bx_t|\bx)=\mN({\bx}, t^2\mathbf{I})$, which can be derived using Ito calculus \cite{ikeda2012ito}. Note that we do not perform this equation to directly obtain the diffused latent vector, \ie, $\bx_T=\bx + T \bz$, where $\bz \sim \mN(\bf{0}, \bf{I})$. This single-step diffusion approach does not adequately capture the distribution discrepancy between the latent vectors of adversarial and clean samples. Instead, we aim to explore this distribution discrepancy through a more detailed process using the score function.
To this end, we next calculate the score function $\nabla_{\mathbf{x}}\log p_{t}(\mathbf{x})$.

For the clean sample $\bx$, we have $\bx_t=\bx + t \bz=\mN({\boldsymbol{\mu}}_{\bx}, (\sigma_{\bx}^2+t^2)\mathbf{I})$, yielding
\begin{equation}
    \nabla_{\mathbf{x}}\log p_{t}(\mathbf{x})=-\frac{\bx_t-{\boldsymbol{\mu}}_{\bx}}{\sigma_{\bx}^2+t^2}=-\frac{1}{\sqrt{\sigma_{\bx}^2+t^2}}  \mN(\bf{0}, \bf{I}).
\end{equation}

For the adversarial sample $\hat{\bx}$, we have $\hat{\bx}_t=\bx +\boldsymbol{\varepsilon}  + t \bz=\mN({\boldsymbol{\mu}}_{\bx}, (\sigma_{\bx}^2+t^2)\mathbf{I})+\boldsymbol{\epsilon}_a$, leading to
\begin{equation}
    \nabla_{\hat{\mathbf{x}}}\log p_{t}(\mathbf{\hat{x}})=-\frac{\hat{\bx}_t-{\boldsymbol{\mu}}_{\bx}}{\sigma_{\bx}^2+t^2}=-\frac{1}{\sqrt{\sigma_{\bx}^2+t^2}}  \mN({\bf{0}}, {\bf{I}})-\frac{\boldsymbol{\epsilon}_a}{\sigma_{\bx}^2+t^2}.
\end{equation}

Summing Eqn. (\ref{eqn: ode forward discrete}) from $t=0$ to $t=T$ and setting $\Delta t=1/T$, $t \in \{0, \frac{1}{T}, \ldots, \frac{T-1}{T}\}$, we get
\begin{align}
    \bx_T &= \bx_0 + \frac{1}{T} \sum_{t=0}^1 t \nabla_{\mathbf{x}}\log p_{t}(\mathbf{x}) \nonumber\\
    &= \bx_0 + \frac{1}{T} \sum_{t=0}^1 \mN({\bf{0}}, \frac{ t^2}{\sigma_{\bx}^2+ t^2} \mathbf{I}) \nonumber \\
    &= \bx_0 + \mN({\bf{0}}, \frac{1}{T} \sum_{t=0}^1 \frac{ t^2}{\sigma_{\bx}^2+ t^2} \mathbf{I}) \label{eqn: sum_N}
\end{align}
As $\Delta t \to 0$ (i.e., $T \to \infty$), Eqn. (\ref{eqn: sum_N}) converges to:
\begin{equation}
\label{eqn: x_T_x_0}
    \bx_T = \bx_0 + \mN(\mathbf{0}, \sigma^2_{cl} \mathbf{I}),
\end{equation}
where $\sigma_{cl}^2=\mathbb{E}_t  \frac{ t^2}{\sigma_{\bx}^2+ t^2}$.

Similarly, for adversarial sample $\hat{\bx}$, we have
\begin{equation}
\label{eqn: x_T_x_0 adv}
    \hat{\bx}_T = \hat{\bx}_0 + \mN(\mathbf{0}, \sigma^2_{cl} \mathbf{I})-\mathbb{E}_t  \frac{ t\boldsymbol{\epsilon}_a}{\sigma_{\bx}^2+ t^2},
\end{equation}
Combining Eqn. (\ref{eqn: x_T_x_0}) and Eqn. (\ref{eqn: x_T_x_0 adv}), we conclude:
    \begin{equation}
         \bx_T -\hat{\bx}_T \sim \mN( \mathbf{0}, 2\sigma_{cl}^2\mathbf{I})+\boldsymbol{\mu}_{\epsilon},
    \end{equation}
    where $\boldsymbol{\mu}_\epsilon=\left(\mathbb{E}_t  \frac{ t}{\sigma_{\bx}^2+ t^2}-1\right)\boldsymbol{\epsilon}_a$. 
\end{proof}
\begin{remark}
    Note that the coefficient $\mathbb{E}_t  \frac{ t}{\sigma_{\bx}^2+ t^2}-1$ in $\boldsymbol{\mu}_\epsilon$ can be further simplified as $\frac{1}{2} \mathrm{ln} (1+\frac{1}{\sigma_{\bx}^2})-1$, which means the coefficient $\mathbb{E}_t  \frac{ t}{\sigma_{\bx}^2+ t^2}-1=0$ if and only if  $\sigma_{\bx}^2=1/99$. However, this is unlikely in practice, as the natural data distributions typically exhibit not only larger but also diverse $\sigma_{\bx}$ values, making such a narrow variance quite rare.
\end{remark}

\subsection{Proof of Proposition \ref{prop: upper_bound}}
\label{app: prop}
\noindent\textbf{Proposition \ref{prop: upper_bound}}
\emph{
Given a test sample $\hat{\bx} $, it holds that optimizing a set of latent vectors $\{\tilde{\bz}_i\}_{i=1}^K$ by Eqn. (\ref{eq: overview}) gives an upper bound on the reconstruction for the clean sample $\bx$:
\begin{equation*}
% \label{eq: upper_bound}
    \frac{1}{K} \sum_{i=1}^K\| f_{\theta}(\tilde{\mathbf{z}}_i) - \mathbf{x} \|_1 +\beta \mathcal{L}_{d}(\tilde{\bz}) \leq C + \mathcal{L}_{a}(\tilde{\bz}, \hat{\mathbf{x}})+\beta \mathcal{L}_{d}(\tilde{\bz}), 
\end{equation*}
where $C$ is a constant related to the adversarial perturbation.
}
\begin{proof}
Based on the adversarial sample $\hat{\bx} = \bx +\boldsymbol{\epsilon}_a$, we have
\begin{align}
    &\frac{1}{K} \sum_{i=1}^K\| f_{\theta}(\tilde{\mathbf{z}}_i) - \mathbf{x} \|_1 +\beta \mathcal{L}_{d}(\tilde{\bz}) \nonumber\\
    = & \frac{1}{K} \sum_{i=1}^K\| f_{\theta}(\tilde{\mathbf{z}}_i) - \hat{\mathbf{x}} - \boldsymbol{\epsilon}_a \|_1 +\beta \mathcal{L}_{d}(\tilde{\bz}) \nonumber\\
    \le & \| \boldsymbol{\epsilon}_a \|_1 + \frac{1}{K} \sum_{i=1}^K\| f_{\theta}(\tilde{\mathbf{z}}_i) - \hat{\mathbf{x}} \|_1 +\beta \mathcal{L}_{d}(\tilde{\bz}) \nonumber\\
    \le & \| \boldsymbol{\epsilon}_a \|_1 + \frac{1}{K}\sum_{i=1}^K \left\| f_{\theta}(\tilde{\mathbf{z}}_i){-} \hat{\mathbf{x}})\right\|_1 \nonumber\\
    &+ \alpha \left(1-  \mathrm{SSIM}(f_{\theta}(\tilde{\mathbf{z}}_i)), \hat{\mathbf{x}}\right) +\beta \mathcal{L}_{d}(\tilde{\bz})  \nonumber \\
    =& \| \boldsymbol{\epsilon}_a \|_1 + \alpha +\mathcal{L}_{a}(\tilde{\bz}, \hat{\mathbf{x}})+\beta \mathcal{L}_{d}(\tilde{\bz})  \nonumber \\
    =& C + \mathcal{L}_{a}(\tilde{\bz}, \hat{\mathbf{x}})+\beta \mathcal{L}_{d}(\tilde{\bz}),
\end{align}
where $C=\| \boldsymbol{\epsilon}_a \|_1 + \alpha$.
\end{proof}

% \vspace{1pt}
\section{More Visualization Results}
\begin{figure}[h]
    \begin{center}
        \subfloat[CIFAR-10, Improved Diffusion \cite{nichol2021improved}]
        {\includegraphics[width=0.46\linewidth]{ 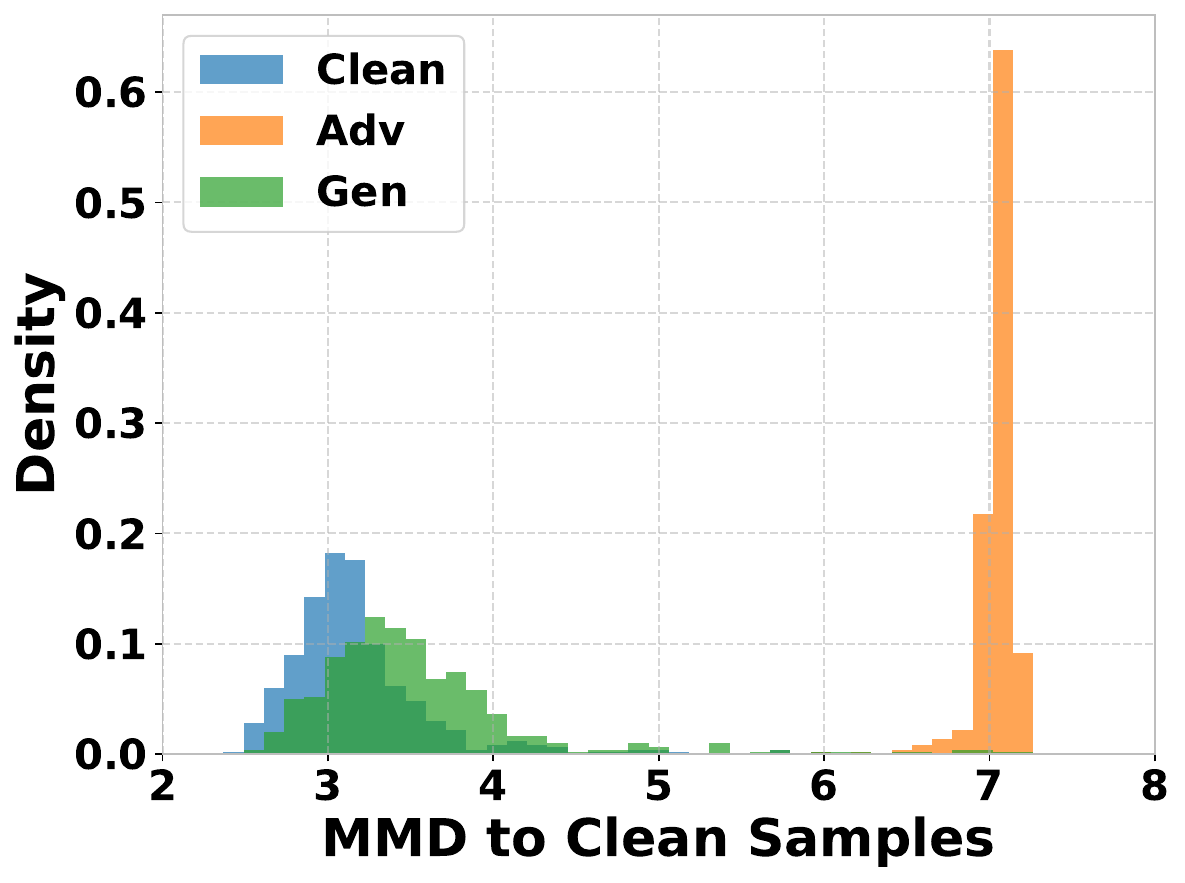}}
        \subfloat[ImageNet, ProGAN \cite{karras2017progressive}]
        {\includegraphics[width=0.49\linewidth]{ 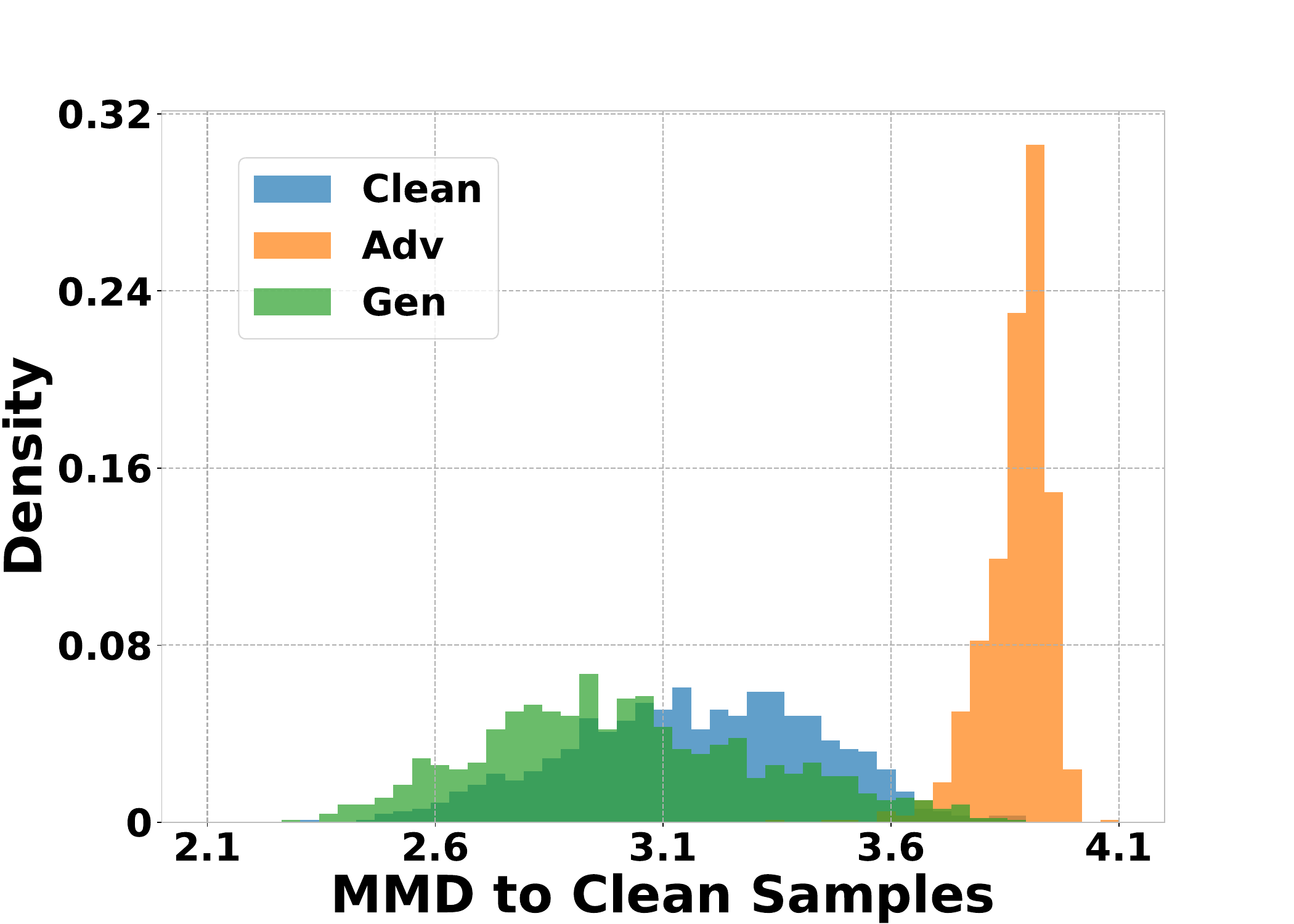}}
        \caption{More histograms cases of MMD distances \cite{gretton2012kernel} between the features of clean (Cln) and clean samples \textit{v.s.} generated (Gen) and clean samples \textit{v.s.} adversarial (Adv) samples and clean samples on CIFAR-10 and ImageNet using different generative models, \eg, Improved Diffusion \cite{nichol2021improved} and ProGAN \cite{karras2017progressive}, showing the common characteristics of different generative domains.}
        \label{fig: motivition_appendix}
    \end{center}
\end{figure}

%-------------------------------------------------------------------------

\begin{figure*}[h]
    \begin{center}
    \subfloat[Consistency Model \cite{song2023consistency}]
    {\includegraphics[width=0.8\linewidth]{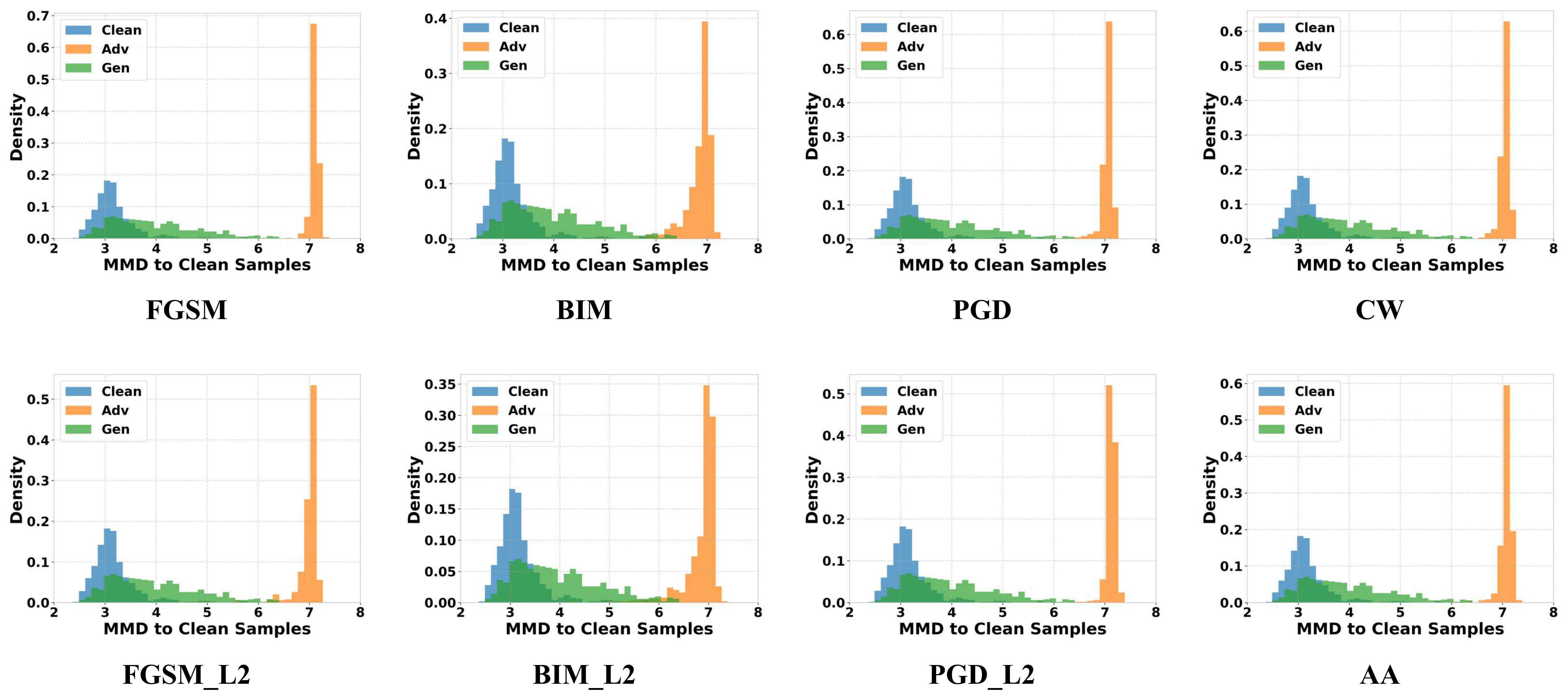}}
    
    \subfloat[Improved Diffusion \cite{nichol2021improved}]
    {\includegraphics[width=0.8\linewidth]{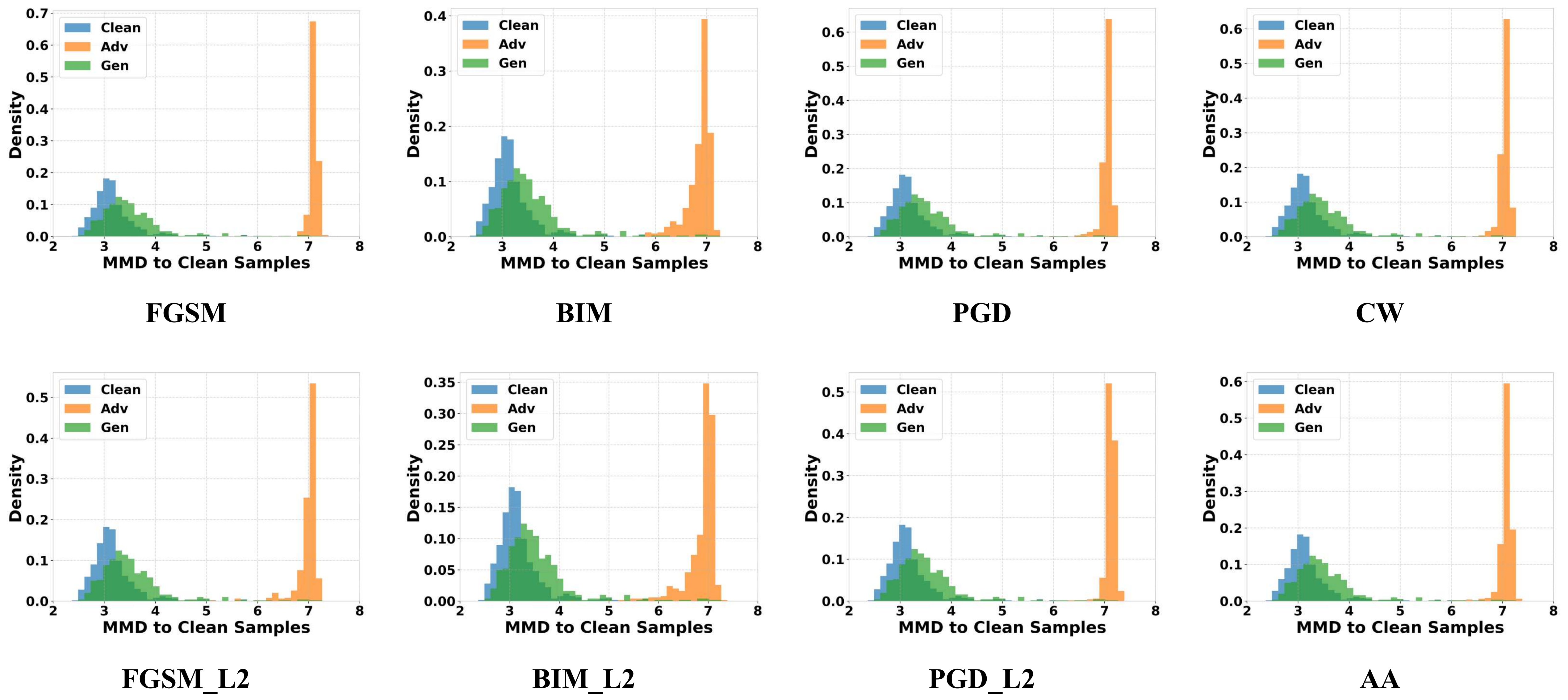}}
    % \vspace{-15pt}
    \caption{ {More histograms cases of MMD distances \cite{gretton2012kernel} between the features of clean (Cln) and clean samples \textit{v.s.} generated (Gen) and clean samples \textit{v.s.} adversarial (Adv) samples and clean samples on CIFAR-10. Here, we consider more types of attacks with $\epsilon=4/255$, including FGSM \cite{goodfellow2014explaining}, BIM \cite{kurakin2018adversarial}, CW \cite{carlini2017towards}, and AutoAttack \cite{croce2020reliable}. The results consistently show the same trend: the features of generated samples remain significantly closer to those of clean samples than to adversarial ones, robustly confirming our original observation across diverse attack families and threat models. }}
    \label{fig: motivation}
    \end{center}
    % \vspace{5cm}
\end{figure*}

\begin{figure*}[h]
    \begin{center}
    \subfloat[CIFAR-10]
    {\includegraphics[width=\linewidth]{ 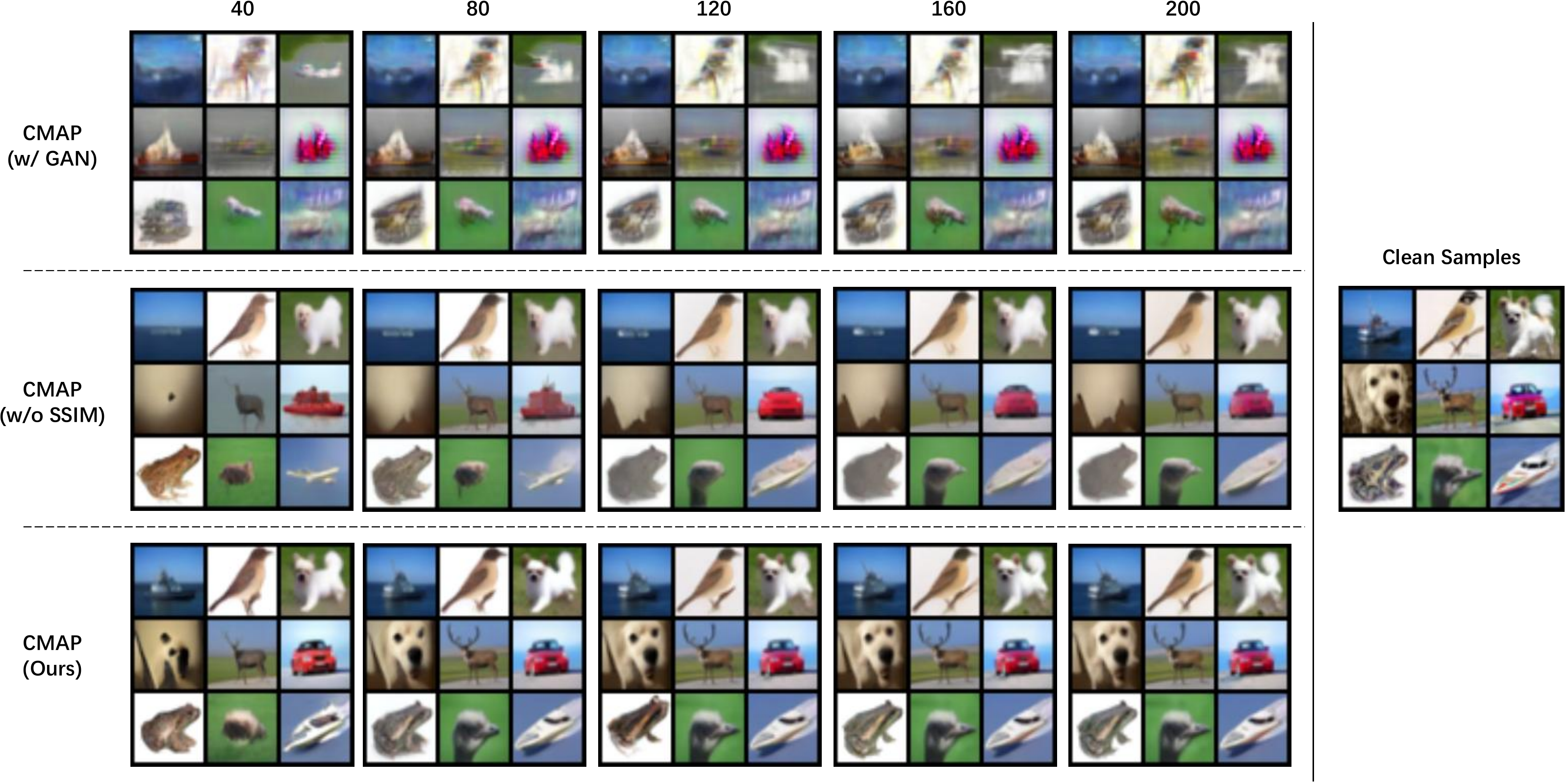}}
    
    \subfloat[ImageNet-100]
    {\includegraphics[width=\linewidth]{ 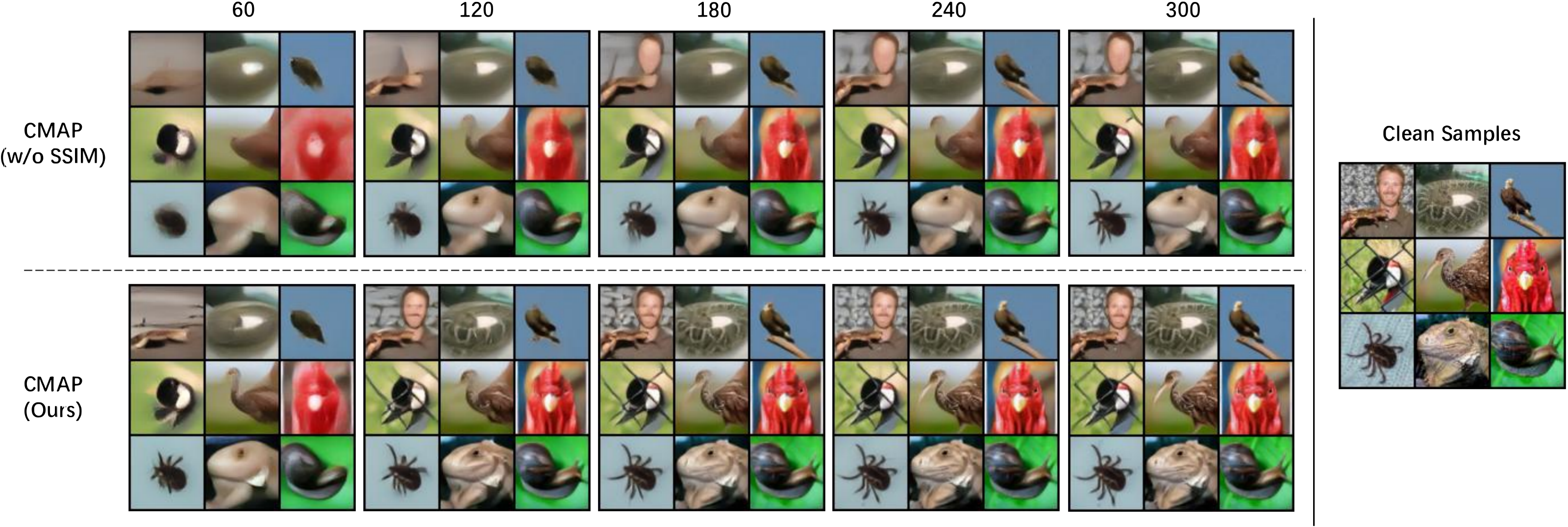}}
    % \vspace{-15pt}
    \caption{Visualizations of the perceptual consistency restoration process using different generative models and optimization objectives. The optimization of latent vectors with a generative adversarial network (BigGAN \cite{brock2018large}) \textbf{(first row in (a))} typically results in images that exhibit structural blurriness and artifacts. Furthermore, when employing only L1 loss as the alignment loss \textbf{(second row in (a), first row in (b))}, the generated samples predominantly preserve overall color and brightness, lacking structural details. In contrast, our \mymethod~successfully generates refined and fidelity images \textbf{(third row in (a), second row in (b))}.}
    \label{fig: visualization}
    \end{center}
    % \vspace{5cm}
\end{figure*}

\begin{figure*}[t]
    \centering

    \includegraphics[width=0.6\linewidth]{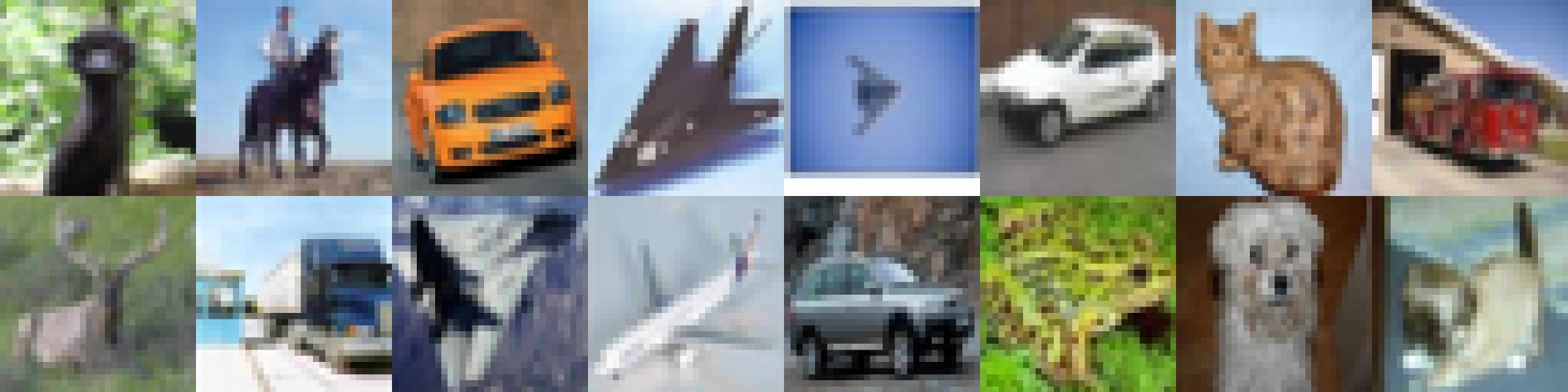}\\
    \small (a) Original clean images randomly selected from the CIFAR-10 dataset.

    \vspace{0.15em}

    \includegraphics[width=0.6\linewidth]{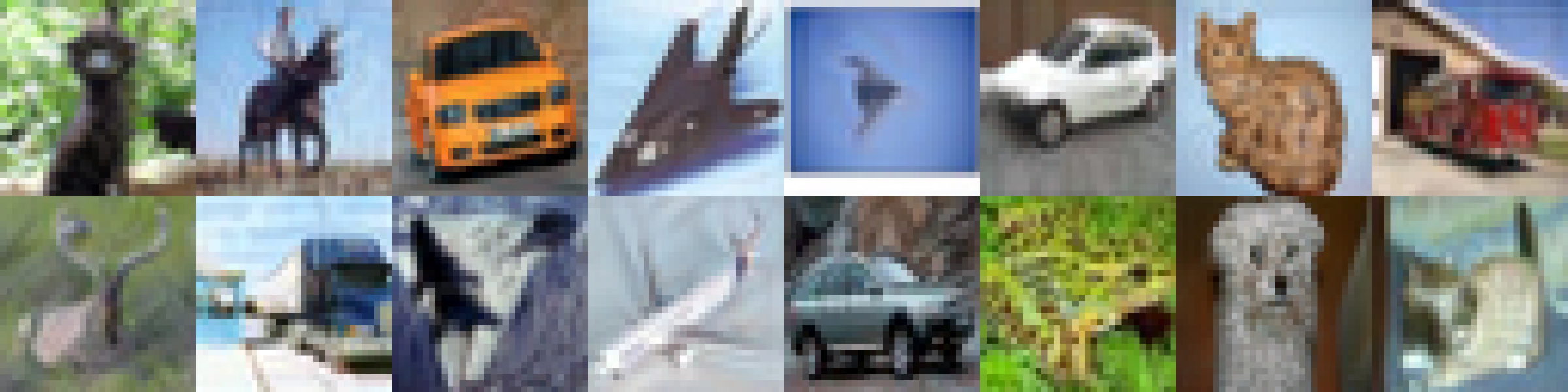}\\
    \small (b) Adversarial examples generated from the original images.

    \vspace{0.15em}

    \includegraphics[width=0.6\linewidth]{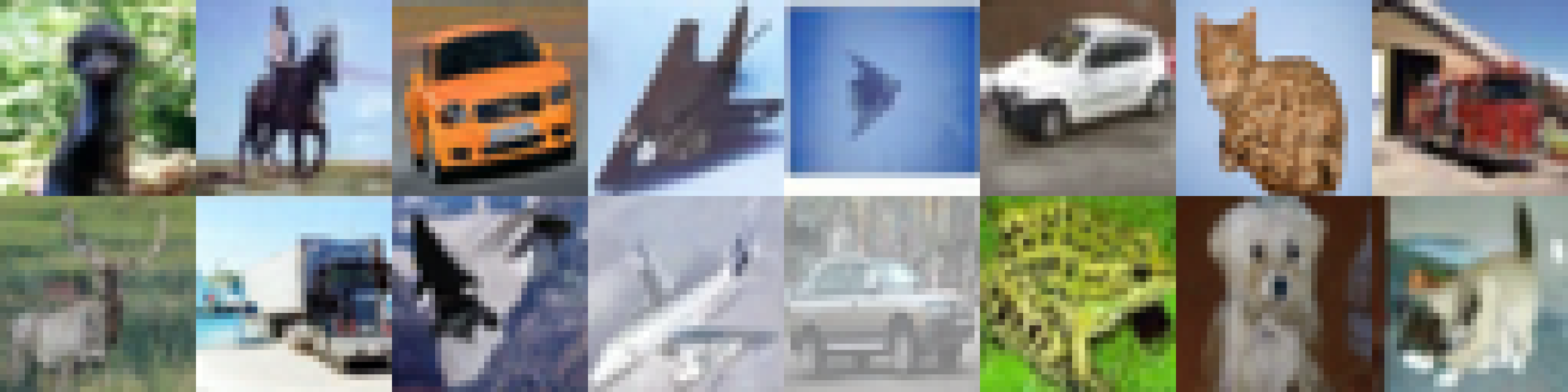}\\
    \small (d) Purified images obtained by \textbf{DiffPure}.

    \vspace{0.15em}

    \includegraphics[width=0.6\linewidth]{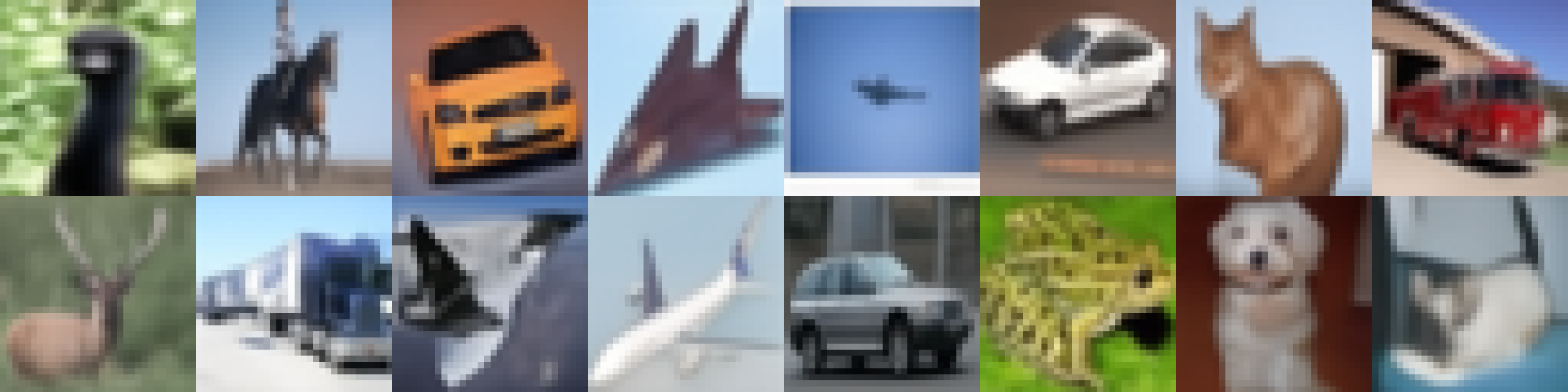}\\
    \small (c) Purified images obtained by \textbf{GNSP}.

    \vspace{0.15em}

    \includegraphics[width=0.6\linewidth]{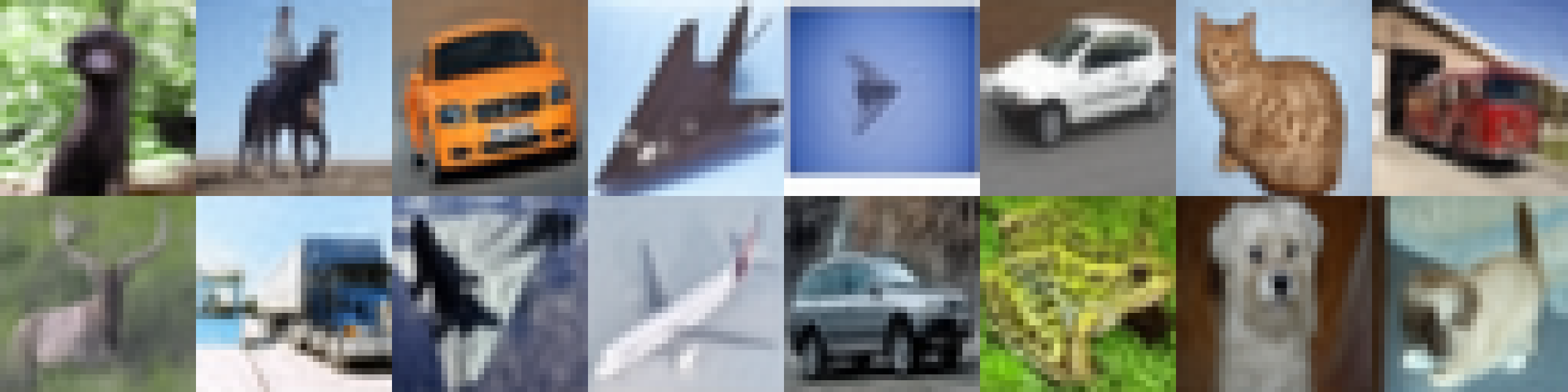}\\
    \small (e) Purified images obtained by \textbf{FreqPure}.

    \vspace{0.15em}

    \includegraphics[width=0.6\linewidth]{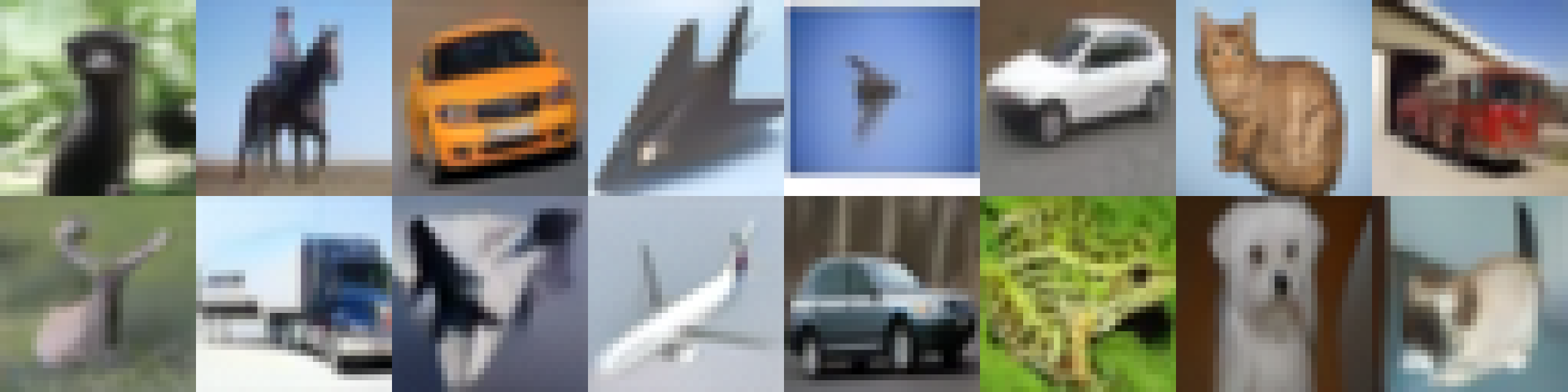}\\
    \small (f) Purified images obtained by \textbf{CMAP (Ours)}.

    \caption{ {Qualitative evaluation and visual comparison of different purification methods on CIFAR-10 against PGD+EOT attacks with $\epsilon=8/255$ (PART I), including DiffPure \cite{nie2022diffusion}, GNSP \cite{lee2023robust}, FreqPure \cite{pei2025diffusion}. Unless otherwise specified, all adversarial examples shown are crafted against our CMAP. Specifically, the adversarial samples in (b) are generated against CMAP, while those corresponding to other purification methods are provided in Fig. \ref{fig:adversarial}.}}
    \label{fig: cifar_part1}
\end{figure*}

\begin{figure*}[t]
    \centering

    \includegraphics[width=0.6\linewidth]{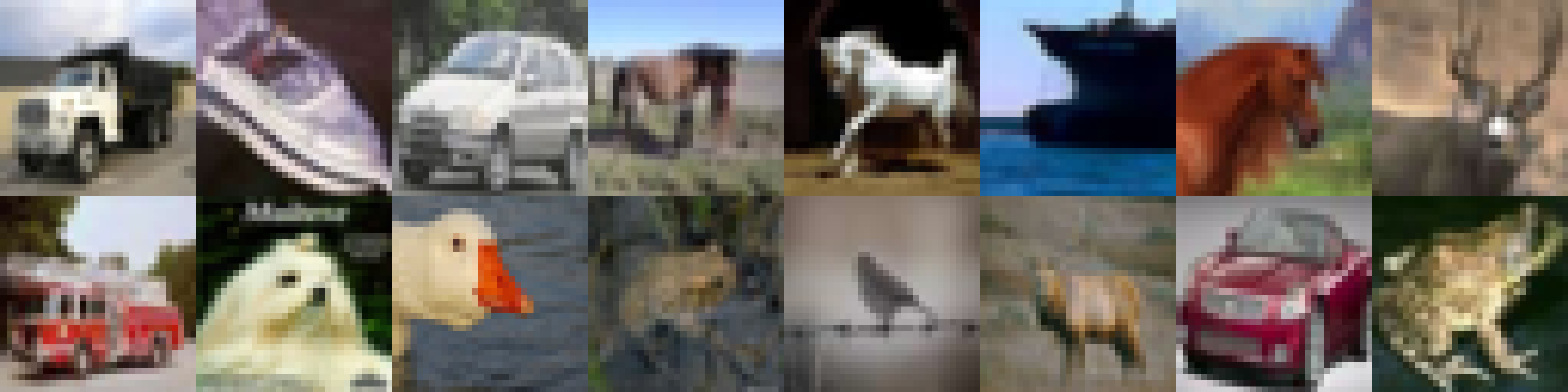}\\
    \small (a) Original examples randomly selected from the CIFAR-10 dataset.

    \vspace{0.15em}

    \includegraphics[width=0.6\linewidth]{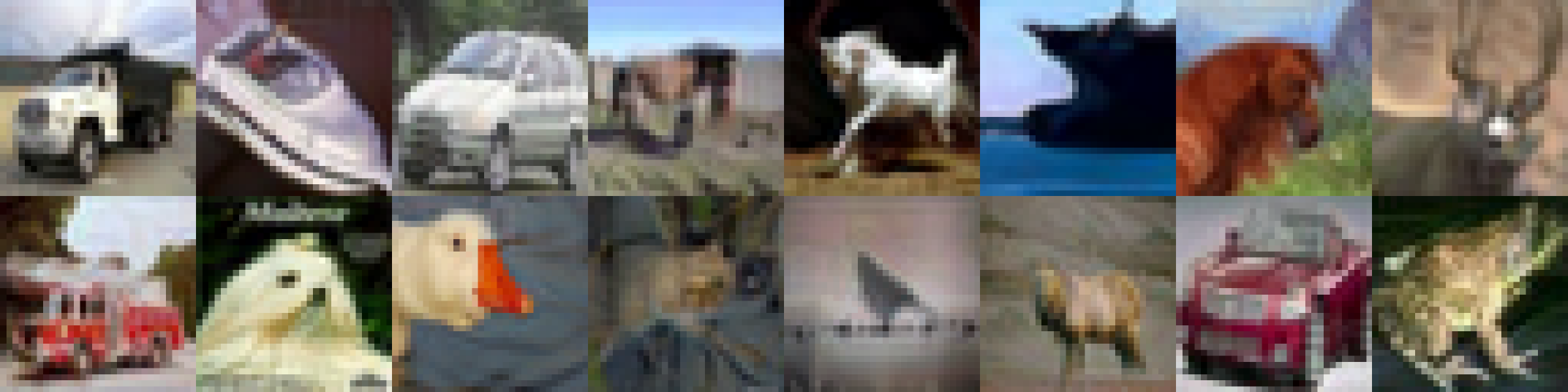}\\
    \small (b) Adversarial examples generated from the original images.

    \vspace{0.15em}

    \includegraphics[width=0.6\linewidth]{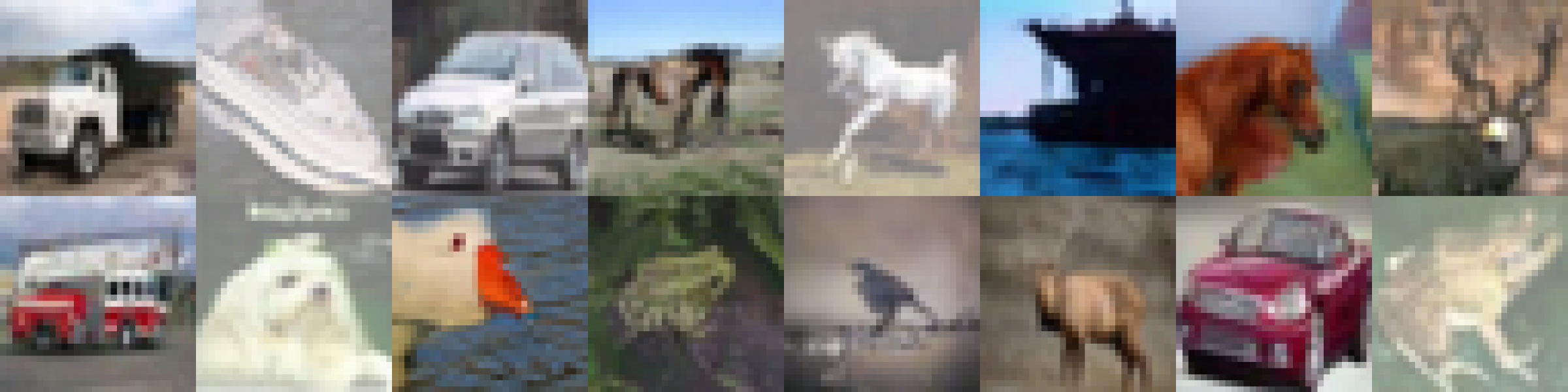}\\
    \small (d) Purified examples obtained by \textbf{DiffPure}.

    \vspace{0.15em}

    \includegraphics[width=0.6\linewidth]{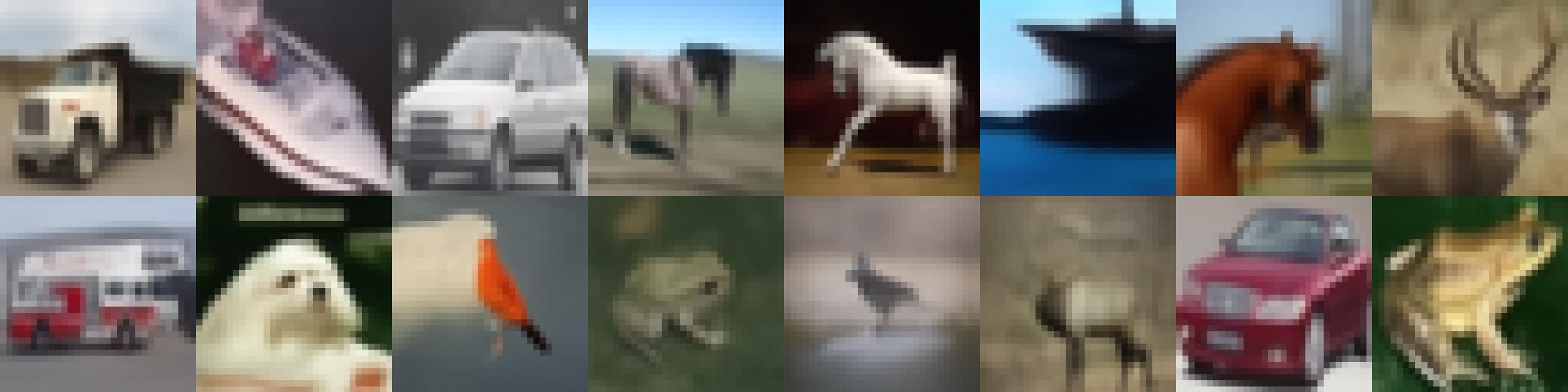}\\
    \small (c) Purified examples obtained by \textbf{GNSP}.

    \vspace{0.15em}

    \includegraphics[width=0.6\linewidth]{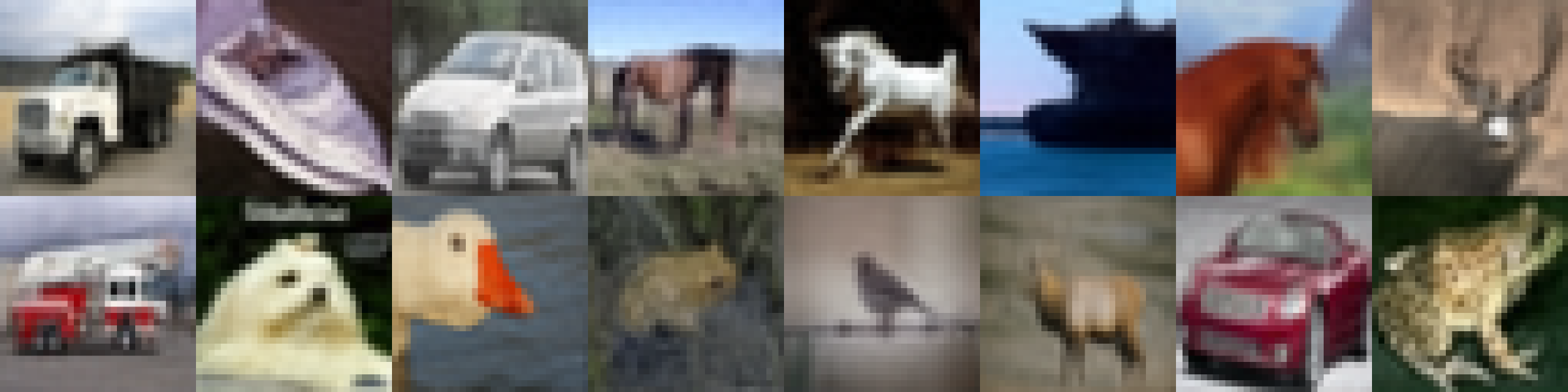}\\
    \small (e) Purified examples obtained by \textbf{FreqPure}.

    \vspace{0.15em}

    \includegraphics[width=0.6\linewidth]{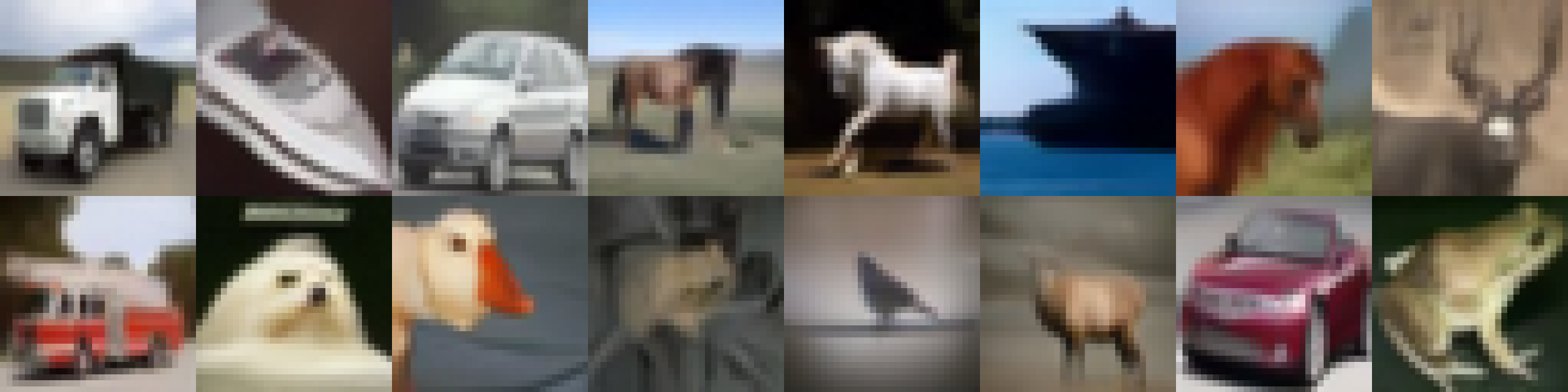}\\
    \small (f) Purified examples obtained by \textbf{CMAP (Ours)}.

    \caption{ {Qualitative evaluation and visual comparison of different purification methods on CIFAR-10 against PGD+EOT attacks with $\epsilon=8/255$ (PART II), including DiffPure \cite{nie2022diffusion}, GNSP \cite{lee2023robust}, FreqPure \cite{pei2025diffusion}. 
    % Here, the adversarial examples in (b) are crafted specifically against our CMAP, while those corresponding to other purification methods are provided in Fig. \ref{fig:adversarial}.
    }}
    \label{fig: cifar_part2}
\end{figure*}

\begin{figure*}[t]
    \centering

    \includegraphics[width=0.6\linewidth]{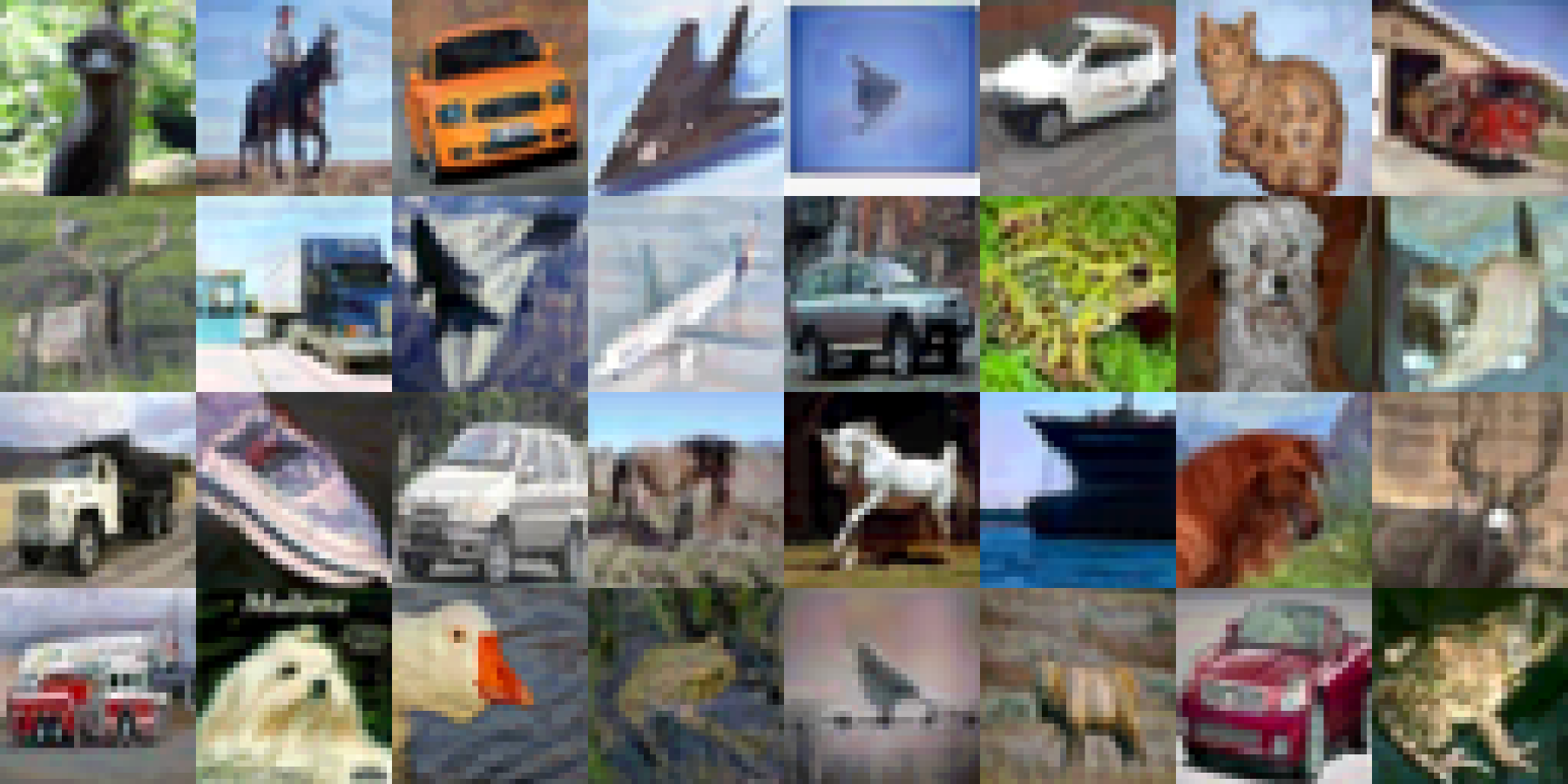}\\
    \small (c) Adversarial examples against \textbf{FreqPure} generated from the original images.

    \vspace{0.15em}

    \includegraphics[width=0.6\linewidth]{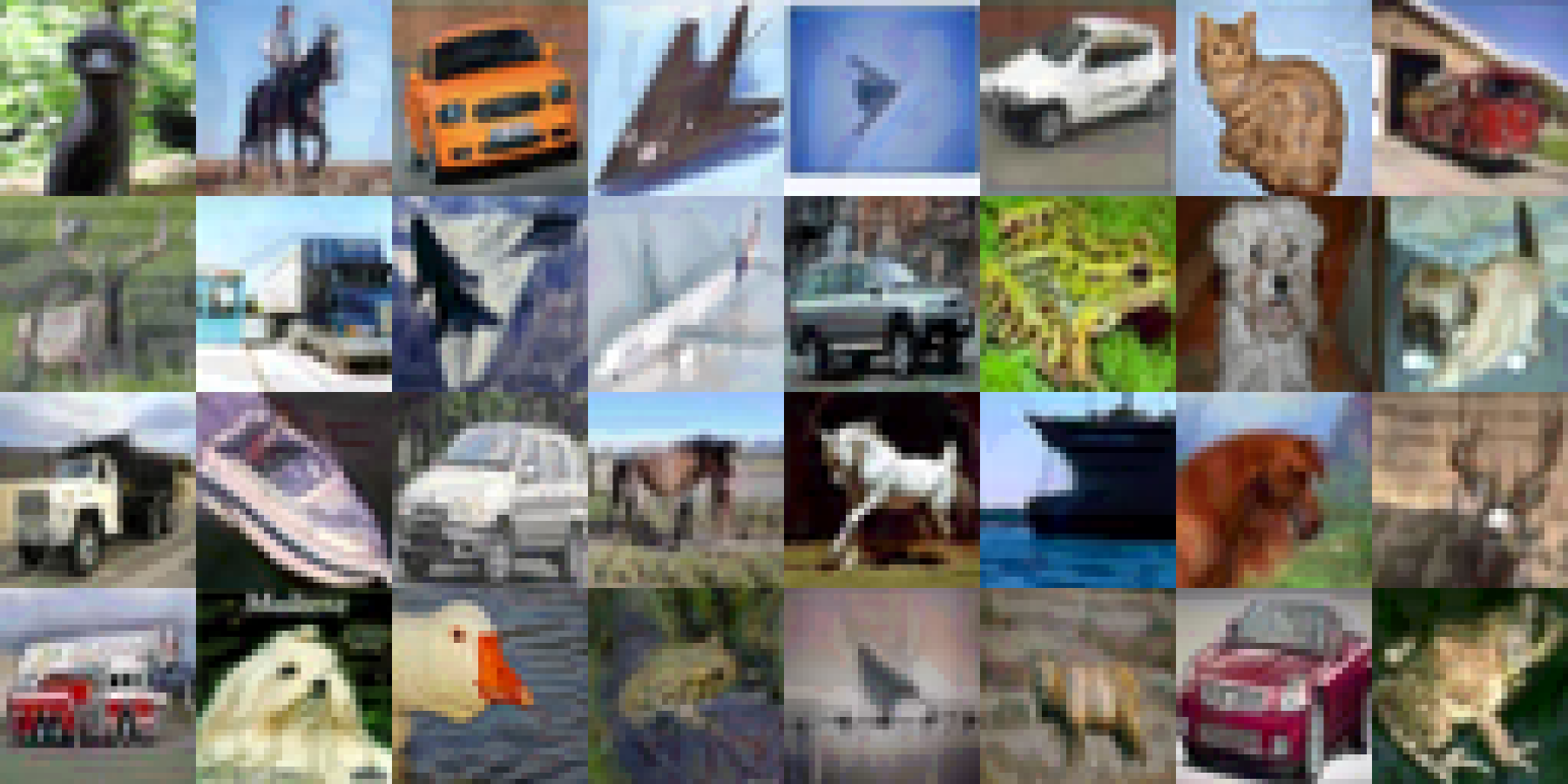}\\
    \small (a) Adversarial examples against \textbf{DiffPure} generated from the original images.

    \vspace{0.15em}

    \includegraphics[width=0.6\linewidth]{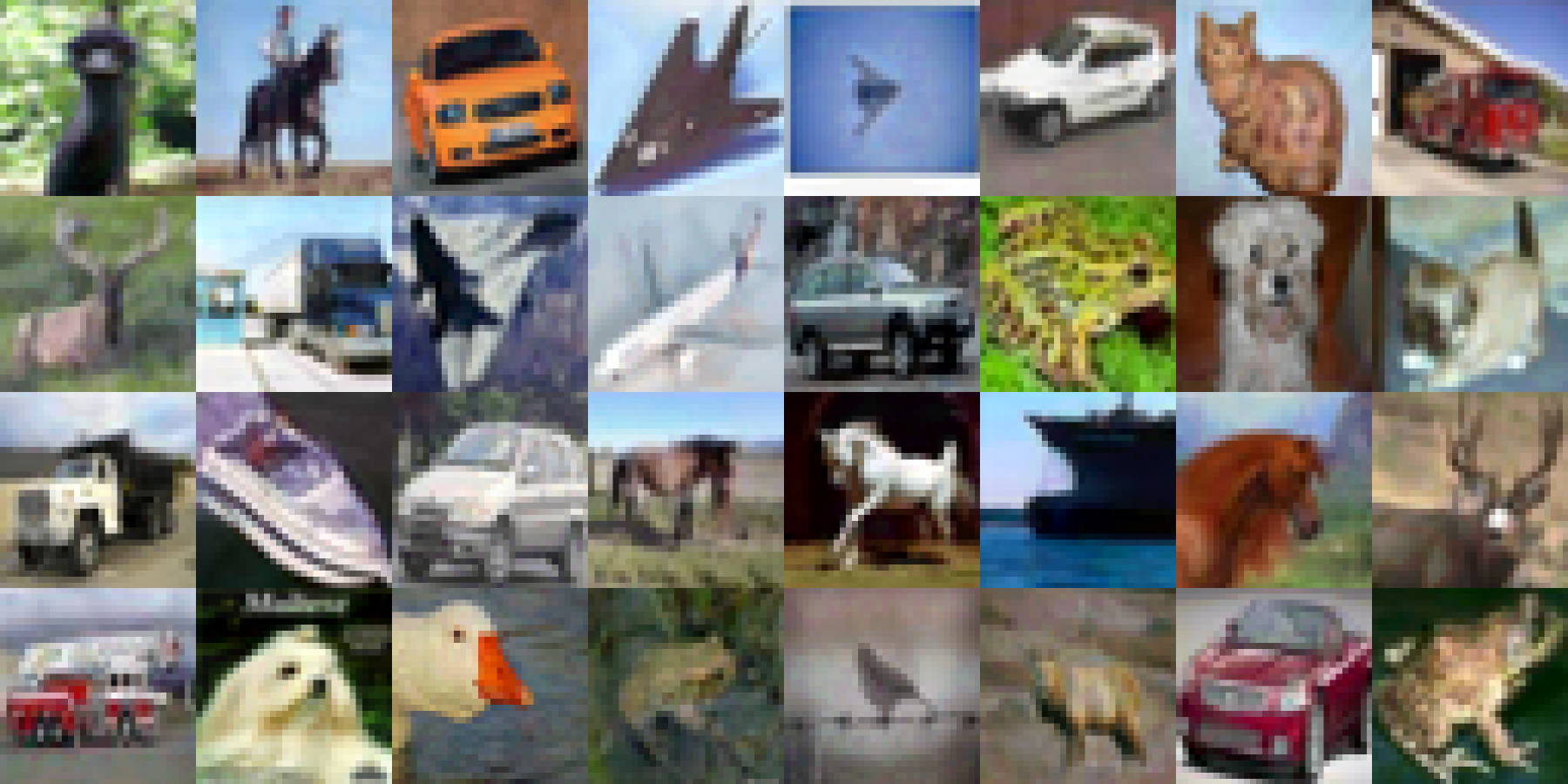}\\
    \small (b) Adversarial examples against \textbf{GNSP} generated from the original images.

    \vspace{0.15em}

    \includegraphics[width=0.6\linewidth]{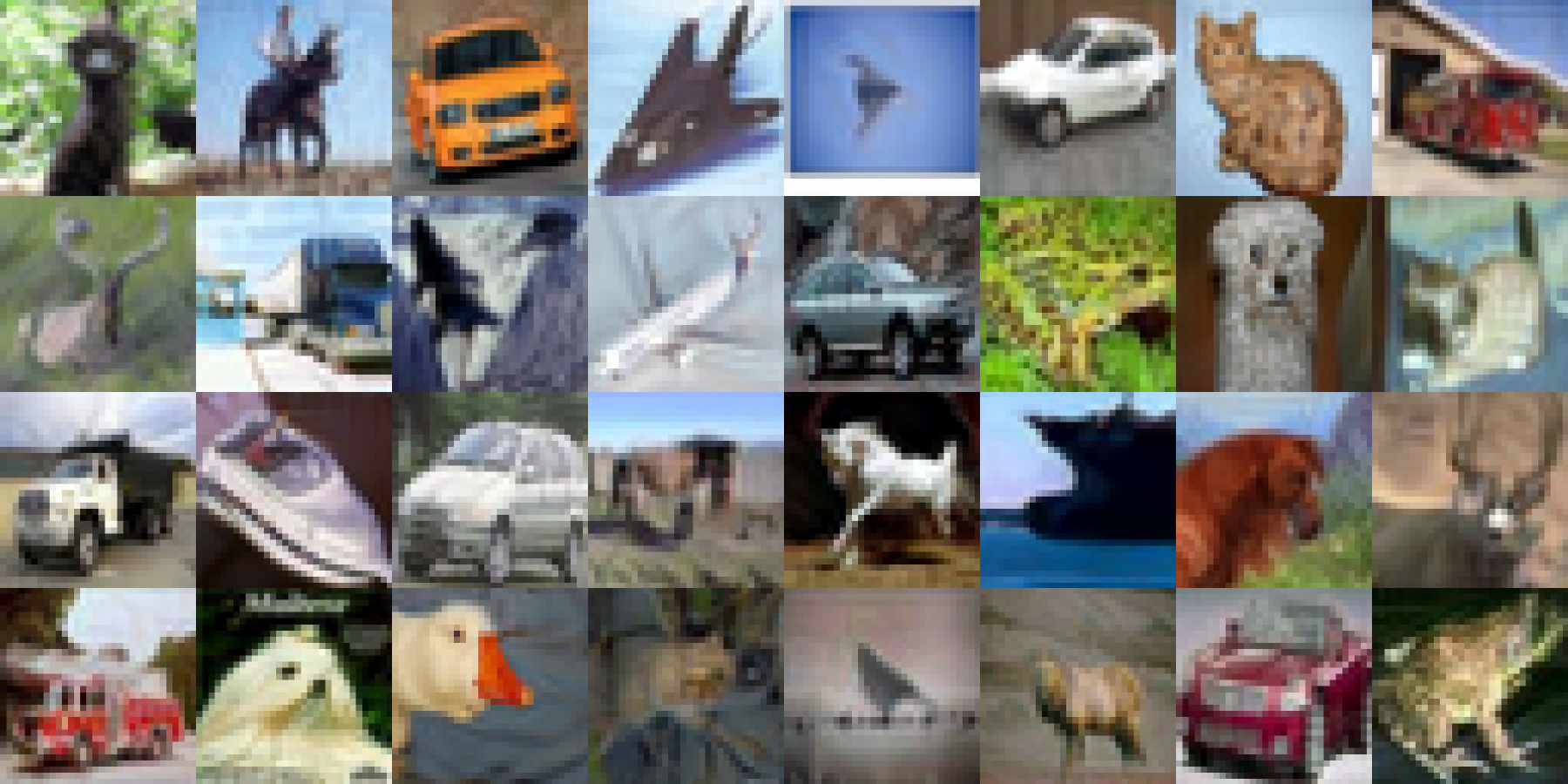}\\
    \small (d) Adversarial examples against \textbf{CMAP} generated from the original images.

    \caption{ {Visualizations of different adversarial samples against different purification baselines on CIFAR-10 with $\epsilon=8/255$.}}
    \label{fig:adversarial}
\end{figure*}

\begin{figure*}[th]
    \begin{center}
    \includegraphics[width=0.7\textwidth]{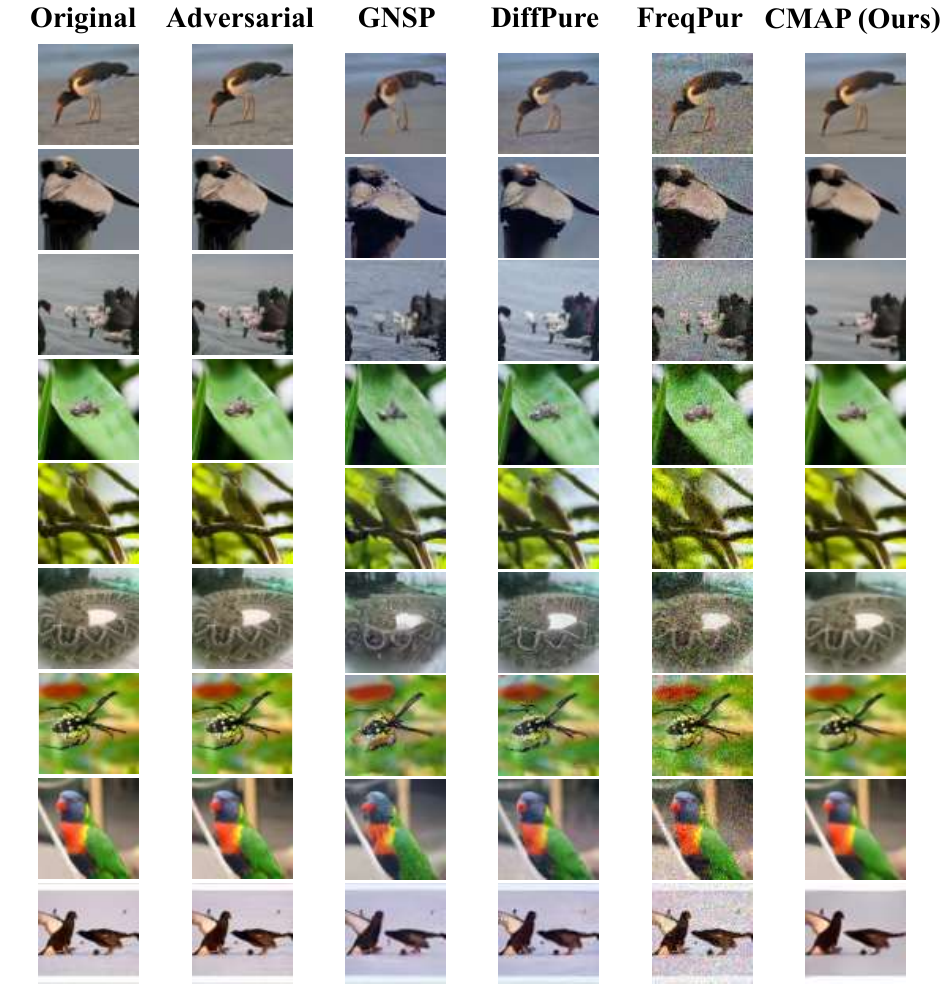}
    \vspace{-7pt}
    \caption{ {Qualitative evaluation and visual comparison of different purification methods on ImageNet-100 against PGD+EOT attacks with $\epsilon=4/255$ (PART I), including DiffPure \cite{nie2022diffusion}, GNSP \cite{lee2023robust}, FreqPure \cite{pei2025diffusion}.}}
    \label{fig: imagenet_part1}
    \vspace{-12pt}
    \end{center}
\end{figure*}

\begin{figure*}[th]
    \begin{center}
    \includegraphics[width=0.7\textwidth]{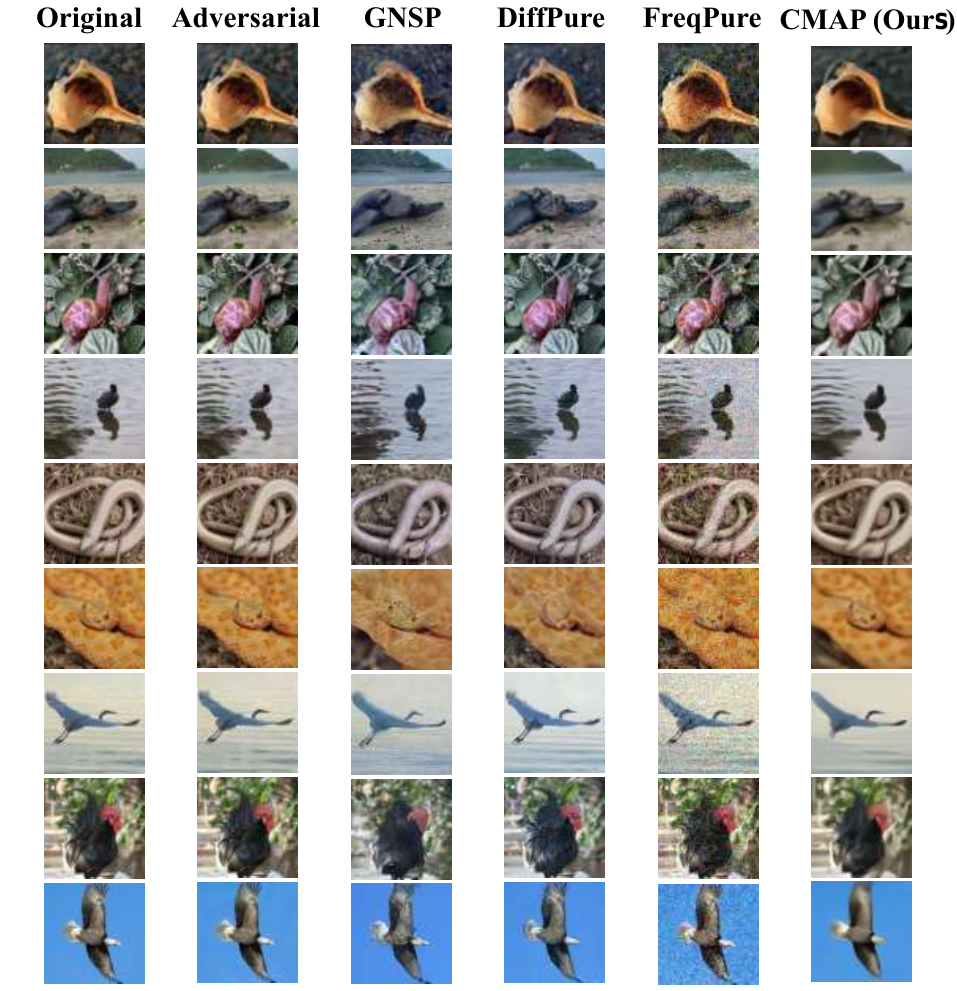}
    \vspace{-7pt}
    \caption{ {Qualitative evaluation and visual comparison of different purification methods on ImageNet-100 against PGD+EOT attacks with $\epsilon=4/255$ (PART II), including DiffPure \cite{nie2022diffusion}, GNSP \cite{lee2023robust}, FreqPure \cite{pei2025diffusion}.}}
    \label{fig: imagenet_part2}
    \vspace{-12pt}
    \end{center}
\end{figure*}

\begin{figure*}[t]
    \centering

    \includegraphics[width=0.56\linewidth]{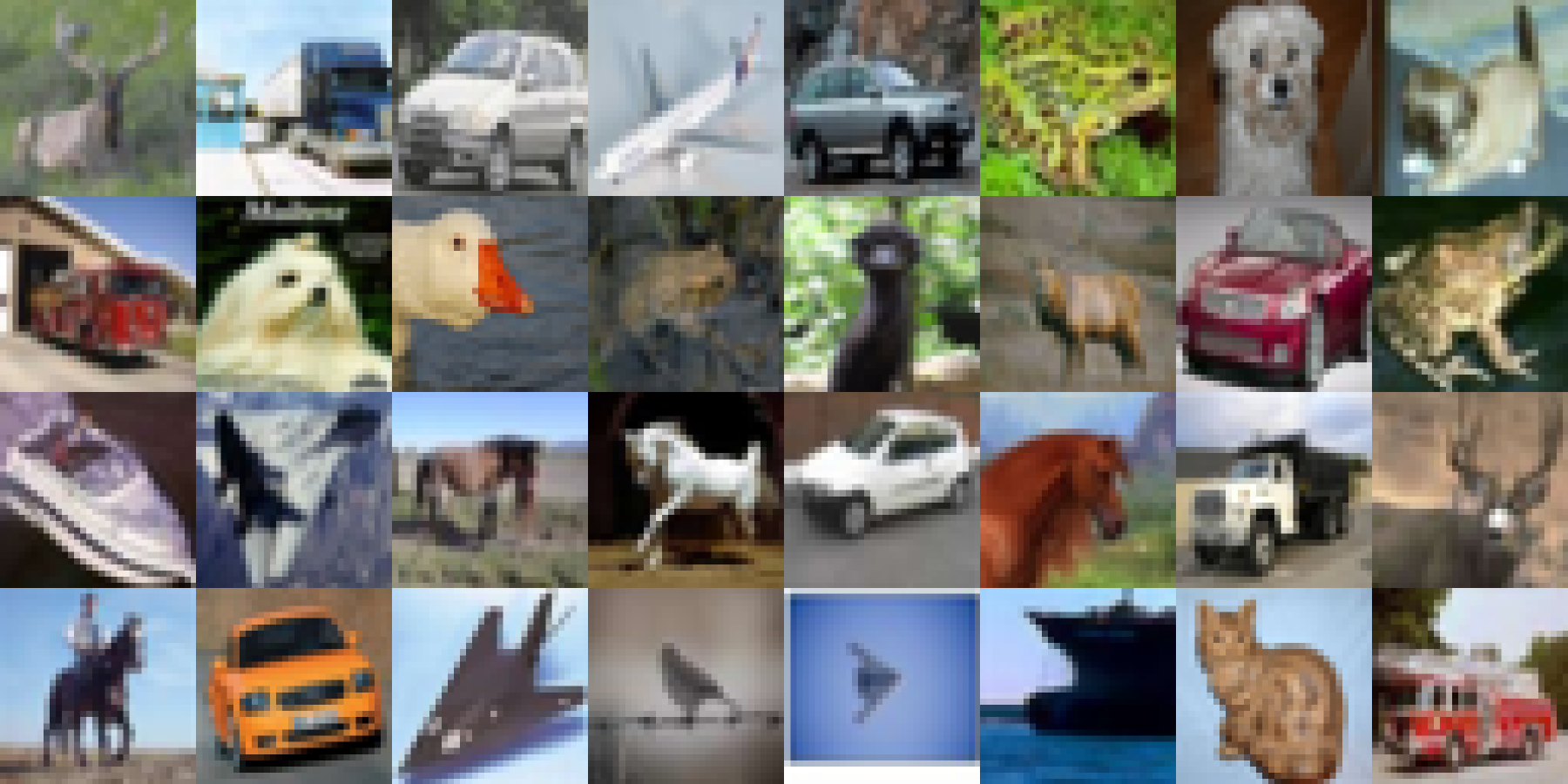}\\
    \small (a) Original examples randomly selected from the CIFAR-10 dataset.

    \vspace{0.15em}

    \includegraphics[width=0.56\linewidth]{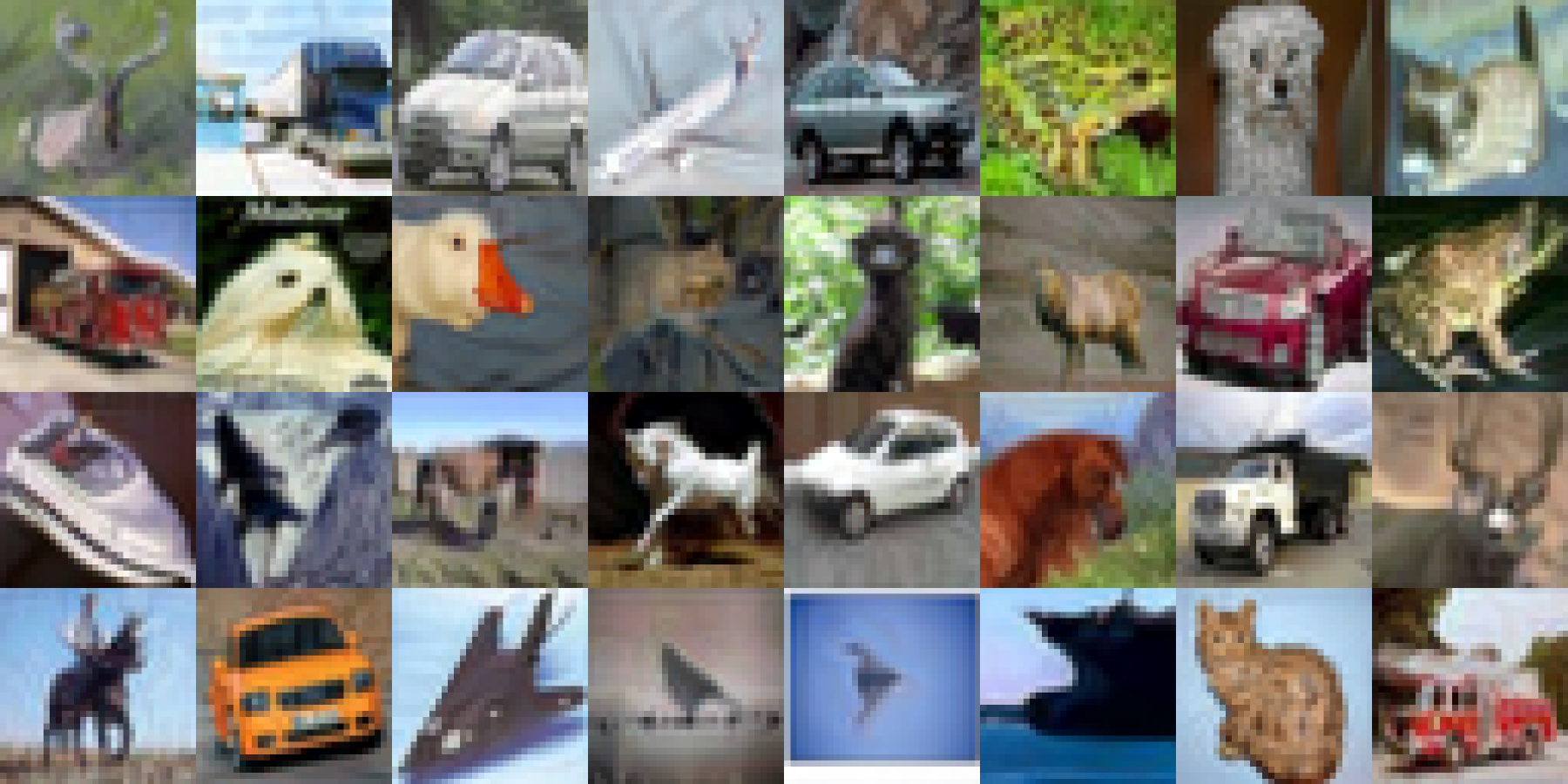}\\
    \small (b) Adversarial examples generated from the original images.

    \vspace{0.15em}

    \includegraphics[width=0.56\linewidth]{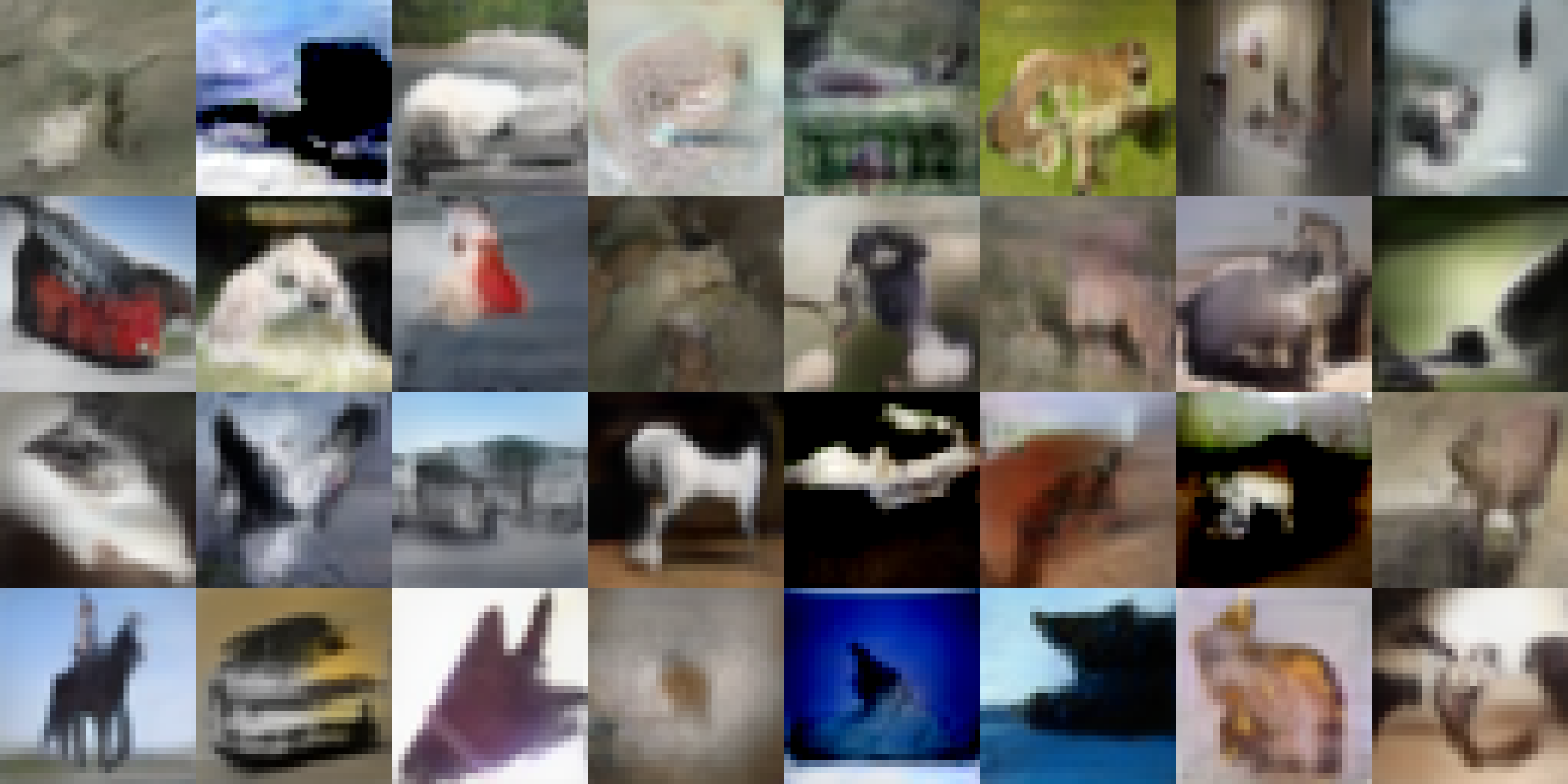}\\
    \small (c) Purified examples obtained by CMAP with GAN as the generator.

    \vspace{0.15em}

    \includegraphics[width=0.56\linewidth]{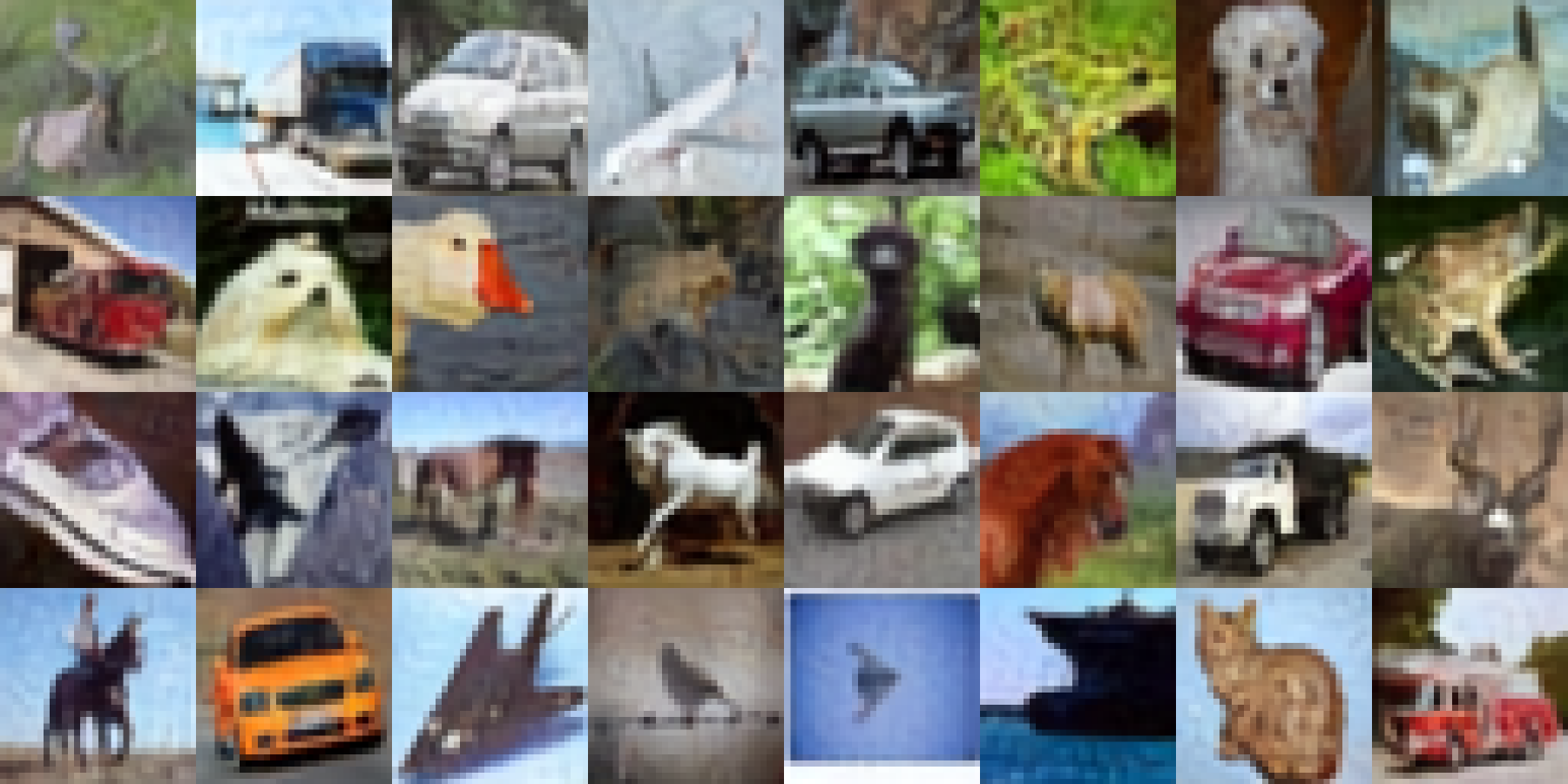}\\
    \small (d) Purified examples obtained by CMAP with VAE as the generator.

    \caption{ {Visualizations of adversarial purification against PGD+EOT with $\epsilon=8/255$ on CIFAR‑10 using our CMAP, with Projected‑GAN \cite{sauer2021projected} and VAE \cite{esser2021taming} as the generators, where the GAN adversarial examples are crafted solely by the classifier. 
    Purified images using Projected-GAN in (c) frequently exhibit semantic blurriness and structural artifacts, illustrating the challenge of direct latent-space optimization in GANs without generator finetuning \cite{salimans2016improved, abdal2019image2stylegan}. By contrast, VAE-based CMAP in (d) yields higher-fidelity reconstructions with better semantic integrity and visual details while removing adversarial perturbations effectively, benefitting from the VAE’s probabilistic latent modeling and smoother, more invertible latent geometry.}}
    %  {TODO} The purified images using GAN exhibit semantic blurriness and structural artifacts, reflecting the inherent difficulty of direct latent‑space optimization in standard GANs—a limitation often alleviated by generator finetuning \cite{salimans2016improved, abdal2019image2stylegan}.}
    \label{fig:comparison_part3}
\end{figure*}

\begin{figure*}[t]
    \centering

    \includegraphics[width=0.6\linewidth]{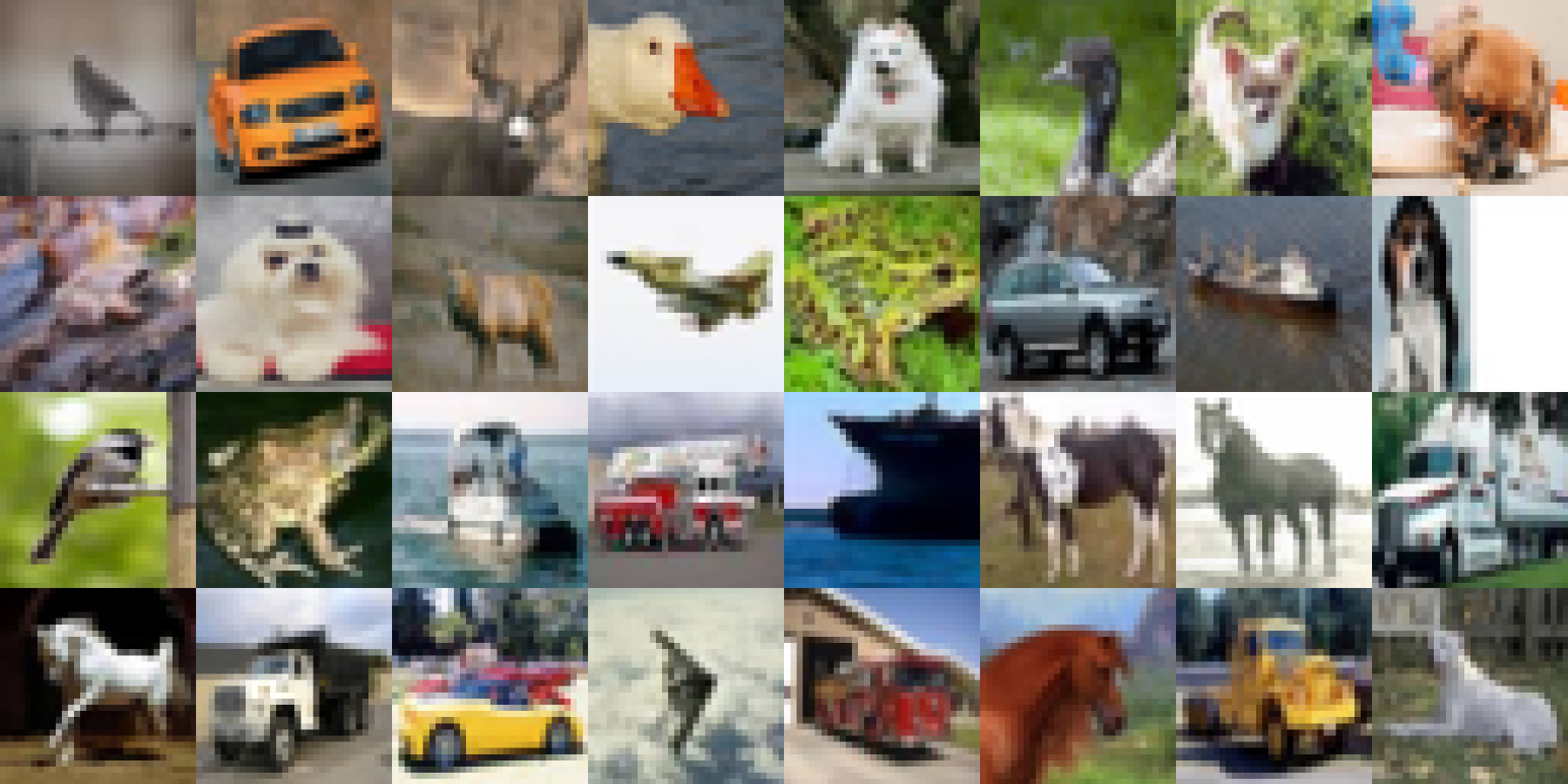}\\
    \small (a) Original examples randomly selected from the CIFAR-10 dataset.

    \vspace{0.15em}

    \includegraphics[width=0.6\linewidth]{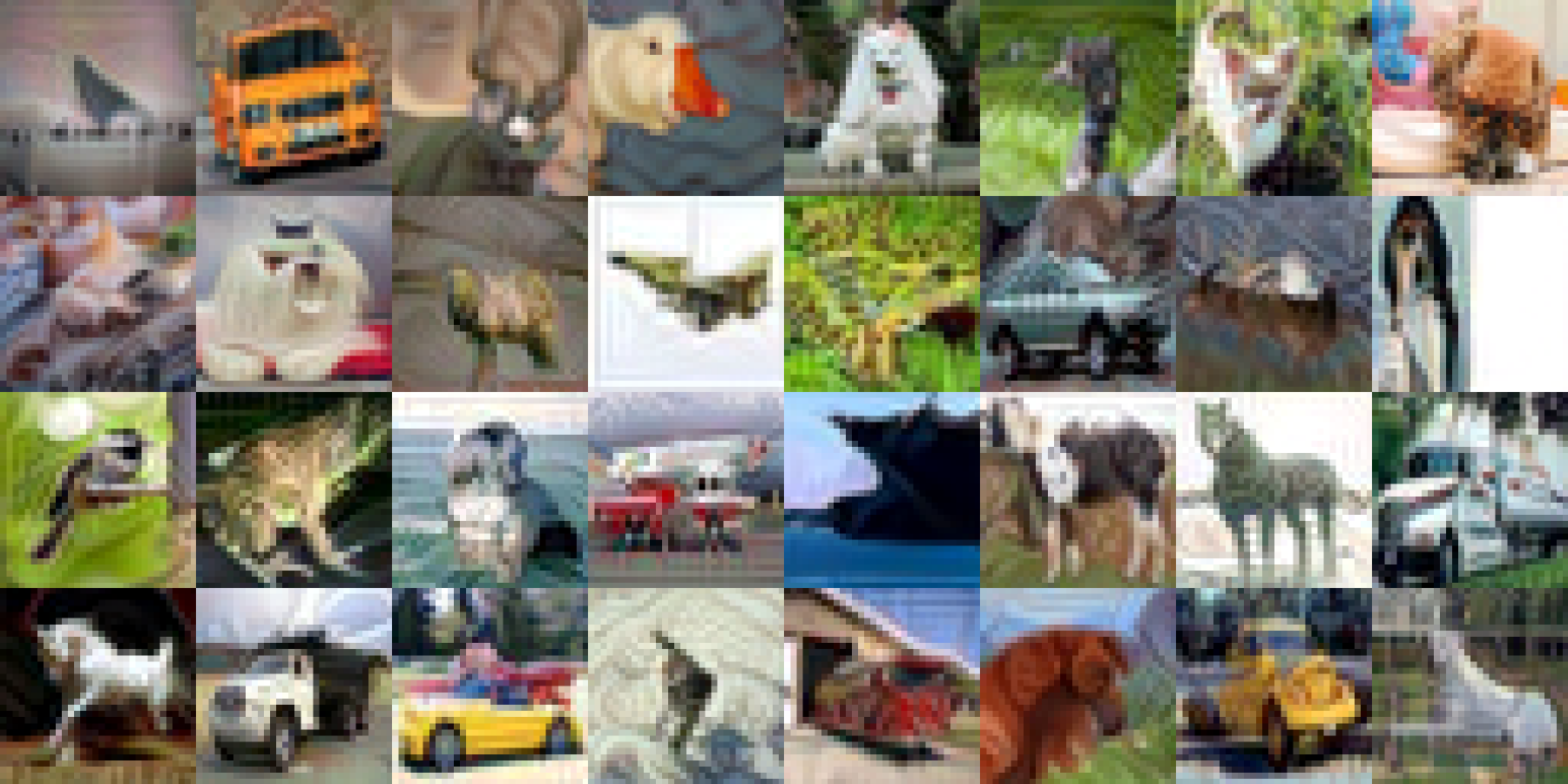}\\
    \small (b) Adversarial examples generated from the original images.

    \vspace{0.15em}

    \includegraphics[width=0.6\linewidth]{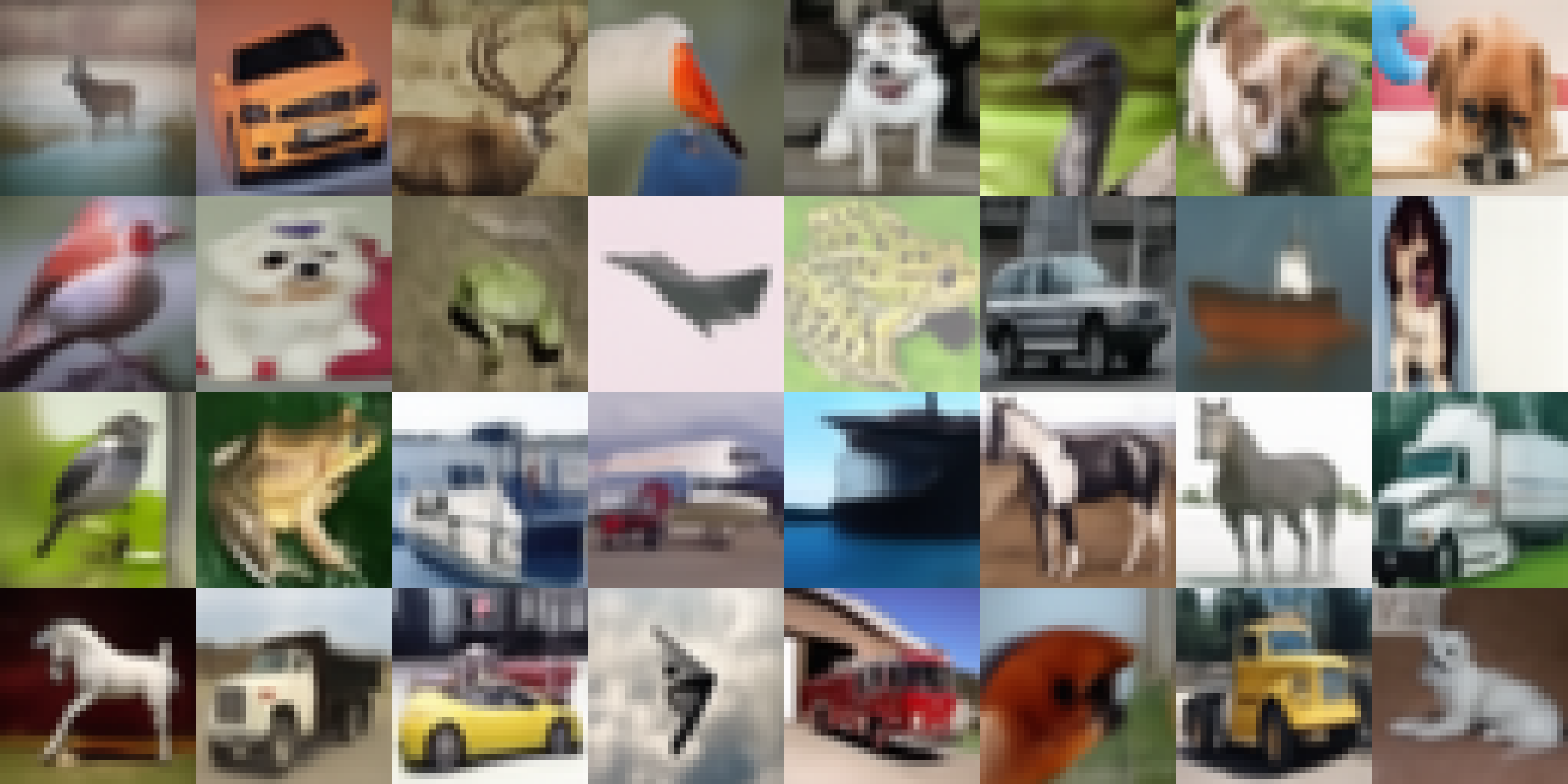}\\
    \small (c) Purified examples obtained by \textbf{GNSP}.

    \vspace{0.15em}

    \includegraphics[width=0.6\linewidth]{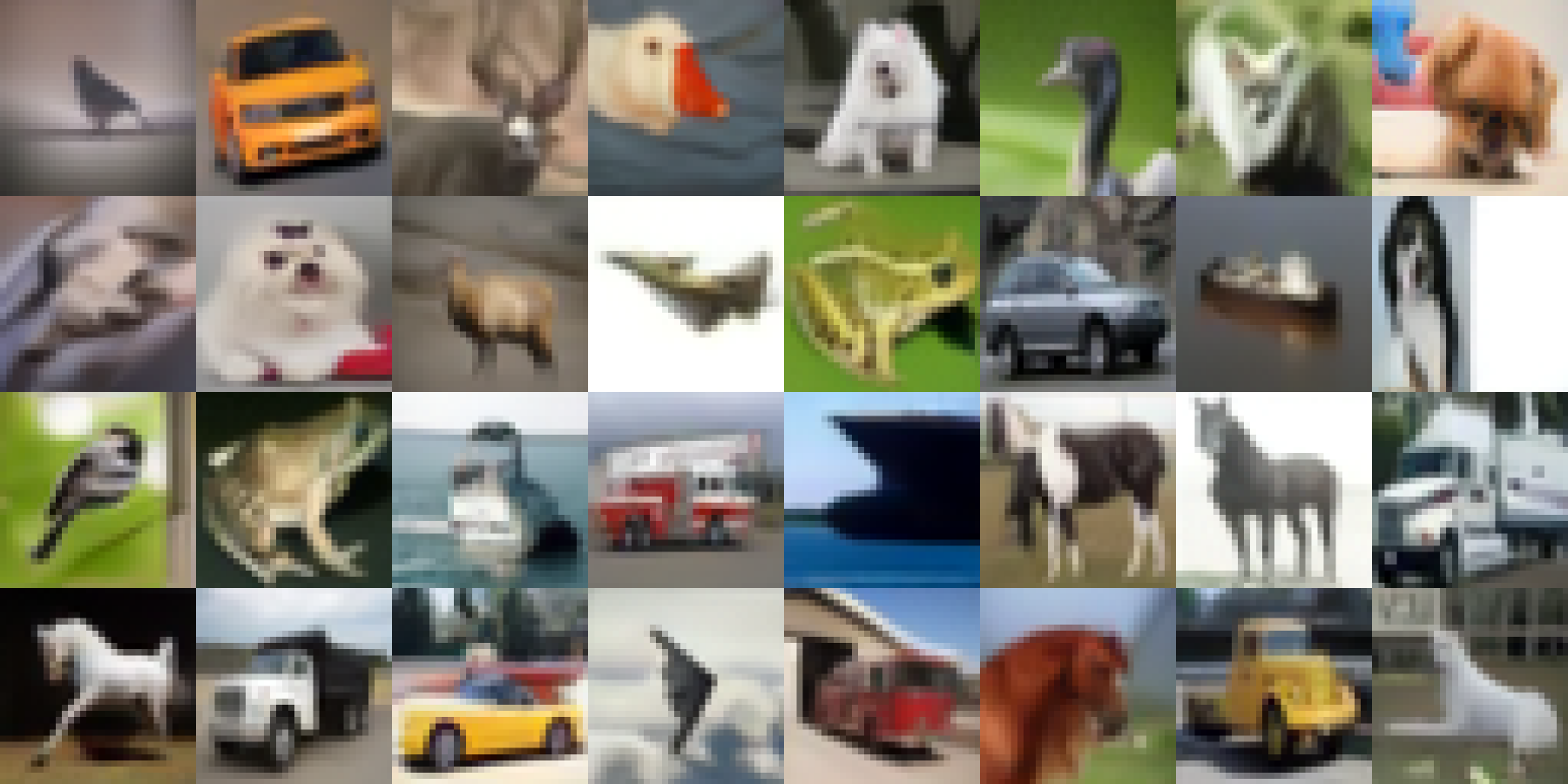}\\
    \small (f) Purified examples obtained by \textbf{CMAP (Ours)}.

    \caption{ {Qualitative evaluation and visual comparison with GNSP \cite{lee2023robust} on CIFAR-10 against PGD+EOT attacks with $\epsilon=16/255$.}}
    \label{fig:comparison_part_cifar_high}
\end{figure*}

\end{document}